\documentclass{article}
\usepackage{arxiv}
\usepackage{nicefrac}       
\usepackage{microtype}      
\usepackage{booktabs}       
\usepackage{doi}
\usepackage{adjustbox}
\usepackage{float}
\usepackage[T1]{fontenc}
\usepackage[utf8]{inputenc}
\usepackage{multicol}
\usepackage{multirow}
\usepackage[normalem]{ulem}

 \usepackage{booktabs}
 \usepackage{graphicx}
 \usepackage{float}
 \usepackage{tabularx}
 \usepackage{hyperref}

 \usepackage[numbers]{natbib} 
 
\usepackage{makecell}

\usepackage{multicol}

\usepackage{booktabs}
\usepackage{multirow}

\title{Explainable artificial intelligence (XAI): from inherent explainability to large language models}
\author{Fuseini Mumuni\thanks{University of Mines and Technology (UMaT), Tarkwa, Ghana} \and \textbf{Alhassan Mumuni}\thanks{Cape Coast Technical University, Cape Coast, Ghana.}}

\date{}

\begin{document}
\maketitle




\begin{abstract}
Artificial Intelligence (AI) has continued to achieve tremendous success in recent times. However, the decision logic of these frameworks is often not transparent, making it difficult for stakeholders to understand, interpret or explain their behavior. This limitation hinders trust in machine learning systems and causes a general reluctance towards their adoption in practical applications, particularly in mission-critical domains like healthcare and autonomous driving. Explainable AI (XAI) techniques facilitate the explainability or interpretability of machine learning models, enabling users to discern the basis of the decision and possibly avert undesirable behavior. This comprehensive survey details the advancements of explainable AI methods, from inherently interpretable models to modern approaches for achieving interpretability of various black box models, including large language models (LLMs). Additionally, we review explainable AI techniques that leverage LLM and vision-language model (VLM) frameworks to automate or improve the explainability of other machine learning models. The use of LLM and VLM as interpretability methods particularly enables high-level, semantically meaningful explanations of model decisions and behavior. Throughout the paper, we highlight the scientific principles, strengths and weaknesses of state-of-the-art methods and outline different areas of improvement. Where appropriate, we also present qualitative and quantitative comparison results of various methods to show how they compare.  Finally, we discuss the key challenges of XAI and directions for future research.
\end{abstract}


\begin{multicols}{2}

\section{INTRODUCTION}

\subsection{Machine learning and explainable AI}

Artificial Intelligence (AI) has received significant attention in recent times due to its immense power and remarkable potential to solve seemingly intractable problems across various real-world domains. Machine learning enables AI systems to extract relevant relationships from data through a training process, and then use this knowledge to make accurate inference on new instances during operation. Several machine learning methods have been developed and used extensively in diverse practical applications. In some of these applications (e.g.,  \cite{nam2019development,haenssle2018man,zhou2021ensembled,brinker2019deep}) the accuracy of machine learning models has reached parity with or even surpassed human experts. Deep learning methods have generally attained far higher accuracies than other machine learning models. Unfortunately, this performance gain comes at a cost of excessive data demand as well as overall model complexity that obfuscates the decision logic of the machine learning system.  Thus, most data-driven deep learning methods are characteristically black box in nature, where inputs are mapped directly to predictions without any accompanying information showing how the decisions are arrived at. However, for high-stake applications in domains like healthcare, finance, and robotics, it is necessary for stakeholders to understand the underlying rationale behind the decisions of artificial intelligence systems.  For this reason, there has been an increasing need to use transparent or \textit{interpretable} models whose predictions can be understood by developers and other relevant stakeholders such as regulators, managers, and end users.  In cases where models are not already interpretable, it is often necessary to devise methods to explain their \textbf{behavior or predictions}. Explainable artificial intelligence (XAI) aims at achieving this goal.

\subsection{Interpretability or explainability?}

Interpretability and explainability are closely related concepts, and these terms are often used interchangeably. However, several authors (e.g., \cite{lipton2018the,gilpin2018explaining,montavon2018methods,rudin2019stop,marcinkevičs2020interpretability})  have attempted to clarify their respective meanings and subtle differences. And, although there is not yet a universal consensus on what the exact technical definitions of these terms should be, a fuzzy notion of what they mean is not a subject in dispute. Specifically, an interpretable model is considered to be one which highlights some link between input and output spaces, or one whose prediction mechanism is obvious and can be understood from its internal architecture or working principle. For example, a sparse decision tree maps its inputs to outputs in an intuitive way. In contrast, explanability is generally understood to mean the ability of a model (or its surrogate) to provide explicit descriptions of what informs the decisions by means of logical rules, question-answering, or similarly rich modes of human-understandable information. Despite differences in meaning, we use these terms loosely and interchangeably in this work, as is commonly practiced in the literature. 

\subsection{Benefits of Explainable AI}

Some of the most important benefits and advantages of using XAI methods in machine learning include:

\begin{enumerate}
	\item \textit{Transparency} \cite{ehsan2021expanding,waltl2018increasing,gyevnar2023bridging}: Explainable AI\textbf{ } enhances transparency of machine learning systems, fostering understanding by human stakeholders. Understanding the operation of artificial intelligent systems allows developers and users to maximize their strengths while minimizing potential risks.

\end{enumerate}
\begin{enumerate}
	\item \textit{Trust} \cite{shin2021the}, \cite{leichtmann2023effects}: Understanding the decisions of machine learning models enhances trust in these systems and increases the willingness of decision-makers to adopt them\textbf{} in mission-critical applications.

	\item \textit{ Fairness} \cite{deck2024a}: Explainability methods can help to uncover bias in model predictions and, thus, ensure that the resulting decisions fair to all stakeholders irrespective of gender, race, religion or any disadvantaged group.

\end{enumerate}
\begin{enumerate}
	\item \textit{Safety} \cite{kuznietsov2024explainable}, \cite{atakishiyev2024safety}: Explainable AI can reveal behaviors of machine learning systems which do not meet safety requirements, allowing these problems to be addressed prior to, or even after deployment. This ultimately leads to improved safety. Similarly, explainability can help to ensure that machine learning models adhere to ethical standards and good practices.

\end{enumerate}
\begin{enumerate}
	\item \textit{Accountability}\textbf{ } \cite{doshivelez2017accountability}:When decisions of machine learning models lead to catastrophic outcomes, model explanations can help to establish the rationales of such decisions or understand why the model failed and whether  it is caused by error or negligence on the part of any stakeholder.  This eventually enables culpable parties to be held accountable for any losses resulting from poor design or misuse.

\end{enumerate}
\textit{ (vi) Model debugging}\textbf{ } \cite{anders2022finding},  \cite{schramowski2020making}: Interpretability of machine learning systems can allow undesirable behavior to be detected and corrected through subsequent refinement. Such refinements can improve prediction accuracy and prevent undesirable prediction outcomes that may arise when a model learns spurious correlations from data. This problem, known as the Clever Hans  \cite{lapuschkin2019unmasking} is encountered frequently  in machine learning.

\subsection{Important XAI Concepts}

We briefly described some of the most important concepts of explainable AI in this subsection. These concepts are used extensively in the literature and throughout this paper. 

\begin{figure*}[!htb]
	\vspace {1mm}
	\centering
	\includegraphics[width=1.0 \linewidth]{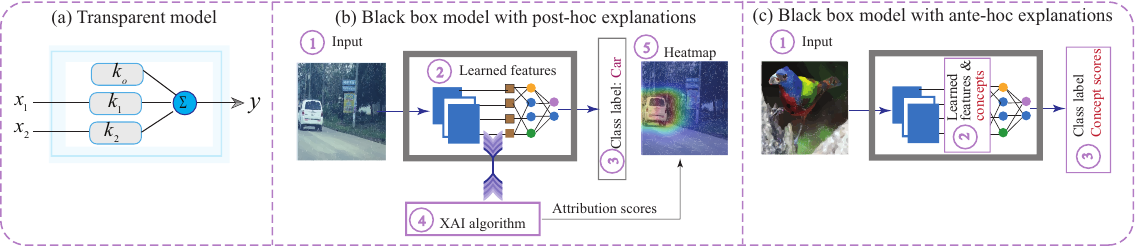} \vspace {-3mm}
	\caption{Illustration of (a) white-box model, (b) and (c) black-box model explainable by post-hoc and ante-hoc methods, respectively. The circled numbers illustrate the sequence of operations from input to prediction and explanations.  
	}\label{fig:1_XAI_Concepts}
\end{figure*}

\begin{enumerate}
	\item \textit{Scope:}\textbf{ }Explanations can have a \textit{global} or \textit{local} scope depending on their coverage.  Global explanations provide insights on a model’s behavior in general whereas local explanations focus on explaining individual predictions. Some explainability methods can explain both global behavior and local instances at the same time.

	\item \textit{Applicability}: An explainability method is described as \textit{model-agnostic} if its implementation is independent of the model it explains. On the other hand, model-specific methods use techniques whose implementations are specific to the machine learning model being explained.  One advantage of \textit{model-specific} explainability methods is that the approach may allow details of the machine learning model to be incorporated to obtain better and tailored explanations. However, some of these techniques may interfere with the original black-box model in ways that degrade prediction performance. In general, model- agnostic explainability methods are more versatile and can be applied to any black box model without ``opening it".

	\item \textit{Implementation stage: }Interpretability methods can be incorporated into machine learning systems at different stages. This leads to two broad families of approach: \textit{Post-hoc }interpretability and \textit{ante-hoc }interpretability\textit{Post-hoc } interpretability methods are applied after training the model. Majority of XAI techniques fall in this group. The key idea of post-hoc\textit{ }explainability is to explain model behavior in an unobtrusive manner. However, the approach may compromise faithfulness as the explanation may not always follow the true rationale of predictions. \textit{Ante-hoc }interpretability techniques (e.g., concept bottleneck models   \cite{koh2020concept} seek to endow models with explainability at training or design time  by simultaneously learning features for classification as well as features that influence predictions. The approach enhances faithfulness but some of the techniques used may impose additional constraints that ultimately harm prediction accuracy.

\end{enumerate}
 Some authors classify \textit{ante-hoc} explainability methods as i\textit{inherently} interpretable, however, although these models acquire explainability at design time, they require explicit interventions in their design to achieve this, unlike true \textit{inherently} interpretable models. Figure \ref{fig:1_XAI_Concepts} illustrates some of the important concepts in XAI.


\begin{figure}[H]
	\vspace {-1mm}
	\centering
	\includegraphics[width=1.0 \linewidth]{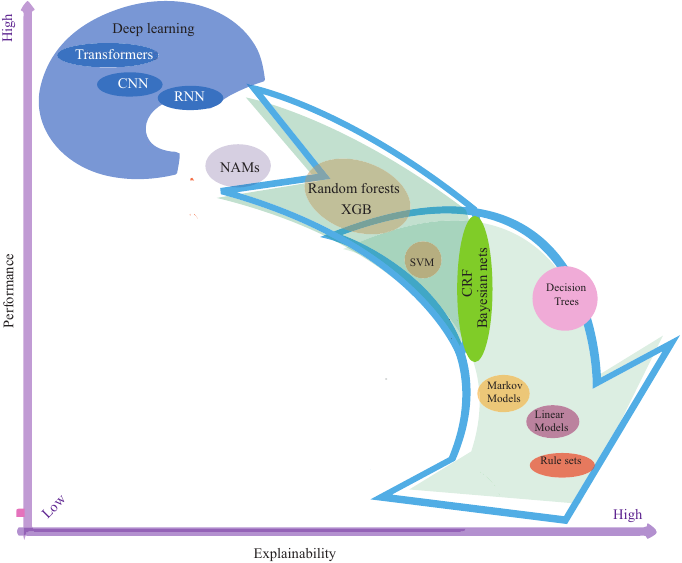} \vspace {-3mm}
	\caption{Accuracy-interpretability relationship for different families of machine learning models. Although the illustration captures simple rule-based and linear models at the lower end of the accuracy spectrum, it should however, be noted that this assumes that the models are applied to complex problems. For simpler problems, these simple models achieve competitive results. 
	}\label{fig:2_Acc_vs_XAI}
\end{figure}

\subsection{Accuracy vs interpretability}

\textit{Inherently or intrinsically interpretable} models are relatively simple frameworks and by virtue of this property, the basis of their predictions can be intuitively understood. Intrinsic interpretability is exhibited in varying degrees across different model families, as shown in Figure \ref{fig:2_Acc_vs_XAI}.  The most interpretable family of models include rule sets, decision trees and linear models. Owing to their simplicity they often yield modest or relatively low accuracies, particularly on complex tasks.

On the other hand, larger and more complex models like deep neural networks possess the capacity to capture desirable information from high dimensional training data and thus, achieve impressive accuracies but their decisions are not readily understood by stakeholders. They are therefore referred to as \textit{opaque} or \textit{black box} models.  As Figure \ref{fig:2_Acc_vs_XAI} shows, the explainability and accuracies of machine learning models generally tend to trend in opposite directions, i.e., the higher the accuracy, the less interpretable. This accuracy-explainability tradeoff, illustrated in Figure \ref{fig:2_Acc_vs_XAI}, is a widely studied phenomenon  \cite{crook2023revisiting},  \cite{assis2025the},  \cite{bell2022its}) with some efforts  \cite{luo2019balancing},  \cite{mori2018balancing}aimed  at carefully balancing these competing goals.

\subsection{Motivation and work outline}

Owing to the practical importance and the ever-increasing volume of work on explainable artificial intelligence, there exist many authoritative surveys (e.g.,  \cite{rudin2021interpretable},  \cite{dwivedi2023explainable},  \cite{adadi2018peeking},  \cite{schwalbe2023a},  \cite{barredoarrieta2020explainable}) on this subject.  There are also reviews dedicated to large language models  \cite{singh2024rethinking},  \cite{zhao2024explainability}or specialized domains like medical  or healthcare applications  \cite{chaddad2023survey},  \cite{vandervelden2022explainable},  \cite{tjoa2021a}. In addition,  another recent study  \cite{mumuni2024improving} reviews state-of-the-art methods that utilize prior knowledge representation in various forms (including logical rules, knowledge graphs) and LLMs to improve the explainability of deep learning models. The study  \cite{mumuni2024improving}also discusses techniques that employ prior  knowledge to improve adversarial robustness and zero-shot generalization. Although there is a proliferation of well-written reviews on the subject, most of the existing works focus on more general concepts of explainability and often present a high-level description of the relevant methods. Only a few works (e.g.,  \cite{dwivedi2023explainable},  \cite{barredoarrieta2020explainable}) provide   adequate depth or present the specific techniques in adequate detail (this includes explaining the working principle or clearly highlighting their strengths and weaknesses). Moreover, to the best of our knowledge, no surveys —not even the most recently published studies at the time of submission of this work — cover some of the emerging but hugely important developments in the field. In particular, no works emphasize the increasing use of vision-language models and large language models in enhancing and automating the explainability of other black box models. For these reasons, we are motivated to fill the gap by comprehensively presenting, and discussing the pertinent issues on these methods. We present a comprehensive review of methods covering all categories of interpretability methods. We dedicate a whole section (section 5) of this work for the approaches that use vision language models and LLMs in their interpretability pipelines. Uniquely, we also present detailed quantitative results on some of the most popular methods. 

\begin{figure*}[!htb]
	\vspace {1mm}
	\centering
	\includegraphics[width=1.0 \linewidth]{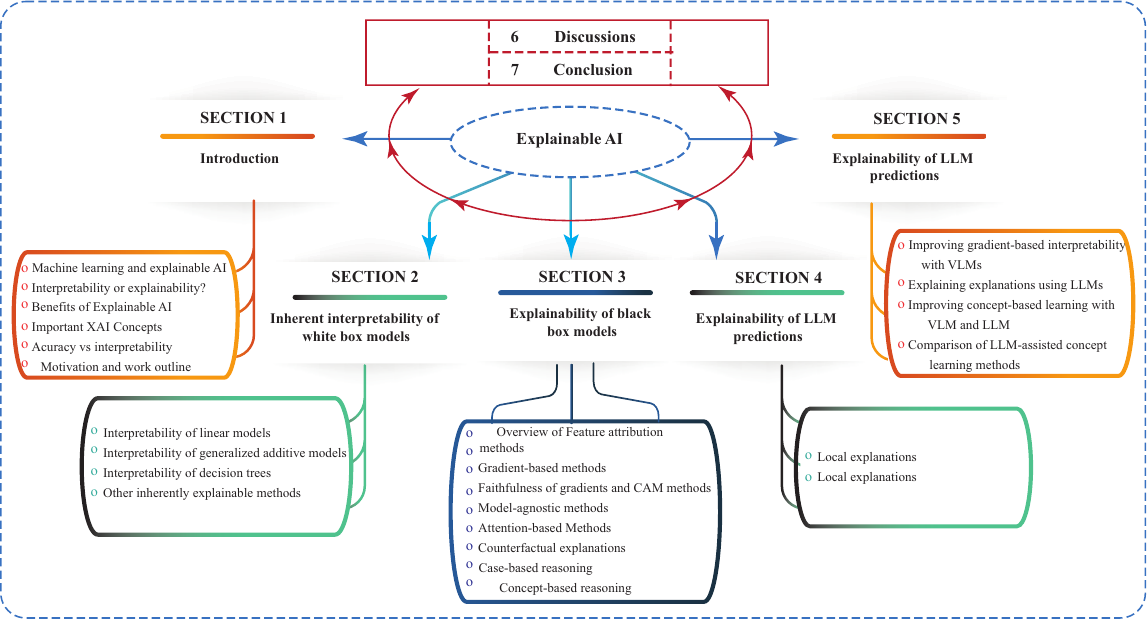} \vspace {-3mm}
	\caption{Outline of paper.  
	}\label{fig:3_Outline}
\end{figure*}

This paper is structured as follows: Section 1 presents general introduction to the study. Section 2 covers intrinsically interpretable frameworks, their key principles, challenges and workarounds. Section 3 covers methods of explaining general black box models. The interpretability of large language models is presented in detail in Section 4. In Section 5 we present techniques that leverage vision-language and large language models for improving and automating explainable AI. In Section 6 we discuss important issues, current state and probable and needed developments in the future. We conclude the work in Section 7 by summarizing the important points. A more detailed outline of the outline of this work is provided in Figure \ref{fig:3_Outline}.


\section{INHERENT INTERPRETABILITY OF WHITE-BOX MODELS}

Research in explainable artificial intelligence is gaining increasing momentum owing to the importance of this field and the opportunities it presents. However, some models are already inherently interpretable by their very design. In other words, they are self-explainable in that they allow developers or users to gauge into their inner workings or decision-making rationale without making additional modifications specifically for this purpose. These inherently interpretable or \textit{white box} models are widely used to solve problems in diverse machine learning tasks. Since interpretability is intrinsic to the models, the explanations are likely to be more faithful to the decision of the models. Additionally, no extra design or training time is invested in ensuring interpretability.  This category of models includes linear models, generalized additive models, decision trees, Bayesian networks, etc. We describe the most popular ones here.

\subsection{Interpretability of linear models}

Linear models are a class of machine learning frameworks that capture input features describing relevant properties and accompanying coefficients or weights that specify the relative strength of each feature.  For linear models, as shown by the simplified equation  \cite{nam2019development}, the contribution of each feature (\textit{x}\textsubscript{1}, \textit{x}\textsubscript{2}, etc.) to the final prediction \textit{y} can be intuitively understood by observing the corresponding coefficients (\textit{k\textsubscript{1}}, \textit{k\textsubscript{2}}, etc.) that weight the features or descriptors. In the equation \textit{a}\textsubscript{0} is a baseline. The magnitude and sign of the coefficient respectively describe the degree and direction (i.e., whether a higher value of the descriptor lowers or amplifies the outcome) of the influence a particular feature exerts on the model’s prediction. 

$$y = {a_0} + {k_1}{x_1} + {k_2}{x_2} +  \cdots  + {k_n}{x_n}$$

 Thus, the interpretability of this class of models is straightforward from their intrinsic components and the understanding of the input data. However, in most practical settings, interpretability of simple linear models is complicated without additional workarounds  \cite{schielzeth2010simple},  \cite{lipton2016},  \cite{salih2024are}. We briefly discuss the common scenarios in the next paragraphs.

 By keeping all other input variables constant, regression coefficients in linear models can be analyzed to understand the contribution of a given feature to the output. Also, zero-order correlation coefficients  \cite{onwuegbuzie1999uses} can help to reveal the importance of a given feature to the model’s  prediction without considering the influence of all other remaining features. This interpretation approach assumes that the input variables are uncorrelated. However, in most practical situations, the presence of feature correlations is inevitable. Feature correlation or collinearity is a property whereby a certain input feature is a function of another. Under this condition, coefficient values no longer provide adequate information about the contribution of input features to the output. Hence, collinearity may undermine the validity of interpretability methods if this behavior is not controlled for. 

 To improve the interpretability of linear regression models under in collinearity conditions, more tailored techniques are required. For instance, Commonality analysis  \cite{raymukherjee2014using},  \cite{nimon2011regression},  \cite{zientek2006commonality} allows the variance in the output to be partitioned into the components associated with each predictor exclusively, and the components of the variance that is due to the interaction (or common effects) of input features. 

   Commonality analysis uses ccommonality coefficients to extract information on predictors and their interactions. Negative commonality values for a model may represent suppressor relations, where the variables act to suppress confounding variance in other predictors, and thus, increase these predictors’ contributions to the model’s overall variance \cite{reichwein2006commonality}. 

The presence of covariates  \cite{cattaneo2018inference},  \cite{atem2017linear} may also undermine the natural interpretability of machine learning frameworks such as linear regression models. Covariates are confounding factors or variables that often mask the true effect of input features. Covariates are especially common in medical domains where extraneous factors beyond those being investigated can significantly influence the predictions of machine learning models. For example, a machine learning model designed to predict the outcome of a new treatment may be influenced by a host of unrelated factors such as a patient’s diet, stress levels and even genetics. In cases like this, the observed features alone cannot explain the behavior of the model when important covariates are not adequately accounted for. Fortunately, many effective techniques  \cite{schielzeth2010simple},  \cite{ning2020robust},  \cite{belloni2014inference},  \cite{angrist2004when} are available for addressing this challenge.  The popular ones include propensity score matching and treating covariates as additional input variables.  

Besides model complexity or size, the interpretability of linear models also depends on the meaningfulness of input features. However, in some application domains, raw feature values may not readily be informative from human perspective.  Moreover, in applications where inputs are high-dimensional or represent low-level features, it becomes challenging for humans to understand the importance of individual input units in the model’s decision. For example, in image classification, pixels are the fundamental variables that are processed to determine the outcome. Yet, individual pixels in isolation do not provide relevant clues on the model’s decision. Furthermore, approaches focusing on individual pixels for interpretability would be vulnerable to noise arising from variations in pixel values that do not affect the perceptual quality of the given image.   

Data misalignment: machine learning models may use different kinds of data or data from different sources during training or at test time. In such situations, the data may not be aligned in terms of their distribution or scale. This may pose significant challenges for linear models which are required to be interpretable. Intrinsic interpretability can be hindered when some skewed inputs in a certain dataset unduly influence the prediction. Similarly, data points of larger values can dominate and obscure smaller ones. These problems in machine learning are usually addressed by data preprocessing strategies such as data cleaning, normalization and standardization. Current approaches of implementing these techniques are quite laborious and inefficient, although automated data preprocessing methods are now available than can achieve impressive results with minimal effort. These automated data pre-processing methods, discussed extensively in a recent review  \cite{mumuni2024automated}, are now beginning to have a significant impact on improving the efficiency of handling undesirable data in machine learning. 

 Although the techniques may improve prediction accuracy and enable the machine learning model to attend to all data points in an equitable manner, they are still not guaranteed to restore interpretability. This is because the preprocessing may change the semantic meaning (e.g., dimension or unit) of some inputs.

\subsection{Interpretability of generalized additive models}

Generalized additive models (GAMs)  \cite{hastie2017generalized},  \cite{hastie1987generalized}are a class of machine learning frameworks where the expected value of the response or prediction is a linear combination of features which are independently modeled by smooth functions, also known as  shape functions. The functions describe each predictor’s influence on the output, and can be non-linear, thus, allowing more intricate behaviors to be captured. In principle, of a generalized additive model can be viewed as a multiple linear model where the weighting coefficients for features\textit{\textbf{ x}} are replaced with appropriate functions \textit{f\textsubscript{1}}, \textit{f\textsubscript{2}}, etc., as shown in equation (2), where \textit{F}(\textit{x}) is the output function an\textit{ }$\beta$ is a bias term.

$$F(x) = {\beta _0} + {f_1}({x_1}) + {f_2}({x_2}) +  \cdots  + {f_n}({x_n})$$

\begin{figure}[H]
	\vspace {-1mm}
	\centering
	\includegraphics[width=1.0 \linewidth]{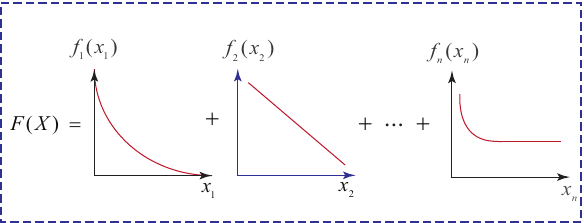} \vspace {-3mm}
	\caption{Generalized additive model composed of smoothed linear and non-linear functions. 
	}\label{fig:Interpret_LinearModels}
\end{figure}

The relevant functions are learned exclusively from training data without the overly restrictive assumption of constant weighting, as is the case for conventional linear models. This flexibility enables better approximation of intricate patterns and relations that map features in the training data to outputs, even when these relationships are non-linear. Figure \ref{fig:Interpret_LinearModels} shows a simple graphical example of a GAM.

The interpretability of GAMs stems from the fact that the outputs of these models are derived from a linear combination of smooth functions, each of which describing a relation that projects a single input feature to the output. By plotting each relation as a function of a specific input feature, the contribution of individual features to the prediction can be visualize separately. For instance, in Figure \ref{fig:Interpret_LinearModels}, it can be seen that the expected value of the response variable \textit{Y} changes linearly in reverse correlation with the input variable \textit{x}\textsubscript{2} when all other variables are kept unchanged. 

 GAMs are highly interpretable if they are adequately simple. This is usually the case when the models are sparse with a small number of shape functions which must necessarily be smooth.  Regularization has been successfully employed to control either sparsity (e.g.,  \cite{ravikumar2025sparse},  \cite{lou2016sparse}), smoothness (e.g.,  \cite{wood2004stable}) or both qualities ( \cite{meier2009high},  \cite{lin2006component}) simultaneously. However, enforcing sparsity and smoothness constraints may compromise accuracy. In particular, higher sparsity can reduce the model’s capacity to learn complex relationships in high-dimensional data while smoothness may limit the ability to model natural discontinuities in data.  Additionally, conventional GAM frameworks assume that input features do not interact. This assumption leads to simple designs that are easy to explain but the enhanced interpretability may compromise accuracy. Consequently, Lou et al. propose to include pairwise interactions of features in the GA2M   \cite{lou2013accurate} extension. The authors show that these two-dimensional interactions substantially improve accuracies of GAMs while still maintaining interpretability.

Another difficulty with the interpretability of generalized additive models concerns how to handle multiclass prediction tasks. With binary classification problems, model decisions are easy to explain by analyzing the shape functions that map input features to final outputs.  However, for multiclass problems, interpretability remains a challenge even for relatively simple models. Specifically, in the multiclass case, the contribution of a particular feature to the final class prediction cannot be understood simply by observing the projection from its shape function to the given class score. In fact, a feature may increase the logit for a given class but reduce the final score for the class after normalization. This can happen if the feature increases the logits for other classes by a larger amount. Thus, multiclass prediction problems pose a unique explainability challenge. 

To address this difficulty, Zhang et al. \cite{zhang2019axiomatic} propose a post-processing technique that enforces compliance of GAMs with two key interpretability axioms: \textit{monotonicity}, which imposes monotonicity constraint on predictor functions and the average class score; and \textit{smoothness}, which requires shape functions to be devoid of noise-induced discontinuities. The method transforms a less interpretable multiclass prediction model into a more interpretable one in a post-hoc manner. 

\subsection{Interpretability of decision trees}

Decision trees are simple but effective machine learning models that have achieved tremendous success in a wide range of tasks, mainly within classification and regression domains. They enjoy a long-standing popularity dating back many decades  \cite{loh2014fifty},  \cite{quinlan1986induction}. Besides their performance, decision trees are also known for their transparency. That is, the internal structures of these models directly provide insights on their predictions. Owing to these qualities, there is significant research interest in interpretability-related properties of decision trees such as model depth, number of nodes, and sparsity.

\begin{figure}[H]
	\vspace {-1mm}
	\centering
	\includegraphics[width=1.0 \linewidth]{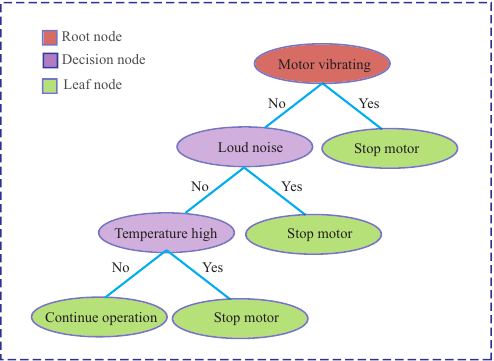} \vspace {-3mm}
	\caption{A simple representation of a decision tree showing hierarchical rule-based structure. 
	}\label{fig:5Interpret_DecTrees}
\end{figure}

  \textit{Decision trees } are organized in hierarchical tree structures with nodes implementing tests of logical conditions, branches denoting the results of these tests, and leaf nodes denoting final labels or predictions that satisfy the evaluated conditions leading to these nodes. Figure \ref{fig:5Interpret_DecTrees} shows the simplified view of a decision tree. A given outcome or prediction of a decision tree can be explained simply by tracing the decision path to the point where the prediction is made.

Traditional decision tree construction methods employ splitting strategies that rely on greedy algorithms. Popular among these techniques are CART  \cite{loh2011classification} and C4.5  \cite{quinlan2025google},  \cite{quinlan1996improved} frameworks  which construct trees in a top-down manner do not provide effective mechanisms to deal with errors at top hierarchies when they are observed later in the construction process. Furthermore, the approaches often lead to only locally optimal trees as these methods do not usually consider future states, and hence struggle to ensure global optimality.

One of the most significant limitations of decision trees is their poor generalization ability. Decision trees tend to easily \textit{overfit} their training data and thus experience a significant performance drop when tested on new data. A model is said to overfit if it learns to produce sufficiently accurate predictions on training data but fails to perform adequately on unseen data. For decision trees, this can happen when the tree is larger (i.e., deeper tree with more branches) and more complex, and thus has more capacity and sophistication to capture more detailed and intricate characteristics (sometimes caused by noise or outliers) specific to the training data rather than general attributes. Some complex trees are constructed as an ensemble of trees involving a large number of subtrees. Conversely, overly simple decision trees do not possess the capacity to capture sufficient number of relevant features from data, and are therefore neither able to perform well on the training set nor generalize to unseen data.

In practice, there are techniques available for mitigating both overfitting (e.g.,  \cite{amro2021instance}) and underfitting (e.g.,  \cite{czajkowski2019decision}) of decision trees. However, considering all factors, larger trees generally achieve better accuracies than simpler or \textit{sparse} decision trees, especially on more complex tasks. Sparsity, on the hand, favors transparency and is therefore regarded as a desirable property from explainability point of view. Consequently, research interest in sparse decision trees remains high. 

  Optimal Sparse Decision Trees (OSDT) \cite{hu2019optimal} and Generalized and Scalable Optimal Sparse Decision Trees (GOSDT) \cite{lin2020generalized} are state-of-the-art methods of achieving sparsity without severely compromising performance. In particular, GOSDT generates sparse and globally optimal decision trees on multiple objectives. The method employs dynamic programming paradigm that enables efficient exploration of the search space of all possible subtrees. GOSDT also addresses the problem of imbalanced data. Although GOSDT is relatively fast, it still takes a significant amount of training time when dealing with high dimensional data. McTavish et al. \cite{mctavish2022fast} propose to reduce the runtime by using guessing technique to mimic a high-performing black-box reference model.

Although a high proportion of studies on the interpretability of decision trees tend to focus on improving their white-box characteristics, some studies attempt to engineer post-hoc methods to interpret opaque trees. For instance, TreeExplainer  \cite{lundberg2020from} extracts local explanations from individual predictions of tree models using S HAP scores   \cite{lundberg2017a}, and then combines multiple local explanations to obtain global interpretability. The method also detects feature interactions using the Shapley interaction index. In contrast to designing decision trees for inherent interpretability, methods based on post-hoc explainability can allow developers to fully focus on performance related metrics. This will reduce the need for increasing model sparsity, which largely harms performance.  However, potential compromise in faithfulness a price to pay for adopting post-hoc explanations in favor of inherent explainability. 

Popular rule-based interpretability approaches like rule sets  \cite{frank1998generating},  \cite{ciaperoni2023concise} are constructed from simple if-then rules that are evaluated to arrive at a decision or prediction.  In classification problems, rule-based methods can be used to learn discriminative rules that relate input features to category labels. Since symbolic rules are easy to understand, methods that utilize these constructs are inherently explainable. Their main limitation is their inability to handle more complex problems that involve high-dimensional data like images. In these domains, rule extraction methods (e.g., extraction methods (e.g.,  \cite{tuo2025interpretable}) can be used to mine symbolic rules from a black box model like neural network that is pre-trained on the relevant data. However, the interpretability of extracted rules may be limited especially for large and complex networks.

\subsection{Other inherently explainable methods}

\subsubsection{Neural additive models}

\begin{figure}[H]
	\vspace {-1mm}
	\centering
	\includegraphics[width=1.0 \linewidth]{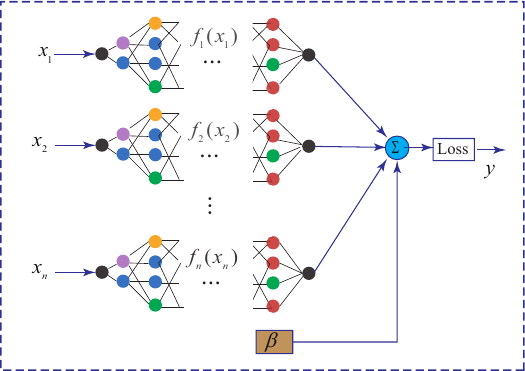} \vspace {-5mm}
	\caption{An illustration of the structure of a neural additive model with input features ${x_i}$ and bias $\beta$. Each sub-network $f\left({{x_i}} \right)$  processes a single feature and the results combines linearly in addition to the bias $\beta$ to yield the output \textit{y}. 
	}\label{fig:6_Additive_Models}
\end{figure}

Neural Additive Models (NAMs)  \cite{zhang2024gaussian},  \cite{r2021neural} represent another class of intrinsically interpretable models  capable of higher performance on complex tasks compared to other intrinsically interpretable models like linear models and GAMs. A Neural Additive Model is essentially a neural network which is composed of sub-networks each of which is responsible for handling a single input feature (see Figure \ref{fig:6_Additive_Models}). The interpretability of NAMs stems from the fact that constituent sub-networks are simple and additive, i.e., the overall model is a linear combination of these components. Also, owing to the power of neural networks. NAMs have a performance advantage over other inherently interpretable methods like regression models, shallow decision trees and sparse generalized additive models. Figure \ref{fig:6_Additive_Models} depicts the concept of neural additive models and Table \ref{tab:T1_ExplainableAI_Principles} describes the features of various inherently interpretable models.

\subsubsection{Scorecards}

Another widely employed intrinsically interpretable method utilizes scorecards  \cite{knaus1991the},  \cite{ustun2016supersparse} for decision-making. Scorecards are linear prediction models that assign numeric weights or points to features to signify their relative importance to the final score. Features can be given positive or negative points to reflect the direction in which they influence the overall score. For  some predictors, scorecards typically use binary features to denote the presence or absence of a given feature. Scorecards are mostly suitable for tasks where predictors have a clear and unambiguous influence on the outcome. Application domains include handling electronic health data and data for credit risk assessment. The important features of this approach and the other common inherently explainable models presented in this section are summarized in Table \ref{tab:T1_ExplainableAI_Principles}.

\begin{table*}[htb!]
	
	\caption{Summary of the core principles, strengths and weaknesses of common inherently explainable models.}
	\label{tab:T1_ExplainableAI_Principles}

\renewcommand{\arraystretch}{1.3}
\begin{adjustbox}{max width=\textwidth}
	
\scalebox{0.90}{
\begin{tabular}{p{4.12cm}p{4.12cm}p{4.12cm}p{4.12cm}}
\hline
\multicolumn{1}{|p{4.12cm}}{\textbf{Method}} & 
\multicolumn{1}{|p{4.12cm}}{\textbf{Principle}} & 
\multicolumn{1}{|p{4.12cm}}{\textbf{Key strengths}} & 
\multicolumn{1}{|p{4.12cm}|}{\textbf{Limitations}} \\ 
\hline
\multicolumn{1}{|p{4.12cm}}{Linear Models (LMs)} & 
\multicolumn{1}{|p{4.12cm}}{Use linearly weighted combination of features. } & 
\multicolumn{1}{|p{4.12cm}}{Capable of handling continuous variables} & 
\multicolumn{1}{|p{4.12cm}|}{Limited to simple problems involving linear relationships \newline
} \\ 
\hline
\multicolumn{1}{|p{4.12cm}}{Generalized Additive Models (GAMs)} & 
\multicolumn{1}{|p{4.12cm}}{Use a superposition of smooth (non-) linear functions to capture the relationship between input and output.} & 
\multicolumn{1}{|p{4.12cm}}{Effective in handling continuous variables \newline
Can capture non-linear patterns, unlike linear models} & 
\multicolumn{1}{|p{4.12cm}|}{Limited to linear and logistic regression problems \newline
Difficult to capture nonlinear interactions} \\ 
\hline
\multicolumn{1}{|p{4.12cm}}{Decision Trees (DTs)} & 
\multicolumn{1}{|p{4.12cm}}{Use a hierarchical arrangement of rules organized in a tree structure} & 
\multicolumn{1}{|p{4.12cm}}{Powerful in handling nonlinear relations, simpler and more interpretable compared to LMs and GAMs of equivalent sophistication.} & 
\multicolumn{1}{|p{4.12cm}|}{Difficulty in dealing with continuous or non-discrete features} \\ 
\hline
\multicolumn{1}{|p{4.12cm}}{Neural Additive Models (NAMs)} & 
\multicolumn{1}{|p{4.12cm}}{Use linear combination of simple neural networks are used for the prediction model } & 
\multicolumn{1}{|p{4.12cm}}{Capable of modeling more complex relationships than linear models, GAMs and DTs. \newline
NAMs can achieve high accuracy owing to their use of neural networks} & 
\multicolumn{1}{|p{4.12cm}|}{Difficulty in dealing with continuous or non-discrete features} \\ 
\hline
\multicolumn{1}{|p{4.12cm}}{Rule-based methods} & 
\multicolumn{1}{|p{4.12cm}}{Learn prediction rules from data or extract rules from trained models} & 
\multicolumn{1}{|p{4.12cm}}{Highly interpretable \newline
Can work with low volumes of data \newline
Can leverage domain knowledge to improve performance} & 
\multicolumn{1}{|p{4.12cm}|}{Poor scalability  \newline
Better suited to cases features are binarized or in discrete intervals \newline
Difficult to capture feature interactions} \\ 
\hline
\multicolumn{1}{|p{4.12cm}}{Scorecards} & 
\multicolumn{1}{|p{4.12cm}}{Features are assigned scores which sum to give a final prediction} & 
\multicolumn{1}{|p{4.12cm}}{Can be easy to build  \newline
Can be constructed or improved by leveraging knowledge from domain experts } & 
\multicolumn{1}{|p{4.12cm}|}{Usually limited to features in binary forms or discrete intervals \newline
Complexity and size can hinder transparency \newline
Difficult to capture feature interactions} \\ 
\hline
\end{tabular}

} 
\end{adjustbox}
\end{table*}

\section{Explainability of black box models}

In this section, we detail the methods used to explain black box models.  Feature attribution methods are by far the commonest group of approaches. We review various sub-categories of this family of approaches, as well as other popular interpretability methods. Figure \ref{fig:7_Table_PPT1}    presents a summary of the categories of methods that we cover in detail in this section. Examples of techniques in each category is provided. 

\subsection{Overview of Feature attribution methods}

\begin{figure*}[!htb]
	\vspace {-1mm}
	\centering
	\includegraphics[width=1.0 \linewidth]{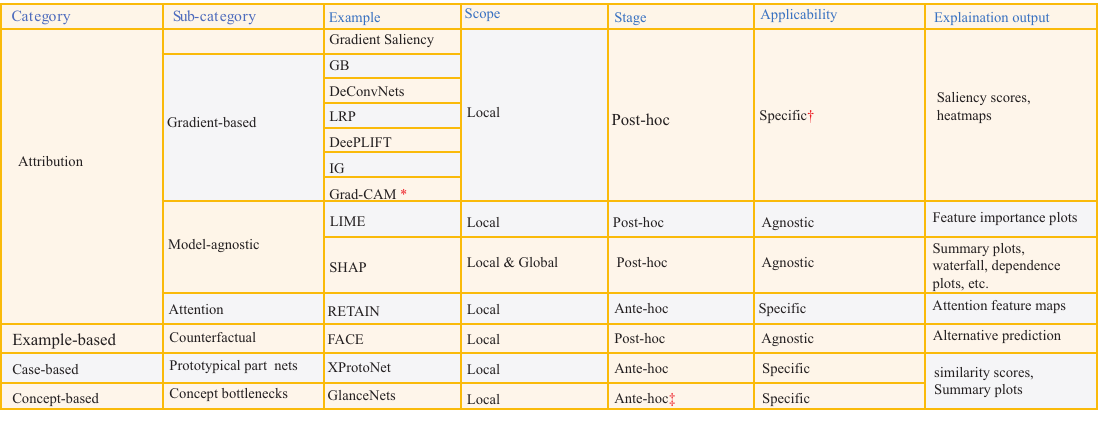} \vspace {-5mm}
	\caption{Summary of interpretability methods for black box models. $^{\ast}$Some class activation mapping techniques (e.g., Score-CAM \cite{wang2020score}) do not strictly use gradients; †Some studies classify gradient-based methods as model-agnostic, however, their implementation depends on the underlying model; $^{\ddagger}$Some concept bottleneck models use post-hoc approaches.
	}\label{fig:7_Table_PPT1}
\end{figure*}

\begin{figure*}[!htb]
	\vspace {1mm}
	\centering
	\includegraphics[width=1.0 \linewidth]{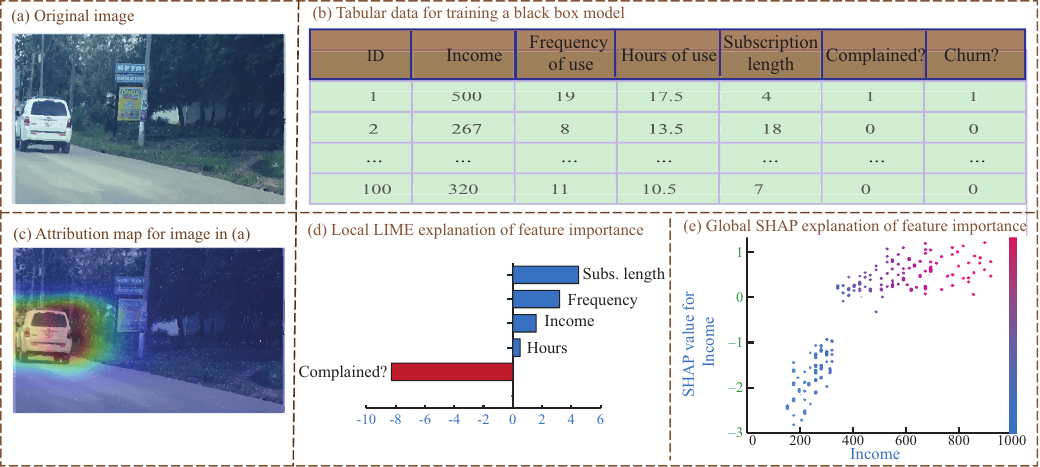} \vspace {-3mm}
	\caption{Examples of attribution-based interpretability methods. (a) is an input image to a classification model while (b) represents tabular data containing information on cable network subscribers. In (c), a CNN predicts the image of a car from (a) and a popular attribution method Grad-CAM \cite{selvaraju2017grad} highlights the region that influence the local decision. In (d), the LIME \cite{ribeiro2016explaining} framework plots the local feature importance showing which factors led to a prediction that a particular customer will span.   Finally, (e) plots the SHAP scores for the feature “Income” showing its global role in on the prediction.  
	}\label{fig:8_Attribution_Methods}
\end{figure*}

Feature attribution methods of interpretability cover a broad range of techniques that use various means to show the contribution of input features to the given prediction by the model. The explanations can be shown in various ways, including heatmaps, bar and scatter plots. This class of approaches include interpretability techniques that can be applied in domains that use images, text and tabular data. Some of the most popular attribution methods are gradient-based approaches and model-agnostic methods like SHAP \cite{lundberg2017a} and LIME \cite{ribeiro2016explaining}. Examples of attribution-based interpretability is presented in Figure \ref{fig:8_Attribution_Methods}. The illustration shows two popular input modalities (image and tabular data) and suitable attribution outputs in various forms. Figure \ref{fig:7_Table_PPT1} also shows different categories of interpretability methods and their common characteristics.

\subsection{Gradient-based methods}

Gradient-based interpretability frameworks are a class of feature attribution methods that leverage the gradient computation in neural networks to determine feature importance. These methods attempt to explain the prediction of AI models by highlighting, in the form of a heatmap, regions in an image that are considered to have contributed to the given decision. Some methods in this category (e.g., Score-CAM  \cite{wang2020score}) do not technically use gradient information per se but are nonetheless classified among these methods owing to their strict membership of a broad family of differentiable approaches. Popular gradient-based interpretability methods are described in the next subsections. A broad summary of the  features of the main classes of approaches is presented in Figure \ref{fig:11_Summary_Gradient_Methods}.

\subsubsection{Gradient saliency}

Saliency visualization methods achieve interpretability by providing simple visualization cues that highlight relevant regions of the sample of interest. Gradient saliency, introduced by Simonyan \textit{et al}.  \cite{simonyan2013deep} for  explaining image classification decisions, generates explanations by constructing visual saliency maps from computed gradients of a neural network. The method is one of the earliest examples of gradient-based frameworks that have remain relevant to date. The technique uses the partial derivative or gradient (through back-propagation) of the final prediction score with respect to the input image. These gradients are used to construct the saliency or attribution map that highlights the relative importance of each pixel in the input. The approach is relatively simple and computationally efficient.

 Despite these strengths, vanilla gradients violate an important sensitivity axiom \cite{sundararajan2017axiomatic} causing the saliency maps to show excessive noise. The sensitivity axiom requires that an attribution method should highlight a feature as relevant if a model produces different predictions for a given input and a baseline which differ only by this feature. This problem is partly caused by the gradient saturation problem, which originates from the backpropagation of gradients through several layers of a neural network with non-linear compression functions. Specifically, saturation can result from the computations at the activation functions like sigmoid or the negative input region of the Rectified Linear Unit (ReLU)  \cite{nair2010rectified}. The problem causes some of the gradients with respect to the input to \textit{diminish} and subsequently results in these smaller gradients being overshadowed and obfuscated by larger ones which may not be relevant to the classification task, and thus, manifesting as noise in the saliency map.

\subsubsection{Enhanced Gradient-based Interpretability}

\textbf{Gradient $\times$ input}: In order to improve the invariance of common gradient-based methods to input shifts, the gradient $\times$ input technique  \cite{kindermans2019the},  \cite{shrikumar2016} utilizes the product of the gradient and input for relevancy attribution. This technique prevents the tendency of the computed saliency scores to be undesirably altered by inconsequential changes (such as small constant shifts) in the input which do not usually affect the prediction itself.  

\textbf{  SmoothGrad}: In SmoothGrad  \cite{smilkov2017smoothgrad}, Smilkov et al. propose to reduce visual noise associated with gradients by first generating a new set of images through the addition of noise to the original image and then averaging the saliency maps for all images to obtain the final heatmap. The intuition behind the technique is that the computed derivative is characterized by noisy fluctuations, but averaging over slightly different maps helps to cancel out this noise. However, SmoothGrad requires a couple of passes through the DNN model, resulting in extra computational overhead. Other modifications of this basic approach have been proposed. For instance, NoiseGrad  \cite{bykov2022noisegrad} applies noise to the network weights instead of the input space  and VarGrad  \cite{adebayo2018local} uses the variance of the  attribution output instead of the average.

\subsubsection{Backpropagation-based methods}

Backpropagation-based methods generate relevancy maps by modifying the backpropagation algorithm (specifically, at the ReLU activation function) and using this new rule to backpropagate the final output through successive layers to the input in order to obtain the importance of each input feature to the prediction. 

\begin{figure}[H]
	\vspace {-1mm}
	\centering
	\includegraphics[width=1.0 \linewidth]{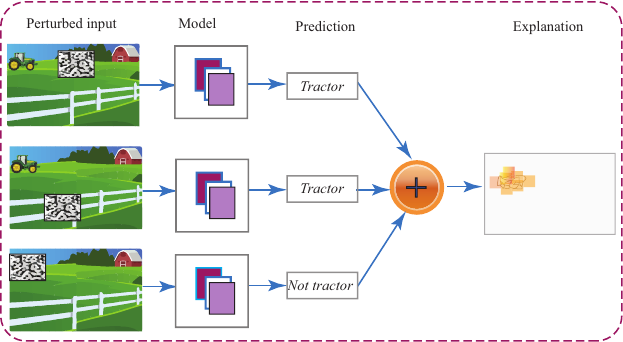} \vspace {-3mm}
	\caption{An illustration of interpretability by occlusion: By systematically occluding parts of the image, the final attribution can detect the region which most influences the prediction of the tractor.  
	}\label{fig:9_Perturbation}
\end{figure}

\textbf{Deconvolutional Networks}: The DeConvolutional Network or DeConvNet  \cite{zeiler2013}framework , proposed by Zeiler and Fergus, selectively projects activations (i.e., all but the target activation is set to zero) backwards to reconstruct the input so as to determine which pixels contributed to the given activations. To achieve this objective, a forward pass is first required to identify so-called \textit{switches}, or positions of maximum values in the max pooling regions. The switches help to place features in their correct locations in the reconstructed signals when they are passed through ReLU activation functions to obtain positive-only outputs. One fundamental difference between DeConvNets and gradient saliency   \cite{simonyan2013deep} lies in how they handle information at  ReLUs. In gradient saliency, during the backpropagation of scores, the gradient passing through a ReLU in the backward pass is masked out (i.e., set to zero) if the information flowing through the same ReLU in the forward pass is negative. However, during the backpropagation of scores through DeConvNets, the output from a ReLU is zero only when the current input is negative (i.e., regardless of the value in the forward pass). The technique is also sometimes referred to as \textit{occlusion} or perturbation-based method because it allows attribution to be computed by systematically occluding (replacing appropriate patches with noise, zero or random values) different patches of the input image and evaluating the resulting reduction in prediction probability score. When attributing feature importance, the occluded patches that result in significant decrease in the class score are considered influential and are thus, shown in the visualization maps. An illustration of interpretability by the input occlusion method is depicted in Figure \ref{fig:9_Perturbation}.

\textbf{Guided Backpropagation}: With guided Backpropagation  \cite{springenberg2014striving}, Springenberg  et al. replace max pooling layers with large-stride convolutions to achieve dimensionality reduction. This modification obviates the need for the forward pass step to estimate switch variables, leading to better reconstruction results and improved visualization quality of the final heatmap. Another fundamental difference is in the propagation rule used by these methods. Specifically, while DeConvNet uses the back-projecting RELU to set all negative entries of the top gradients to zero, Guided Backpropagation combines the information processing rules at ReLUs adopted by Gradient Saliency and DeConvNets. That is, the method masks out entries to ReLUs for which either the gradients in the backward pass or the data in the forward pass are negative. By this design, Guided Backpropagation achieves higher interpretability quality with sharper visualizations. 

 The main strengths of DeConvNet and Guided Backpropagation techniques lie in their relatively high-resolution heatmaps. However, these methods are not class discriminative. Class-discriminative behavior (illustrated in Figure \ref{fig:10_SaliencyMaps}) describes the ability of interpretability methods to correctly highlight target classes in images containing multiple classes. Moreover, owing to the masking out of negative entries, neither method is suitable for capturing features that contribute negatively to the prediction. Furthermore, like gradient saliency, these backpropagation-based methods do not satisfy the sensitivity axiom owing to the propagation of gradients through deep neural networks. This problem ultimately leads to degraded performance.

\subsubsection{Activation propagation-based methods}

This class of methods compute attributions by assuming that the activation of a given neuron in the final layer of the network represents the relevance of the neuron to the prediction. These neuron activations are then propagated backward to the input. The various frameworks (e.g., (LRP)  \cite{bach2015on}, DTD   \cite{montavon2017explaining}, and  DeepLIFT  \cite{shrikumar2017learning}) differ by their relevancy propagation details. In general, the approaches work by decomposing (e.g., through Taylor decomposition) successive outputs into contributions of various units from previous layers. The technique helps to address some of the challenges associated with using gradients, such as saturation. Like other attribution-based methods, the final output is a heatmap that highlights relevant parts of the input.  

\textbf{   Layer-wise Relevance Propagation}: Layer-wise Relevance Propagation (LRP)  \cite{bach2015on} seeks to quantify the relevance or contribution of each pixel of an input image to the final prediction of a classifier by generating pixel-wise heatmaps based on the backpropagation of scores from the output.   LRP can determine the strength of the contribution of each unit in any layer of a network by computing and backpropagating the attribution scores from the output layer to the given layer using specially designed propagation rules.  

 Layer-wise relevance propagation is also considered a decomposition-based interpretability method as the relevance scores are iteratively decomposed and redistributed from the current to the next layer (i.e., when traversing the network in the direction of the shallower layers). Bach et al. \cite{bach2015on}also show that  LRP can be approximated by Taylor decomposition.
 
 \begin{figure}[H]
 	\vspace {-1mm}
 	\centering
 	\includegraphics[width=1.0 \linewidth]{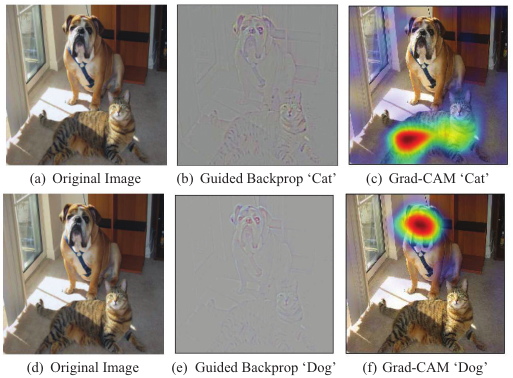} \vspace {-3mm}
 	\caption{Illustration of some important properties of saliency maps. The input image is shown in (a) and (d). The outputs illustrate high resolution heatmaps from Guided Backpropagation (GB) \cite{springenberg2014striving} (i.e., (b) and (e)) and low-resolution maps from Grad-CAM in (c) and (f). Also, class-discriminative behavior is illustrated by Grad-CAM \cite{selvaraju2017grad} which can highlight either a cat or dog. Partial activation problem is also illustrated in (f) where only the most discriminative region (in this case the head of the dog) is highlighted. 
 	}\label{fig:10_SaliencyMaps}
 \end{figure}

 Although LRP was originally proposed to enhance the interpretability of image classification decisions, the method has since been extended to solve interpretability problems in domains that use other input data types including audio (e.g., \cite{becker2018audiomnist}), textual (e.g., \cite{arras2017what}) and tabular (e.g.,  \cite{ullah2021explaining}) data.

By virtue of its decomposition approach, LRP not only determines the magnitude of the contribution of features but can also show its direction, where positive attribution values indicate feature importance for the given prediction while negative values provide evidence contradicting the existence of the target. The method is relatively computationally efficient and is supported by principled theoretical procedures. Despite these strengths, the method has some drawbacks, notably, its limited flexibility owing to the need to design propagation rules for different types of layers while the class-discriminativeness of the method has also been questioned. Consequently, more recent techniques like class Contrastive LRP LRP \cite{gu2019understanding} and  Softmax-Gradient LRP (SGLRP)  \cite{iwana2019explaining} enable class-discriminative explanations.  Contrastive LRP constructs class-discriminative explanations by comparing the relevance of the target class with the non-target class. SGLRP utilizes the gradient of the softmax as the relevance that is backpropagated to the input. Both methods obtain their final attribution maps by subtracting the relevance of non-target classes. Class-discriminative property is illustrated in Figure \ref{fig:10_SaliencyMaps}.


\begin{figure*}[!htb]
	\vspace {1mm}
	\centering
	\includegraphics[width=1.0 \linewidth]{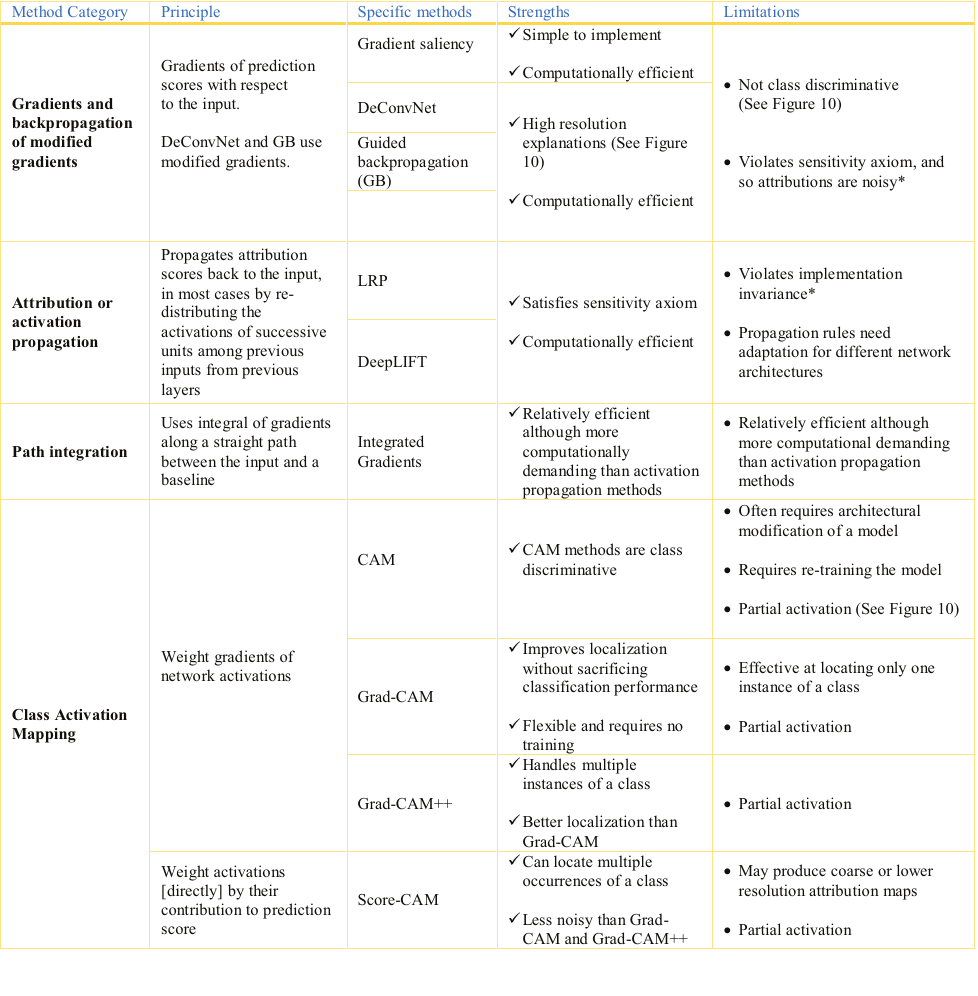} \vspace {-3mm}
	\caption{Summary of common gradient-based methods. $^{\ast}$The sensitivity axiom and implementation invariance are illustrated by Sundararajan et al. \cite{sundararajan2017axiomatic}.
	}\label{fig:11_Summary_Gradient_Methods}
\end{figure*}


\textbf{Deep Taylor Decomposition:} Deep Taylor Decomposition (DTD) or DeepTaylor proposed by Montavon et al.  \cite{montavon2017explaining} applies Taylor decomposition to network activations obtained in the forward  pass. The method decomposes activation values starting from the classification output, and backpropagates through successive layers down to the input, while at each point distributes the values into the contributions of constituent units from the next layer that feeds the current one.

\textbf{ DeepLIFT}: To overcome saturation and the loss of information due to discontinuities in the propagation of gradients, DeepLIFT  \cite{shrikumar2017learning} avoids computing relevance scores directly  from the backpropagation of gradients and instead uses activation differences. That is, the relevance score of each neuron \textit{is} the difference between the activation of the neuron in response to the original input \textit{and} its activation to some reference input. The method propagates this difference backward to obtain the contribution of input features to the outputs at specified locations where the last layer activations were recorded. intuition behind DeepLIFT is that it is more informative to understand how an output changes as the input deviates from a baseline (usually zero-valued), than to quantify the gradient at a point, which measures how an output changes as the input changes infinitesimally in the neighborhood of the point. By using activation differences, DeepLIFT can propagate meaningful information even in situations where the true gradient is zero.

\subsubsection{Integrated Gradient-based methods}

\textbf{  Integrated Gradients}: To mitigate gradient saturation, reduce noise and improve the resulting attribution map, integrated gradients (IG) \cite{sundararajan2017axiomatic} first introduced by  Sundararajan et al., computes integral of the average gradients along a continuous path between the network’s input and a selected reference or baseline (often, zero-valued input). Owing to its formulation Integrated Gradients possesses desirable properties. In fact, the original idea behind IG is motivated by the need for attributions to satisfy two critical axioms – sensitivity to changes in relevant features; and implementation invariance, which previous gradient-based methods could not satisfy. Other methods (e.g., DeepLIFT and LRP) try to satisfy sensitivity axiom requirement by trying to compute discrete gradients with the help of a baseline, rather than computing instantaneous gradients with respect to the input.

 Unfortunately, as noted by Sundararajan the chain rule cannot be applied directly to discrete gradients. Therefore, the activation propagation-based approach employs custom (network-specific) propagation rules that cause the method to violate \textit{implementation invariance} requirement. A given method is implementation invariant if its attributions for functionally equivalent neural networks are always identical. Meanwhile, two networks are said to be functionally equivalent if different implementations of these networks produce identical outputs for all possible inputs. Arras et al. \cite{arras2017explaining} design a propagation rule to work with recurrent neural network (RNN) models which have  multiplicative connections between layers. Specifically, the authors address the propagation of relevance scores through Long ShortTerm Memory (LSTM) and Gated Recurrent Units (GRUs).

  \textbf{Expected gradients:} Expected Gradients \cite{sturmfels2020visualizing},  \cite{erionlearning} have been proposed to overcome a number of inherent deficiencies of approaches based on Integrated Gradients. \textbf{ }Most notably\textbf{, }In Integrated Gradients, the baseline is assumed to be uninformative–i.e., it does not contain information relevant to the target. However, in the image domain, the default all-zero baseline may actually represent valid data points within the target, specifically, black regions of the target object. In this situation, Integrated Gradients fails to highlight black regions as contributing to the class prediction even when they do.  Although different choices of baseline have been proposed (see   \cite{sundararajan2017axiomatic}), they also face a similar problem – IG does not correctly attribute regions that similar to the baseline. To circumvent the issue of the baseline which may overlap with the target class, recent studies  \cite{sturmfels2020visualizing},  \cite{erionlearning}propose to\textbf{ } obviate the need for selecting an arbitrary baseline by integrating over the training data distribution. However, owing to the complexity of computing integrals over entire datasets, the approach, aptly referred to as expected gradients, relies on approximating these integrals by their expectations.

\textbf{Guided Integrated Gradients:} Kapishnikov et al. \textbf{ } \cite{kapishnikov2021guided} show that in Integrated Gradients, noisy saliency maps can result from the accumulation of \textbf{ }noise along the path used for integration. As a result of this serious shortcoming, the authors devise an alternative, noise-robust gradient-based technique, Guided Integrated Gradients, to compute saliency maps. In Guided Integrated Gradients, the authors propose to use an adaptive path which depends not only on the input but also on the model. This path replaces the straight line adopted in the original study.  Guided IG achieves higher quality heatmaps with less noise compared with IG.

\subsubsection{Class Activation Mapping techniques}

The idea behind this popular family of interpretability methods is that the pooled features at the last layers of convolutional neural networks (CNNs) carry important localization information about the target. However, this information is lost at the classification output owing to the use of fully-connected layers for class prediction. Various techniques have therefore been proposed for recovering the localization information to provide explanation for classification decisions.

\textbf{Class Activation Mapping (CAM)}:  The Class Activation Map (CAM)   \cite{zhou2016learning} technique  was originally proposed by Zhou et al   \cite{zhou2016learning}, who leverage the localization ability of the global average pooling (GAP) layer in CNNs to perform classification and sparse localization without requiring bounding box regression. Thus, in CAM, the authors replace the last fully-connected layer with the GAP layer. This layer computes the weighted sum of class-discriminative features (or class activation map) of the input image from the final convolution layer. The class activation map is computed across spatial dimensions to capture the importance of the activations at different spatial locations.

\textbf{  Grad-CAM:} CAM usually requires architectural modifications to existing CNN models (specifically, the introduction of the GAP layer), a procedure which can potentially harm performance. To overcome this drawback, Selvaraju et al. \cite{selvaraju2017grad} propose Gradient-weighted Class Activation Mapping (Grad-CAM), which computes activation maps from the gradients (with respect to the input) of the predicted class. The method is capable of producing visual explanations from existing CNN models without further modifying nor needing a re-training of the model. This flexibility implies that the technique can readily be applied across many different CNN architectures. The technique utilizes the global weighted average of gradients of the target class at the final convolutional layer to construct class-discriminative attribution. To obtain fine-grained pixel-level attributions which are also class-discriminative, the study in Grad-CAM also designs Guided Grad-CAM by upsampling and fusing saliency maps of Grad-CAM with the high-resolution outputs of Guided Backpropagation. High-resolution and class-discriminative properties of visualization maps are shown in Figure \ref{fig:10_SaliencyMaps}.

\textbf{ Grad-CAM++:} Besides the increased flexibility, Grad-CAM achieves superior results to CAM. However, neither method is capable of highlighting multiple instances of a target object in a single input frame. Grad-CAM++  \cite{chattopadhay2018grad} addresses this limitation by utilizing the weighted average (rather than the global average employed by Gad-CAM) of positive partial derivatives of learned features to obtain heatmaps for multiple instances of a given class in an image. Grad-CAM++ attributions also have better coverage of relevant pixels of target objects.

\textbf{ Layer-CAM}: While earlier CAM versions like Grad-CAM and Grad-CAM++ typically compute class activation maps from the last convolutional layer of CNN models, Jiang et al. \cite{jiang2021layercam} propose Layer-CAM, which combines class activation maps from shallower to deeper layers. The intuition is based on the fact that the deeper layers are  only capable of capturing course information whereas shallower layers encode high spatial resolution or fine-grained details about object location. Thus, by combining activations across shallower and deeper layers, Layer-CAM leverages both global information and more accurate object details to achieve better coverage and improved interpretability. However, the visual explanations inherit accumulated noise by combining gradient information from multiple layers.

\textbf{Score-CAM}: Owing to the well-known problems with gradients, such as vanishing gradients and noise, CAM methods that rely on computing gradients suffer from impaired visualization outputs. Score-CAM \cite{wang2020score} constructs its heatmap by aggregating activation maps (i.e., across designated channels) which are derived by weighting final confidence scores on the predicted classes. The activations are first upsampled to match the dimensions of the input, followed by normalization to enhance class discrimination capability. Subsequently, the normalized activations serve as masks to extract informative regions of the input image. The relevancy weights are the post-softmax outputs after the normalization operation. Since the weights directly correspond to the prediction score on the target class, Score-CAM can independently localize each target object in an image with high prediction confidence scores. However, like other class activation mapping techniques, the visualizations produced by the method do not cover all target regions. Furthermore, Score-CAM has high computational overhead and is prone to producing coarse maps owing to the upsampling of activation maps. This problem is addressed by more recent methods which utilize various techniques such as image augmentation-based activation maps in Augmented Score-CAMS \cite{ibrahim2022augmented}; smoothing operations in SS-CAM  \cite{wang2020ss}; and fusion of Grad-CAM++ and ScoreCAM by Soomro et al. \cite{soomro2024gradscorecam}. Additionally, the computational demand of Score-CAM is also addressed in recent studies (e.g., FIMF score-CAM \cite{li2023fimf}.

\subsection{Faithfulness of gradients and CAM methods}

\begin{figure}[H]
	\vspace {-1mm}
	\centering
	\includegraphics[width=1.0 \linewidth]{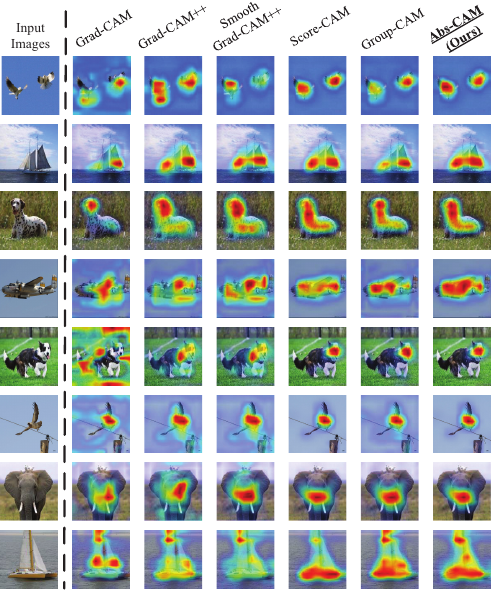} \vspace {-5mm}
	\caption{Summary of common gradient-based methods \cite{zeng2023abs}.  
	}\label{fig:12_HeatMaps}
\end{figure}

Visualization results in the form of visualization maps like those illustrated in Figure \ref{fig:12_HeatMaps} provide a basis for comparing the performance of different attribution methods. But this assessment criterion is highly subjective and may lead to wrong conclusions about the quality of explanations.  To mitigate this limitation, other performance metrics such as mean Intersection over Union (mIoU) allow to quantitatively evaluate the localization accuracy of heatmaps with respect to a ground truth segmentation mask. To achieve this objective, some authors (e.g.,  \cite{jiang2021layercam}, \cite{shi2020zoom} treat class activation mapping as a weakly-supervised semantic segmentation problem and therefore seek to evaluate the accuracy of saliency maps using object localization metrics such as mIOU to measure how well the heatmap coincides with the ground-truth segmentation mask.  However, since the primary use of heatmaps is to explain the reason behind a classification decision, localization accuracy does not suffice as this metric does not measure the underlying factors that influence the generated heatmap. Faithfulness metrics overcome this limitation. The faithfulness of an interpretability method measures the extent to which the explanations reflect the true rationale of a model’s decision. Metrics designed to measure this property focus on assessing the impact of perturbing the highlighted features on the prediction accuracy.

\subsubsection{Faithfulness of saliency maps}

The faithfulness of heatmaps generated by different interpretability methods can be quantitatively evaluated by ablating features in the input space. One of the most widely employed metrics based on this principle relies on removing a subset of features from the input and recording the resulting classification accuracy. For this metric, two main protocols  \cite{samek2017evaluating} are commonly used: 1) Most Relevant First (MoRF) is based on removing features in order of importance (from highest to lowest) while quantifying the corresponding decrease in accuracy. The idea is that, if the attribution method is faithful, removing features that are important to the model’s decision should cause a significant reduction in the prediction. 2) Least Relevant First (LeRF) follows a similar logic except that input features are removed in the reverse order of importance. 


\begin{figure*}[!htb]
	\vspace {1mm}
	\centering
	\includegraphics[width=1.0 \linewidth]{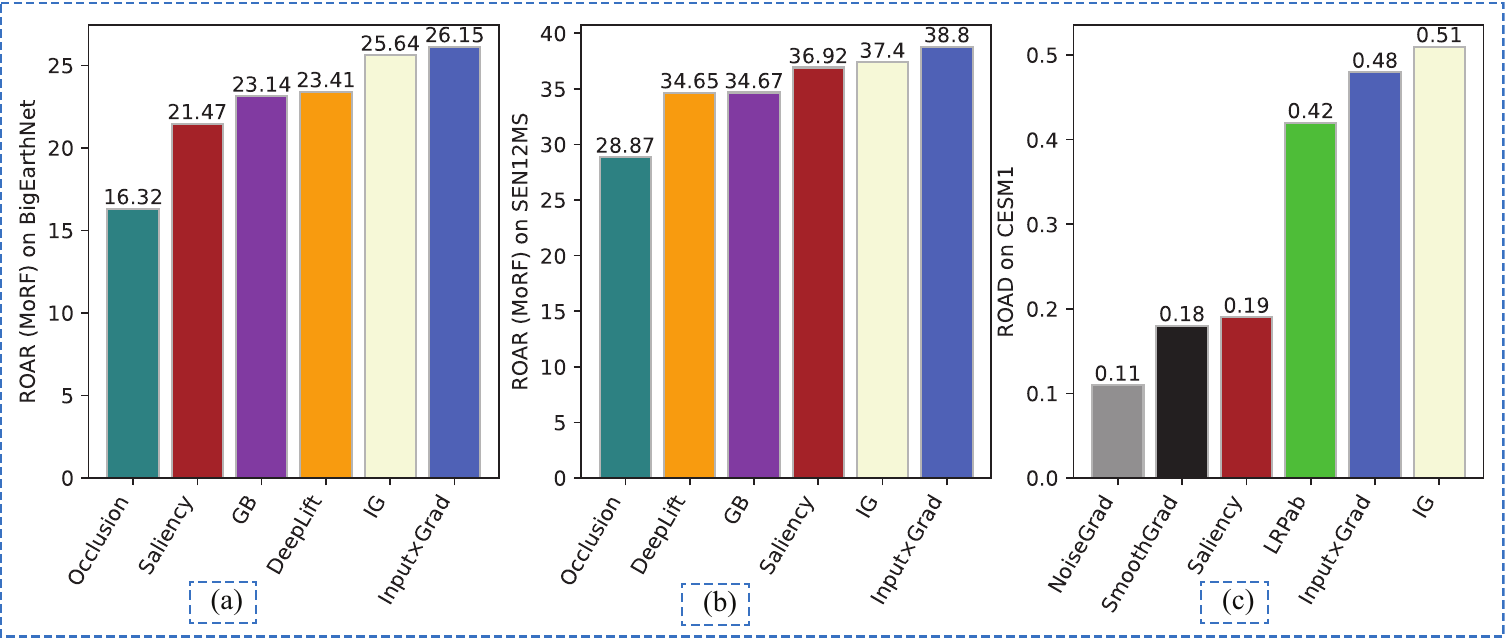} \vspace {-3mm}
	\caption{Ranking faithfulness of popular gradient-based attribution methods under three settings: (a) and (b) ROAR-MoRF scores from \cite{hurrell2013the}, \cite{kakogeorgiou2021evaluating} on BigEarthNet \cite{sumbul2020bigearthnet} and (c) ROAD score using custom network \cite{bommer2024finding} and CESM1 data \cite{hurrell2013the}. The results show that although gradient-based methods cannot be consistently   by popular faithfulness metrics, Occlusion generally performs high whiles IG and input×gradient do not.       
	}\label{fig:13_Gradient_Attribution}
\end{figure*}


 Despite the popularity of this metric, Hooker et al.  \cite{hooker2019a} argue that feature removal alone cannot account for the decline in prediction accuracy as this modification causes a shift in the distribution of the test data relative to the training samples, a situation which reduces prediction accuracy. The authors therefore propose a new method, RemOve And Retrain (ROAR), which addresses this distribution shift by re-training the model on the modified samples after feature removal. Subsequent studies (e.g.,  \cite{tomsett2020sanity},  \cite{rong2022a}) reveal that most training-free methods, as well as the ROAR framework produces inconsistent rankings of attribution methods under the MoRF and LeRF protocols, where some methods can achieve high faithfulness scores when evaluated in a particular order but tend to record low faithfulness when tested in the opposite order. Rong et al. \cite{lipton2016}) introduce ROAD - RemOve And Debias – that leverages the noisy linear imputation technique to overcome information leakage that undermines consistency ROAR. The method thus improves consistency and efficiency through time savings on re-training.

    Generally, evaluating the faithfulness of attribution methods through feature ablation remains a serious challenge as there are practical limitations of the method. For instance, since the input dimension is fixed for image data, the ``removed" pixels are actually replaced by a patch of zero (i.e., a black patch), mean (i.e., an arbitrary batch with edges) or random (i.e., some noisy patch) values or sometimes values sampled from a predefined distribution. 

Popular quantitative evaluation metrics for attribution methods, including ROAD, show a great deal of inconsistency across different settings or configurations (e.g., different models, datasets, or training epochs). Despite this limitation, some degree of uniformity can be observed for the results produced by these metrics. For example, in the representative studies presented in Figure \ref{fig:13_Gradient_Attribution}, methods like Integrated Gradients and \textit{input $\times$ gradient}, where the input is multiplied by the gradient to obtain the attribution scores, frequently achieve low faithfulness. This behavior may be caused by the underlying computations that generate the final heat map. 

For instance, multiplication by input pixels may emphasize individual features which may not capture rich semantic information relevant for recognizing the target class.  Additionally, pixel values of zero in the input can eventually diminish corresponding weights in the heatmap.  For these reasons, some regions in the heatmap which may not be deemed so important could actually be crucial in in the prediction task while seemingly unimportant regions could be more informative. Thus, removing these pixels may not produce desired change in accuracy. In contrast, as shown in Figure \ref{fig:13_Gradient_Attribution}, methods like DeconvNet (Occlusion), SmoothGrad and NoiseGrad frequently achieve relatively higher scores on faithfulness metrics. DeconvNet produces high resolution heatmaps which strongly highlight fine-grained features and less noise. And, since these features mostly capture high-level semantic information useful for classification, removing pixels that correspond to edges should indeed result in a significant drop in accuracy. Hence, the low ROAR and ROAD scores are expected. Meanwhile, averaging saliency maps of the original image and its perturbed replicas provides smoothing effect in SmoothGrad and NoiseGrad, resulting in sharp maps and reduced noise or artifacts. This may cause output maps to focus on more discriminative features whose removal can harm accuracy.


\begin{figure*}[!htb]
	\vspace {1mm}
	\centering
	\includegraphics[width=1.0 \linewidth]{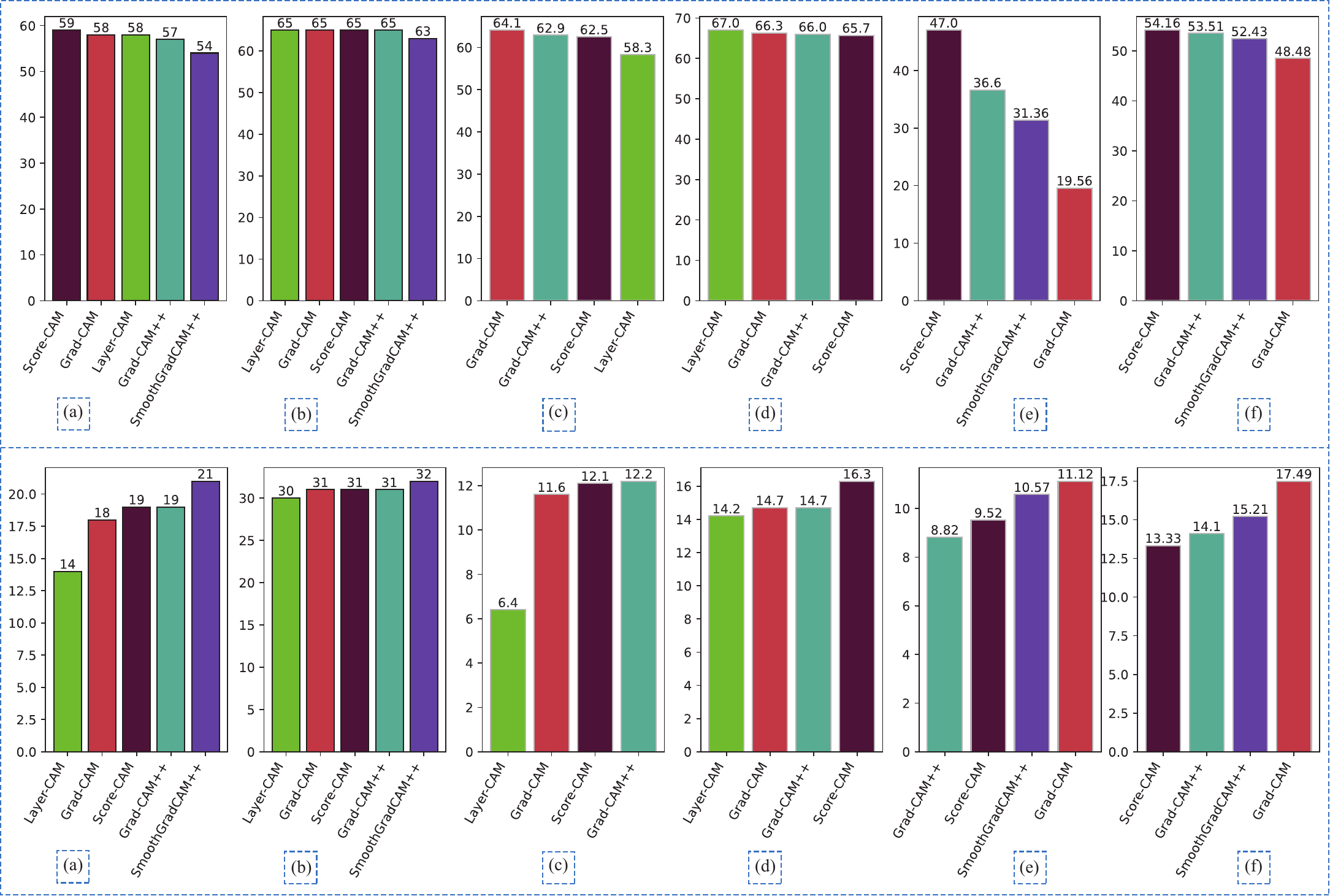} \vspace {-3mm}
	\caption{Ranking of CAM methods by the insertion and deletion metrics computed from various studies across different settings: (a) and (b) Poly-CAM \cite{englebert2024poly}; (c) and (d) Opti-CAM \cite{zhang2024opti}; (e) and (f) Fake-CAM \cite{poppi2021revisiting}. The top row shows plots for the insertion metric while bottom row is for deletion. Note that for insertion higher score indicates better performance while lower is better for deletion. The Figure shows two plots for each study where the first uses a VGG architecture while the second uses ResNet.     
	}\label{fig:14_CAM_Methods}
\end{figure*}


\subsubsection{Faithfulness of class activation mapping techniques}

Many recent studies (e.g., Ablation-CAM  \cite{desai2020ablation}, CAPE  \cite{chowdhury2024cape}, GT-CAM  \cite{ligt}) frequently evaluate the quality of heatmaps of class activation maps by analyzing how changes in the relevant pixels affect the prediction. That is, these metrics measure the faithfulness of attribution methods. The commonly used metrics include.  1) Average Drop (AD) and Average Increase (AI)  \cite{chattopadhay2018grad}; 2) Insertion and Deletion  \cite{petsiuk2018rise}.

\textbf{     Average Drop (AD): }This metric computes the average drop (in terms of percentage) in prediction confidence when the network is given only the heatmap as input, rather than the original image. The intuition here is that masking out portions of the image that do not explain the prediction should cause a smaller drop in the class probability score. \textbf{Average Increase (A}I): AI computes, over the entire dataset, the number of times the prediction confidence increases when irrelevant regions are masked out. 

\textbf{Insertion and Deletion}  \cite{petsiuk2018rise}: Insertion computes faithfulness as the increase in class probability score as relevant pixels within the explanation heatmap are progressively added to a baseline (i.e., uninformative reference) image while d\textbf{eletion} measures the drop in the class score as relevant pixels are removed and their positions are filled with pixels from a baseline image. Both insertion and deletion are expressed as the total Area Under the Curve (AUC) of the prediction score. Higher values of insertion indicate higher faithfulness, while for deletion, lower values signify higher faithfulness. Figure \ref{fig:14_CAM_Methods} illustrates that heatmaps generated by different CAM methods under different settings (network architectures and configurations) do not show consistency in faithfulness. However, Score-CAM achieves higher faithfulness than most other methods. This may be mainly to the fact that the method avoids the use of gradients and hence does not suffer many of its effect like saturation.   

\subsubsection{The challenge of faithfulness metrics}

The results presented in Sections 3.3.1 and 3.3.2 illustrate that saliency maps. Moreover, the results of different evaluation metrics not usually agree, as shown by the difference between the ROAR, ROAD, insertion and deletion rankings in Figures 12 and 13. The inconsistency in faithfulness is partly a limitation of the faithfulness metrics. In \cite{poppi2021revisiting}, Poppi et al. show that a bizarrely incorrect attribution, called Fake-CAM, can outperform state-of-the-art methods on existing metrics. The authors propose a new metric called \textit{maximum coherency} that correctly returns poor score for Fake-CAM. Similarly, Zhang et al.  \cite{zhang2024opti} introduce Average Gain (AG), as an alternative to (AI) and (AD). As desired, AG which gives low performance for Fake-CAM.  However, these methods still do not address the underlying weakness of faithfulness evaluation, which is caused by a fundamental limitation of the underlying principle: specifically, samples generated from perturbed inputs may be of a different distribution from the original training images. Yet, it is well known that CNNs do not generalize well outside their training data distributions.  Furthermore, Nie et al.  \cite{nie2018a} show that even when an adversarial attack causes CNNs to wrongly mis-assign the class label (implying a change in the intermediate network activations), the output visualizations computed with Guided Backpropagation remain unchanged. These observations show that predictions on perturbed images are not guaranteed to elicit correct model behavior. Consequently, deletion-insertion, AI-AD, and improved faithfulness metrics that adapt these methods fail to capture the true reasoning of underlying model. Moreover, an influential study \cite{adebayo2018sanity} conducts ``sanity checks" on state-of-the-art gradient-based methods and finds that many techniques produce saliency visualizations that are insensitive to data and model parameter randomization. This means that the attribution methods are neither dependent on the learned features of the model nor on the relationship between training examples and corresponding labels. These observations underscore the inconsistency of faithfulness metrics for attribution methods.

\subsection{Model-agnostic methods}

Model-agnostic interpretability methods are incorporated into machine learning frameworks after training. These methods derive their explanations to a model’s behavior by utilizing only input data and final prediction outcomes. Unlike other post-hoc interpretability methods, model-agnostic interpretability techniques are independent of the underlying model architecture. For example, the technicalities in applying a model-agnostic method to explain a neural network model that predicts credit score does not differ significantly from what it takes to implement the same method for a random forest classifier on the same task. On the other hand, the implementation of a Grad-CAM or DeepLIFT explainability framework for a neural network is completely different from achieving this for a random forest model.

\subsubsection{Shapley additive explanations (SHAP)}

The most influential method among the perturbation-based approaches is the SHapley Additive explanations (SHAP)  \cite{lundberg2017a} framework . SHAP learns the relevance of different features and the model decision in a game-theoretic setting where the goal is to determine the strength of the contribution through the Shapley value – a measure of the importance of a given feature to the output or outcome. In the game, SHAP treats the given features as players and the output of the black-box model as the payoff. The idea of Shapley values is to score input variables using sampled subsets or coalitions of features based on their relative importance to the outcome. This scoring is carried for various coalition of features, and the weights or contributions of features are determined by computing the outputs and observing the difference in prediction obtained with and without the given feature.

KernelSHAP is a model-agnostic framework, proposed in the original SHAP study  \cite{lundberg2017a}, which utilizes approximation method based on sampling subsets of available features. Frameworks like KernelSHAP that approximate Shapley values by sampling subsets of features can be considered perturbation-based approaches, and for some data types (e.g., images), are also often referred to as occlusion because of how feature perturbation is implemented. The method has a number of desirable properties, including consistency (i.e., always converges when a sufficiently large number of samples are used) and unbiasedness. However, Covert and Lee  \cite{covert2020improving} observe that KernelSHAP compromises bias in order to achieve low variance. Consequently, the authors introduce a technique to reduce variance without incurring bias penalty.

  Also, the computational overhead when applied to large-scale deep learning models is prohibitive. Owing to this drawback, several techniques have been proposed to facilitate efficient computation of Shapley values. Among them are those that retain the model-agnostic characteristic of SHAP.In addition, model-specific versions are also available for accurate and computational tractable realization of SHAP.

In general, model-agnostic implementations employ approximation techniques that avoid evaluating the model with all possible subsets of input features. Some methods (e.g., SHAP@k  \cite{kariyappa2024shapk} achieve computational efficiency by  utilizing appropriate algorithms to select a limited set of features for evaluation. Other methods of speeding up model-agnostic SHAP frameworks include FASTSHAP  \cite{jethani2021fastshap}, which employs a separate machine learning model specifically trained to predict SHAP scores. A comprehensive review of these approaches is presented by Chen et al.  \cite{chen2023algorithms}.

   Among the model-specific techniques are the methods such as: LinearSHAP  \cite{chen2006true} designed to explain the predictions of  linear models; TreeSHAP  \cite{lundberg2020from} for tree-based models like decision trees; Graphsvx  \cite{duval2021graphsvx} for graph neural networks, and HarsanyiNet \cite{chen2023harsanyinet} for computing attribution scores in a single forward pass for deep convolutional neural networks. Additionally, the method proposed by Covert et al.  \cite{covert2022learning} for vision transformers (ViTs) is based on reviewing individual predictions of a vision transformer on subsets of image patches. Covert et al. \cite{covert2022learning} employ a surrogate model to learn SHAP scores by masking attention values for missing image features. Model-specific techniques achieve a level of efficiency by making simplifying assumptions that rely on knowledge of the underlying black-box model’s type. For this reason, these approaches can be applied only to the class of models they are designed for. Furthermore, there are universal also state-of-the-art methods aimed at speeding up the computation of Shapley values that are not restricted to specific model types. In general, these techniques (e.g., FastSHAP  \cite{jethani2021fastshap}, SHAP@k \cite{kariyappa2024shapk} can be applied to a range of machine learning  frameworks. 

   Besides their computational complexity, SHAP methods are also criticized for their feature sensitivity. Moreover, recent methods highlight severe faithfulness issues of SHAP explanations \cite{huang2024on}. Another present-day challenge SHAP is the ability to handle feature interactions.  Some recent works (e.g., Faith-SHAP  \cite{tsai2023faith}, SHAP-IQ  \cite{fumagalli2023shap}) now seek to efficiently compute Shapley interaction indices, paving the way for faithful and accurate attributions for cases involving feature interactions.

\subsubsection{Locally interpretable model-agnostic explanations (LIME)}

Locally interpretable model-agnostic explanations (LIME)) \cite{ribeiro2016explaining} provide  interpretability by learning the behavior of a black-box model with the aid of an interpretable surrogate model which provides local explanations for the predictions of the original black box mode. For this reason, LIME is also referred to as surrogate-based interpretability framework. The surrogate model is described as \textit{local} because it provides explanations for individual predictions, rather than elucidating the general behavior of the original model. LIME is simultaneously a perturbation-based method because, in its implementation the input is perturbed with various permutations and the effect on the output is used to measure feature importance. The idea is that by observing the effect of various perturbations of the input on the particular prediction, the input most relevant to given prediction can be identified.

LIME has many clear advantages. First, it is model agnostic and flexible to deploy in a wide variety of models. Second, it supports different data types such as text, images tabular data. Consequently, the approach is very popular and has many useful practical applications  \cite{band2023application},  \cite{vimbi2024interpreting},  \cite{gaspar2024explainable}. Despite its strengths, LIME also has important limitations . Notably, the performance degradation that may result from fitting a simpler and potentially less accurate surrogate model to a large complex model. In general, interpretable models cannot usually match the accuracies of complex black box models. Therefore, LIME may suffer from a costly trade-off where accuracy is sacrificed for interpretability. 

Stability and fidelity are also major concerns of LIME-based explanations. Fitting the surrogate requires random sampling of points which may lead to instability and/or infidelity, causing non-deterministic explanations or different behavior from the original black box model. Stability describes the tendency of the attribution method to generate inconsistent explanations for different runs of the model on the same input. Fidelity refers to how the local interpretable model matches the predictions of the original black-box model. Instability or inconsistent explanations cannot engender trust in machine learning systems because when different explanations are provided for a prediction, only one can be the true rationale. Therefore, a model that produces unstable explanations must be producing some false explanations as well. Hence, such explanations cannot always be trusted. Lee et al.  \cite{lee2019developing} note that , in image classification for example, instability of saliency map explanations can arise from the binary manner of assigning attributions: regions are classified positive or negative for the target if the output weights are above or below a fixed threshold. For this reason, a small difference in output weight over successive iterations with the same instance can flip the attribution for a given region. The authors  \cite{lee2019developing} address this problem by  averaging output weights over multiple runs through the model, resulting in consistent saliency maps that eliminate noisy fluctuations. A major contributing factor to the instability of LIME is the random nature of the process involved in generating perturbed samples. Therefore, another line of work aimed at overcoming the stability problem of LIME is focused on improving sample selection techniques. For instance, DLIME  \cite{zafar2019dlime} replaces random perturbation in the conventional LIME approach with clustering-based techniques.  Specifically, DLIME employs hierarchical clustering (HC) to group input data and k-nearest neighbor clustering to select the cluster in which the given instance belongs. G-LIME  \cite{tan2023glime} creates  a local interpretable model around samples generated in the neighborhood of the given input. Most conventional perturbation techniques rely on sampling points in non-local neighborhoods which inevitably leads to a general bias towards the selected reference. This may also result in poor local fidelity. To address these problems, G-LIME ‘s samples are drawn from a local distribution and are independent of fixed reference points. Consequently, the method achieves higher stability and local fidelity. Still, other prominent methods exist for generating samples. For example, ALIME \cite{shankaranarayana2019alime} trains an auto-encoder on the training data to serve as the weighting function for generated samples while S-LIME  \cite{zhou2021s} relies on hypothesis testing to adaptively compute perturbations.

To mitigate the fidelity problem, US-LIME  \cite{saadatfar2024us} leverages uncertainty sampling  to generate high-fidelity training samples close to the instance to be explained and to the decision boundary. A denoising auto-encoder achieves a similar goal in ALIME. Both US-LIME and ALIME excel in promoting local fidelity of the distilled surrogates while achieving better stability compared to conventional baselines. Despite the efforts in this direction to improve LIME frameworks, it is practically difficult to train light-weight surrogates to mimic the behaviors of state-of-the-art large-scale models. Also, dense sampling can significantly increase the computational burden of the LIME framework. Therefore, for satisfactory performance, the LIME framework needs enormous capability from model and data perspective in order to reach the desired predictive accuracy to imitate the original black-box models. However, this requirement essentially leads to a trade-off between explainability and performance as large and complex surrogate models would be less interpretable compared with simpler ones.

\subsubsection{Comparison of SHAP and LIME}

SHAP and LIME share a number of properties. However, they also differ in a number of ways. Table 2 summarizes their most important features.

\begin{table*}[htb!]
	
	\caption{A comparison of SHAP and LIME Methods}
	\label{tab:T2_Comparison_SHAP_and_LIME}
	
\renewcommand{\arraystretch}{1.3}
\begin{adjustbox}{max width=\textwidth}
	
	\scalebox{0.90}{
\begin{tabular}{p{2.53cm}p{4.13cm}p{3.17cm}p{6.66cm}}
\hline
\multicolumn{1}{|p{2.53cm}}{\textbf{Attribute}} & 
\multicolumn{1}{|p{4.13cm}}{\textbf{SHAP}} & 
\multicolumn{1}{|p{3.17cm}}{\textbf{LIME}} & 
\multicolumn{1}{|p{6.66cm}|}{\textbf{Remark}} \\ 
\hline
\multicolumn{1}{|p{2.53cm}}{Principal concept} & 
\multicolumn{1}{|p{4.13cm}}{Game theoretic principle with feature sampling to reduce computational cost.} & 
\multicolumn{1}{|p{3.17cm}}{Feature perturbation with interpretable surrogate as proxy for the black-box model.} & 
\multicolumn{1}{|p{6.66cm}|}{Both SHAP and LIME are \textit{post-hoc} methods as they are applied to a pre-trained black-box models.} \\ 
\hline
\multicolumn{1}{|p{2.53cm}}{Scope} & 
\multicolumn{1}{|p{4.13cm}}{Both local and global} & 
\multicolumn{1}{|p{3.17cm}}{Local} & 
\multicolumn{1}{|p{6.66cm}|}{SHAP methods can aggregate information for the entire data to obtain produce explanation.} \\ 
\hline
\multicolumn{1}{|p{2.53cm}}{Applicability} & 
\multicolumn{1}{|p{4.13cm}}{Model-agnostic} & 
\multicolumn{1}{|p{3.17cm}}{Model-agnostic} & 
\multicolumn{1}{|p{6.66cm}|}{Some SHAP methods (e.g., TreeSHAP 
	\cite{lundberg2020local}, HarsanyiNet \cite{chen2023harsanyinet}) are model-specific.} \\ 
\hline
\multicolumn{1}{|p{2.53cm}}{Stability} & 
\multicolumn{1}{|p{4.13cm}}{High} & 
\multicolumn{1}{|p{3.17cm}}{Low} & 
\multicolumn{1}{|p{6.66cm}|}{LIME relies on sampling inputs in a manner that causes instability. } \\ 
\hline
\multicolumn{1}{|p{2.53cm}}{Computational overhead} & 
\multicolumn{1}{|p{4.13cm}}{Extremely high/High} & 
\multicolumn{1}{|p{3.17cm}}{Moderate} & 
\multicolumn{1}{|p{6.66cm}|}{Instead naïve computation of SHAP values most implementations rely on approximations to lessen overhead } \\ 
\hline
\multicolumn{1}{|p{2.53cm}}{Adversarial robustness} & 
\multicolumn{1}{|p{4.13cm}}{Low} & 
\multicolumn{1}{|p{3.17cm}}{Low} & 
\multicolumn{1}{|p{6.66cm}|}{SHAP and LIMEcan be fooled to produce biased outputs (see Slack et al. \cite{slack2020fooling})} \\ 
\hline
\multicolumn{1}{|p{2.53cm}}{Handling of feature interaction} & 
\multicolumn{1}{|p{4.13cm}}{Yes$\ast$, with some recent techniques} & 
\multicolumn{1}{|p{3.17cm}}{No\textsuperscript{+}} & 
\multicolumn{1}{|p{6.66cm}|}{$\ast$Methods like Faith-SHAP \cite{tsai2023faith} and SHAP-IQ \cite{fumagalli2024shap} can compute interaction indices to account for correlations among features. \newline
\textsuperscript{+}LIME generally assumes local behavior to be linear, although tree-based surrogates can help to relax this assumption.} \\ 
\hline
\multicolumn{1}{|p{2.53cm}}{Visualization options} & 
\multicolumn{1}{|p{4.13cm}}{Various (e.g., summary plots, waterfall, force plots)} & 
\multicolumn{1}{|p{3.17cm}}{Feature importance (bar) plot} & 
\multicolumn{1}{|p{6.66cm}|}{Unlike lime, SHAP implementations provide multiple visualization options.} \\ 
\hline
\end{tabular}

} 
\end{adjustbox}
\end{table*}

\subsubsection{Faithfulness metrics for SHAP and LIME}

SHAP and LIME are versatile and popular interpretability methods with many open source implementations. Their relative strengths and limitations are outlined in Table \ref{tab:T2_Comparison_SHAP_and_LIME}. The choice between SHAP and LIME for practical applications also depends on the faithfulness of the selected method. In the case of LIME, since the explanations are generated for a surrogate model, the resulting faithfulness may not represent that of the original model. For this reason, it is also important to know the fidelity of the surrogate, an attribute that measures how accurately the surrogate model mimics the behavior of the original black box in the vicinity of the given sample. Feature ablation-based metrics that are commonly used for evaluation gradient-based methods and class activation maps (discussed in Sections 3.3.1 and 3.3.2) are also widely applied to SHAP (e.g., \cite{meng2022interpretability}) and LIME (e.g., \cite{kakogeorgiou2021evaluating,zhang2024opti}). However, as highlighted in Section 3.3.3, these methods have severe limitations in correctly measuring the faithfulness of attribution methods.

\subsection{Attention-based Methods}

Attention mechanism is an effective technique that is primarily used in neural networks, particularly transformer architectures  \cite{vaswani2017attention}, to improve performance. Trainable attention as a performance-enhancing concept that has long been applied in several domains and has particularly proven effective in fields like natural language processing  \cite{bahdanau2014neural},  \cite{shen2018disan},  \cite{vaswani2017attention} and computer vision  \cite{guo2022attention},  \cite{hassanin2024visual}.  Although the primary role of attention in deep learning is in improving predictive performance, their use in explaining neural network decisions has been vastly successful  \cite{zhang2017top},  \cite{jetley2018learn},  \cite{oktay2018attention}. This use case is motivated by the observation that attention weights directly correspond to feature importance scores. Consequently, attention has also been explored as a means to provide explanations for both CNN and transformer-based models. With this approach, trainable attention layers are used to the approximate importance weights that correspond to the degree to which a network ``pays attention" to different parts of the input when making a prediction. RETAIN \cite{choi2016retain} addresses the interpretability of a recurrent neural network (RNN) using the attention mechanism. The goal is to circumvent the accuracy-interpretability tradeoff training two RNNs sub-modules to generate attention weights for explanation and still "retain" the predictive power of a black box model.

The explanations provided by attention methods are similar to those of gradient-based methods in that both lines of approaches simply learn weights that score portions of the input on importance to the prediction. Thus, like other conventional attribution methods, attention-based interpretability methods do not rely on prior domain knowledge about the input to generate their heatmaps. Consequently, the approach is susceptible to input noise that can cause the model to learn spurious relations.  Furthermore, for transformers in particular, the self-attention heads encode much more information than just attribution scores. Hence obtaining pure relevancy-appropriate information from attention weights scores is challenging. To overcome this problem, some recently proposed methods for extracting relevancy information from transformer attention modules rely on other attribution techniques like Deep Taylor Decomposition (as employed in  \cite{chefer2021transformer} and LRP (as used by Voita et al.  \cite{voita2019analyzing}). In fact, several studies \cite{bastings2020the},  \cite{jain2019attention}, \cite{bai2021why},  \cite{serrano2019is} that investigate the suitability of both CNN and transformer attention weights as explanations have concluded that standard attention methods are incapable of providing true explanations for model decisions. Therefore, the use of trainable attention for this purpose in high stake applications, especially in medical domains, should be approached with caution, if not completely discouraged.

The format of the explanations generated by attention-based interpretability are similar to other attribution methods in that they rely on highlighting relevant features of the input to show their influence on the prediction to humans. In natural language processing, this attribution method may highlight words, tokens, phrases or whole sentences. For vision transformers and CNNs for image classification tasks, saliency scores computed from attention mechanisms are used to construct saliency maps similar to the outputs of gradient-based interpretability methods. Owing to this similarity, faithful evaluation metrics for attention-based explanations follow similar to logic to gradient-based methods. Specifically, the commonest methods rely on feature perturbation to determine the importance of individual pixels or linguistic units. The ROAD, ROAR, AI-AD and insertion-deletion metrics, as well as various variants \cite{deyoung2019eraser,wu2024faithfulness} of these feature ablation methods are suitable options. However, as demonstrated in Sections 3.3.1 and 3.3.2 of this paper, these evaluation metrics do not provide consistent results for different model configurations. Moreover, the metrics do not technically measure true rationales behind predictions and their results across different settings do not align well. Consequently, it is challenging to reliably compare the faithfulness of different attention-based techniques.

\subsection{Counterfactual explanations}

Counterfactuals help to infer outcomes for unobserved, alternative scenarios. In general, counterfactual explanations  \cite{wachter2017counterfactual},  \cite{karimi2020model} seek to answer the question ``what change in the input will cause the prediction to change to a specified outcome. In most cases, the desired outcome is a direct opposite of the original prediction.  Counterfactual examples are usually produced by varying the input in a way that leads to the desired prediction outcome. Counterfactual explanations foster human understanding of complex machine learning models by allowing causal relationships between inputs and decisions to be observed. In multi-class prediction tasks, the goal could be finding a minimal change in the input that results in a predefined label being predicted or a new prediction probability distribution for the predicted classes. The simplest scenarios are binary classification tasks where multiple factors contribute to an outcome with only two possibilities, counterfactual explanations aim to understanding the given outcome by finding an alternative input (or a change in one or more of the factors) that causes the prediction to flip. A typical example is the loan application problem  \cite{guidotti2024counterfactual},  \cite{grath2018interpretable}, where the decision can either be ``approve" or ``decline", and the factors that influence the decision may include age, income and credit score rating. If the initial decision is ``decline", a plausible counterfactual example could be to increase monthly income by an appropriate amount. In addition to enhancing users’ understanding of the predictions of machine learning models, counterfactual explanations can also humans to achieve real-world goals. This can be accomplished through \textit{actionable} \textit{counterfactuals}  \cite{poyiadzi2020face},  \cite{mothilal2020explaining} which mainly focus on crafting counterfactual examples  from variables that correspond to factors in real-life which can be controlled by the humans that the decisions affect. For instance, in the loan application example, counterfactuals based on income and credit score rating can be ``acted upon" by the borrower to influence the decision in the future but factors like age cannot be altered at will. FACE \cite{poyiadzi2020face} proposes mechanisms for algorithmically determining “feasible paths” that transform such difficult-to-influence counterfactuals to actionable ones.

LLM-generated counterfactual explanations (e.g.,  \cite{bhattacharjee2024towards},  \cite{gat2023faithful}) can be used to provide useful insight into the decision-making of other back-box models. To accomplish this goal, the instruction-following capability of LLMs can be exploited to obtain some important but not-so-obvious features of the input that would cause the original model to make a specified prediction.  The LLM can then me prompted to minimally change the input to alter the original prediction. In this context, for example, Bhattacharjee et al.  \cite{bhattacharjee2024towards}  evaluated the ability of three state-of-the-art LLMs –GPT-3  \cite{brown2020language}, GPT-3.5  \cite{ali2023performance} and GPT-4 \cite{achiam2023gpt}– to generate counterfactual explanations for a black-box DistilBERT  \cite{sanh2019distilbert} model applied in classification. The technique involves prompting the LLM to list latent features that influenced a given prediction, followed by another query to identify the words in the text than it associated with the latent features. Finally, the LMM is asked to edit the original input such that the modified text causes the DistilBERT classifier’s prediction to flip. Results from the study show that the LLMs, in particular GPT-4, have a high potential to act as experts in providing causal explanations to the predictions of black box models.

A major advantage of counterfactual explanation methods is that these techniques do not require knowledge of the model’s internal structure or parameters. Additionally, the techniques can be implemented without re-training or modifying the model. For these reasons, counterfactual explanations can be implemented at test time without interfering with performance.

\subsection{Case-based reasoning}

\begin{figure*}[!htb]
	\vspace {1mm}
	\centering
	\includegraphics[width=1.0 \linewidth]{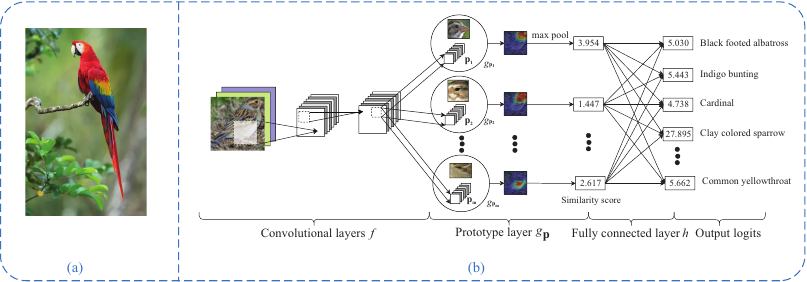} \vspace {-3mm}
	\caption{(a) Scarlet macaw parrot \cite{bright2019multifactorial} with easily distinguishable parts. Prototypical networks use this knowledge to explain predictions. (b) Architecture and computational stages of ProtoPNet.    
	}\label{fig:15_Scarlet}
\end{figure*}

Humans generally recognize object categories by relating their physical appearances to shared characteristics of members of their class, using a paradigm of ``\textit{this} looks like \textit{that}". More importantly, human reasoning about visual concepts is greatly enhanced by the knowledge of compositionality  \cite{palmer1977hierarchical},  \cite{tversky1984objects}, which breaks complex objects or scenes into component parts to make the task of recognition easier. For example, the scarlet macaw parrot shown in Figure \ref{fig:15_Scarlet} (a) can be recognized by its distinctive curved white beak, long colored tail and red neck.  Consequently, decisions that rely on these features can easily be explained. Apart from enhancing recognition accuracy, compositional or case-based reasoning can also help to explain prediction rationales. The explainability in case-based learning frameworks is exemplified in Prototypical Part Networks (ProtoPNets)  \cite{chen2019this},  \cite{wolf2024keep}. ProtoPNets ), whose idea was first introduced by Li et al.  \cite{li2018deep} and refined by Chen et al.  \cite{chen2019this}, uses a case-based learning approach to discover informative image parts or prototypes from training data. ProtoPNet is made up of three main parts: (1) a regular CNN feature extractor; (2) a prototype layer that computes the similarities between features of image patches learned by the CNN layer and discriminative parts; and (3) a fully connected layer which makes final decisions by computing class probabilities based on evidence from learned prototypes. The architecture of this framework is presented in Figure \ref{fig:15_Scarlet} (b).


ProtoPNets are ante-hoc interpretable models that extract explanations from CNNs by jointly learning useful features and their correlations with relevant part-prototypes in an end-to-end manner. As such, they generally do not suffer some drawbacks inherent in post-hoc explainability methods such as instability or even compromised faithfulness  \cite{sixt2020when} . Thus, unsurprisingly, several studies demonstrate the competitiveness of ProtoPNets in practical applications like healthcare target shapes can be modelled as prototypical parts. For instance, the techique has been applied in the medical domain for diabetic retinopathy grading  \cite{hesse2022insightr}, and kidney stone identification  \cite{floresaraiza2023deep}.

 Standard ProtoPNets assume geometrically rigid relationships among parts of an image that can be represented by a finite number of regularly arranged part-prototypes. This limits the ability of these prototypes to capture objects with significant geometric distortions. Deformable ProtoPNet  \cite{donnelly2022deformable} relaxes this assumption by employing prototypes with composite parts that are allowed to assume different spatial relationships to reflect various geometric transformations of the input image. Deformable  ProtoPNet has been shown in qualitative test to boast high performance in situations where objects exhibit significant pose variations.

 Another major limitation of ProtoPNet is that prototypes are learned exclusively on class bases. Consequently, prototypical parts cannot be readily re-used across different classes even if required. ProtoPShare \cite{rymarczyk2020protopshare} allows instances of parts to be shared across different classes by pruning and merging redundant or semantically similar prototypes during training. This results in a more compact design with fewer but more informative prototypes, and the ability to discover similarities among classes.

Furthermore, standard ProtoPNets match fixed-size patches from the given image to part-prototypes. Consequently, learned prototypes may not align with intended targets. For example, in medical imaging applications, these patches struggle to match some disease- specific features with variable sizes. Consequently, in a multi-diseases diagnosis task, Kim et al  \cite{kim2021xprotonet} propose   XProtoNet to extract disease-specific features from dynamically sized regions of the input image, resulting in highly competitive performance gains. Test results on publicly available chest radiography database shows the superior prediction accuracy of XProtoNet along with good interpretability.

  ProtoPNets are not guaranteed to learn semantically meaningful part-prototypes that human experts would consider relevant to the given task. Moreover, since part-prototype networks do not leverage domain from experts, they can also be influenced by the Clever Hans  \cite{lapuschkin2019unmasking} problem , where learned prototypes may not correlate with target classes. Furthermore, part-prototypes may be inconsistent across different images  \cite{huang2011evaluation}, or may be sensitive to image perturbation  \cite{hoffmann2021does}. Accordingly, the faithfulness of their explanations has been questioned by critics (e.g.,  \cite{bontempelli2022concept},  \cite{xudarme2023sanity}). Therefore, human-in-the-loop frameworks such as ProtoPDebug  \cite{bontempelli2022concept}, are proposed to mitigate this challenge by incorporating experts who analyze the explanations and provide feedback to correct wrong outputs created by confounding part-prototypes.

\subsection{Concept-based reasoning}

To address the drawbacks with standard interpretability, some recent interpretability deigns introduce case-based learning. However, the human-interpretable concepts used by this method are automatically extracted from training data. For this reason, they may differ from actual features that humans may use to discriminate target classes. To overcome this limitation, concept learning methods utilize high-level, human-defined concepts the training data which are learned to explain model decisions. These concepts are explicitly incorporate through data annotation with concept labels. Testing with concept activation vectors (TCAVs) and concept bottleneck models (CBMs) are the two groups of concept-based learning method. CBMs are by far the most popular family of approaches.


\begin{figure*}[!htb]
	\vspace {1mm}
	\centering
	\includegraphics[width=1.0 \linewidth]{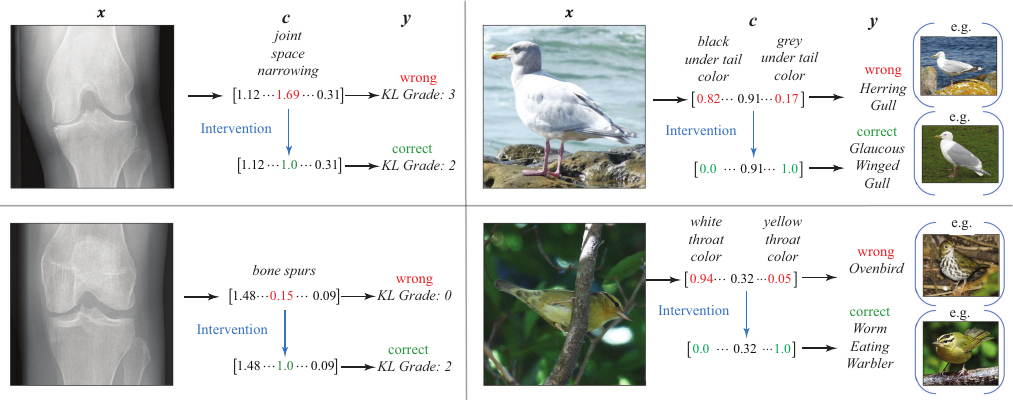} \vspace {-3mm}
	\caption{An illustration of concept edits in CBMs \cite{koh2020concept} for arthritis prediction (LEFT) and bird species identification (RIGHT). For wrong classification decisions, concept scores predicted by the model are edited by changing incorrect values (this process is shown by vertical blue arrows) in order to correct the prediction.    
	}\label{fig:16_CBMs}
\end{figure*}


\subsubsection{Testing with concept activation vectors}

Kim et al.  \cite{kim2018interpretability} achieve interpretability   through the so-called Testing with\textit{ }Concept Activation Vectors (TCAVs) paradigm\textit{. } The method produces explanations by determining the correlation between human-defined visual concepts and a model’s class scores. Concept Activation Vectors (CAVs) are obtained from the weights learned by a linear classifier that seeks to distinguish between activations of human-provided concept examples from random examples. In most practical situations, developers must provide additional concept annotations to be used for training. This process is time consuming to carry out on a large scale. Consequently, ACE  \cite{ghorbani2019towards} is proposed to automatically extract dataset-wide concepts and to obviate the need for manually-labeled concept example. The TCAVs approach has a number of advantages. First, it is a post-hoc method designed to extract explanations from black-box networks, hence the risk of compromising performance.

\subsubsection{Concept bottleneck models}

A seminal work by Koh et al. \cite{koh2020concept} introduced Concept Bottleneck Models (CBMs) that learn high-level concepts directly from raw image data.   The original implementation of the method was used to predict arthritis severity from knee x-ray images using a set of human-interpretable clinical concepts like ``narrowing of joint space", or the presence of ``bone spurs". To accomplish this, the network first learns pre-defined clinical concepts from the input images, and then, using these concepts together with learned low-level features, computes the final class scores. In the same study, the method was tested on bird species recognition using the Caltech-UCSD Birds-200 dataset (CUB)  \cite{wah2011}. Both applications showed promising results.

As shown in Figure \ref{fig:16_CBMs}, CBMs also allow experts to edit mis-predicted concept values at inference time, ensuring the correct concepts are used for the final prediction. Koh et al.  \cite{koh2020concept}show that the involvement of human expert s in the prediction task through concept correction helps improve accuracy.

CBMs generally achieve interpretability by sacrificing some degree of accuracy due to additional constraints that they introduce, although this problem can be mitigated by other means such as test time intervention. Furthermore, empirical studies by Margeloiu et al.  \cite{margeloiu2021do} reveal that CBMs do not always learn desired high-level concepts as intended when trained with a prediction goal. They observe that the models learn additional features that are not related to useful concepts. The presence of confounding factors is inevitable since these methods mainly focus on highlighting the contribution of important concepts, but they do not directly suppress spurious features in the training data. This may limit the predictive value as well as the interpretability of concepts. Since the revelation of this phenomenon, also known as \textit{information leakage}, several workarounds  \cite{lockhart2022towards},  \cite{havasi2022addressing},  \cite{marconato2022glancenets} aimed at mitigating the problem have emerged. GlanceNets  \cite{marconato2022glancenets}, for instance, formulates interpretability goals to facilitates alignment with the network’s data generation process, and then uses disentangled representation learning to achieve the alignment. Lockhart et al.  \cite{lockhart2022towards} use Monte-Carlo (or inference-time) dropout to estimate concept prediction uncertainty, which enables the information leakage to be reduced.  Additionally, some works (e.g., RCC ( \cite{stammer2021right}) utilize Neuro-Symbolic learning frameworks that allow users to interactively edit the association of irrelevant semantic concepts with model predictions. However, to reliably exclude unwanted concepts from being used for prediction, these spurious concepts must necessarily be known in advance. To address this limitation, Yan et al.  \cite{yan2023towards} propose a concept-based human-in-the-loop skin cancer diagnosis model which allows direct intervention to remove spurious features during training. The model is designed to automatically learn both relevant and confounding concepts based on the approach of TCAV and co-occurrence logic. At each step, the model presents its prediction along with an explanation of the decision logic to a human expert can then remove any confounding concepts (e.g., the presence of dark corners, dense hairs, rulers) that influenced the decision by rewriting the first-order-logic formula that underlie the given prediction.    Other approaches aimed at improving CBMs include the selection of appropriate intervention mechanisms  \cite{shin2023a} and the use of a large number of concepts by introducing additional unsupervised concepts  \cite{sawada2022concept}.

\section{EXPLAINABILITY OF LLM PREDICTIONS}

Large language models (LLMs) achieve their tremendous performance by relying on massively large transformer architectures with the capacity to encode huge feature sets through pre-training. By virtue of this complexity, and their use of diverse and huge volumes of data in pre-training, LLMs acquire a significant amount of knowledge. Owing to this knowledge, they are highly proficient in a wide range of tasks, boasting impressive zero-shot generalization capability.  Unfortunately, their sophisticated design comes at a high cost of poor interpretability: end-users do not usually understand the reasoning behind the models’ decisions (or responses). However, interpretability of LLMs is needed to maximize their use. Explainabibility can also help to identify hallucinations  \cite{tonmoy2024a},  \cite{zhang2023sirens},  \cite{huang2023a} in LLM predictions. Hallucinations are wrong predictions that do not conform with real world knowledge.

Therefore, given the widespread use of LLMs and the inevitable need for understanding the decisions of deep learning models in critical applications, research interest aimed at addressing this problem is growing at a fast pace. These efforts in explaining the decisions of large language models usually focus on two main lines of approaches: 1) local explanations–i.e., explaining the rationales of individual predictions; and 2) global or mechanistic explanations–i.e., providing insights into the general decision logic of the model as a whole.

\subsection{Local explanations of LLM predictions}

\begin{figure}[H]
	\vspace {-1mm}
	\centering
	\includegraphics[width=1.0 \linewidth]{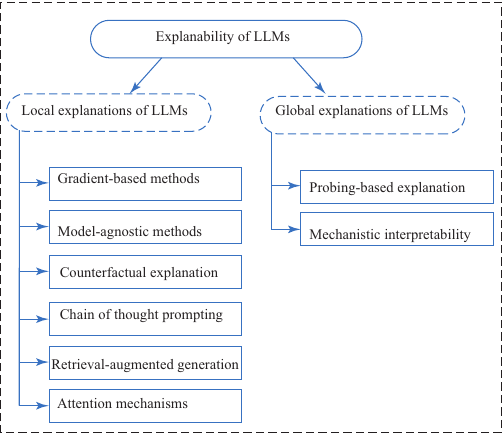} \vspace {-3mm}
	\caption{Taxonomy of LLM interpretability methods covered in this section
	}\label{fig:17_LMM_Outline}
\end{figure}

By focusing on explaining individual predictions, local explanations could be useful to practitioners or end-users who apply an LLM to solve important problems in an ad-hoc manner. In these situations, users are more concerned about the validity of the prediction of interest rather than the overall performance of the model. A common concern for end-users is to know, on a case-by-case basis, whether the outcomes of predictions should be accepted. On the other hand, developers or experts in machine learning systems may be more interested in having a ``big picture" or global view of a model’s decision-making process. Figure \ref{fig:17_LMM_Outline} shows the categories of interpretability methods for LLMs that are covered in this section. 

\subsubsection{Local explanations of LLM via gradient-based methods}

In general, gradient-based methods characterize feature importance from the partial derivative of the prediction with respect to the input. These interpretability methods are already popular techniques for explaining the decisions of machine learning models, particularly in computer vision tasks like image classification. In these tasks, model inputs are usually images which are composed of spatial arrangements of pixels that represent features of different objects present in the given image. Therefore, computing the sensitivity (or gradient) of a class prediction with respect to input provides spatial localization of pixels that contribute most strongly to the given prediction and can thus serve as the basis for the decision. On the other hand, in natural language processing tasks, input features are the tokens or words supplied in the prompt query. In these tasks, the goal of gradient-based interpretability is to assign relevancy scores to input tokens to capture their influence on the model’s prediction. Again, feature relevance may be computed from the gradient of the predicted text with respect to the given input word. This gradient specifically considers the change in the prediction in response to minute changes in the input features. However, for LLM and other NLP applications, the feature vector in question is typically discrete rather than a continuous space of values. As a result, making infinitesimally small changes is often a challenge. Consequently, it is difficult to directly apply gradient-based techniques which operate in the continuous space. These methods often require significant workarounds in order to utilize the non-discrete gradients in  natural language processing (NLP) domains.  A common technique is to avoid using the raw input directly and instead rely on the one-hot-encoding or, on making feature changes in the embedding space. Unfortunately, since most of the state-of-the-art LLMs are closed-set and do not provide direct access to their feature embeddings, this method is impractical in most cases.

In view of this difficulty, applications of conventional gradient-based interpretability methods in NLP have largely been restricted to conventional large-scale language models like BERT  \cite{devlin2018bert}; their use in modern large language models has still not achieved the desired success. Other notable limitations of vanilla gradients in NLP applications include their susceptibility to input shifts, gradient saturation, and the discrete nature of text inputs.

 In order to improve the invariance of the gradient-based method to input shifts, the \textit{gradient $\times$ input} technique \cite{kindermans2019reliability,shrikumar2016not} is often used. This technique prevents the tendency of the computed saliency scores to be undesirably affected by inconsequential changes such as small constant shifts in the input which do not usually affect the prediction itself.  Implementations of this framework in NLP tasks \cite{mohebbi2021exploring} typically multiply the computed gradient by the word embedding. In \cite{mohebbi2021exploring}, Mohebbi et al. apply this technique to successfully estimate the influence of individual input tokens on the predicted labels of a BERT text classifier. 

Also, given the sheer depth and overall size of state-of-the-art NLP models, interpretability frameworks that utilize vanilla gradient often show excessive noise owing to the gradient saturation problem. Although this problem can be ameliorated by using less saturation-prone activation functions such as the GELU  \cite{hendrycks2016gaussian} employed  in some NLP models like BERT, Integrated Gradients (IG) shows less susceptibility to these gradient problems even for models that use functions like RELU. In these implementations of IG in NLP applications (e.g.,  \cite{mudrakarta2018did}, \cite{bastings2021will}), an empty text label can serve as this baseline. Despite its adoption and relatively less noise, IG takes the integral of gradients along paths which are continuous straight lines, and thus, is not compatible with input features in NLP domains which are essentially discrete words. This hampers the accuracy of computed attribution scores. Discretized Integrated Gradients (DIG) \cite{sanyal2021discretized} addresses the limitation by interpolating between the input and baseline along non-linear paths to obtain discretized points which are close to the original discrete words. DIG therefore achieves a more accurate attribution than standard IG on sentiment classification. However, with DIG and IG, it is not always straightforward which baseline to use, although this choice may be critical to performance (see  \cite{bastings2021will}). Moreover, Enguehard  \cite{enguehard2023sequential} notes that recovering whole sentences from the interpolated points can result in distortions in meaning.

To alleviate this shortcoming, the author proposes Sequential Integrated Gradients (SIG) \cite{enguehard2023sequential} which sequentially computes the attribution separately for each word at a time while keeping remaining words intact while generating interpolated points between the baseline and the target word. Also, SIG simplifies the choice of baseline by using the [MASK] token exclusively, which results in good performance. However, the computational demand of this framework can be significantly higher than vanilla IG since this method computes attributions separately for each word. Table \ref{tab:T3_Apps_gradientbased} summarizes the LLM interpretability methods using gradient-based techniques

\begin{table*}[!htbp]
	
	\caption{Applications of gradient-based methods in the interpretability of large language models}
	\label{tab:T3_Apps_gradientbased}
	
\renewcommand{\arraystretch}{1.3}
\begin{adjustbox}{max width=\textwidth}
\begin{tabular}{p{4.09cm}p{4.05cm}p{3.1cm}p{5.25cm}}
\hline
\multicolumn{1}{|p{4.09cm}}{\textbf{Goal}} & 
\multicolumn{1}{|p{4.05cm}}{\textbf{Approach}} & 
\multicolumn{1}{|p{3.1cm}}{\textbf{Example usage}} & 
\multicolumn{1}{|p{5.25cm}|}{\textbf{Major limitations}} \\ 
\hline
\multicolumn{1}{|p{4.09cm}}{Computationally-efficient computation of attribution scores} & 
\multicolumn{1}{|p{4.05cm}}{Computing saliency score using vanilla gradients } & 
\multicolumn{1}{|p{3.1cm}}{Not suitable enough for NLP tasks} & 
\multicolumn{1}{|p{5.25cm}|}{1-Gradient saturation \newline
2-Noise \newline
3- Issue of input shifts \newline
4-Relatively low performance} \\ 
\hline
\multicolumn{1}{|p{4.09cm}}{Improve shift invariance of input features} & 
\multicolumn{1}{|p{4.05cm}}{Gradient $\times$ input \cite{kindermans2019reliability,shrikumar2016not} } & 
\multicolumn{1}{|p{3.1cm}}{Mohebbi et al. \cite{mohebbi2021exploring} } & 
\multicolumn{1}{|p{5.25cm}|}{1-Gradient saturation \newline
2-noise \newline
3-improved but still unsatisfactory performance} \\ 
\hline
\multicolumn{1}{|p{4.09cm}}{Mitigate gradient saturation, reduce noise and improve quality of explanation} & 
\multicolumn{1}{|p{4.05cm}}{Integrated Gradients (IG) \cite{sundararajan2017axiomatic} } & 
\multicolumn{1}{|p{3.1cm}}{Bastings et al. \cite{bastings2021will} } & 
\multicolumn{1}{|p{5.25cm}|}{1-Relatively high computational complexity \newline
3-Slightly compromised performance owing to continuous path used by IG \newline
3- Difficulty in selecting appropriate baseline} \\ 
\hline
\multicolumn{1}{|p{4.09cm}}{Capture feature interaction} & 
\multicolumn{1}{|p{4.05cm}}{Integrated Directional Gradients (IDG)  \newline
} & 
\multicolumn{1}{|p{3.1cm}}{Sikdar et al. \cite{sikdar2021integrated} \newline
} & 
\multicolumn{1}{|p{5.25cm}|}{1-relatively high computational complexity \newline
} \\ 
\hline
\multicolumn{1}{|p{4.09cm}}{Improve performance by discretizing the continuous path} & 
\multicolumn{1}{|p{4.05cm}}{Discretized Integrated Gradients (DIG) \cite{sanyal2021discretized} } & 
\multicolumn{1}{|p{3.1cm}}{Sanyal and Ren 2021 \newline
  \cite{sanyal2021discretized}} & 
\multicolumn{1}{|p{5.25cm}|}{1-Relatively high computational complexity \newline
2- Difficulty in selecting appropriate baseline} \\ 
\hline
\multicolumn{1}{|p{4.09cm}}{Address the problem of baseline selection} & 
\multicolumn{1}{|p{4.05cm}}{Sequential Integrated Gradients (SIG) \cite{enguehard2023sequential}} & 
\multicolumn{1}{|p{3.1cm}}{Enguehard et al. \cite{enguehard2023sequential}} & 
\multicolumn{1}{|p{5.25cm}|}{2-Even higher computational demand compared to IG and DIG} \\ 
\hline
\end{tabular}

} 
\end{adjustbox}
\end{table*}

\subsubsection{Local explanations of LLM via model-agnostic methods}

Local explanations can be generated for LLM via model-agnostic or perturbation-based methods. LIME and SHAP (see Section 3.4) are classic examples of model-agnostic attribution frameworks used to generate post-hoc explanations for machine learning systems regardless of model type. Since these techniques achieve interpretabilty by utilizing only the information on a model’s input and output, they are typically flexible enough to be applied to a wide range of deep learning frameworks, including language models. In essence, model-agnostic\textbf{ }methods only require access to the model’s input and outputs. One of the simplest and most effective ways of establishing the influence of a feature on a model’s prediction based on this framework is by excluding the feature from the input and observing the change in prediction or by sampling a subset of features to analyze the response for. Hence, in natural language processing, this interpretabilty by perturbation can be accomplished by analyzing the change in generated text when a word (or a group of words) is removed from the input sentence.

 Leave-One-Out (LOO) is one of the simplest interpretabilty frameworks for NLP models. The approach determines the importance of a word or phrase in producing a prediction by computing the change in the output when the feature of interest is removed from the input. A feature (in this case, a word or phrase) is usually removed by masking it out with zeros (e.g.,  \cite{prabhakaran2019perturbation}) or by replacing it with an [UNK] token using BERT (e.g.,  \cite{harbecke2020considering},  \cite{kim2020interpretation}) and marginalizing over all predicted words. These frameworks achieve decent results and are fairly simple in design.  However, removing single words or phrases fails to capture interactions among certain words and phrases. In most cases, the meanings or roles of certain words in sentences are influenced by other words that occur in these sentences. Hence, computing the contribution of individual words in isolation is not guaranteed to achieve optimum results. To address this limitation, some methods have been proposed to capture interactions among input features  \cite{jin2019towards},  \cite{chen2020generating}. 

     The HEDGE framework  \cite{chen2020generating} detects feature interactions and iteratively clusters these features into individual words and shorter phrases based on their interactions. This framework employs the SHAP algorithm for feature relevance attribution. Jin et al.  \cite{jin2019towards}achieve hierarchical explanation by determining context independent contribution of constituent features, i.e., by comparing the influence of two words or phrases on the prediction when these components appear together against the sum of their effects when they are used separately in the input sentence. 

  The methods described are mainly designed to generate explanations for relatively simple tasks like sentiment classification where simpler language models like BERT are used. Furthermore, in simple classification tasks, feature attribution can be achieved by occluding a token and analyzing the resulting change in prediction so as to quantify the contribution of the occluded feature to the model’s output. On the other hand, text generation is a progressive process where previous words help to provide context to the current one, and can thus heavily influence the next-word prediction. 

For some text generation tasks performed using large language models like ChatGPT, the input query can be a long-form text with complex structure involving the interactions of words, phrases and sentences in ways that make the information difficult to decipher by simple frameworks. To effectively analyze the complex linguistic structure, understand syntactic relationships that define the appropriate context in which the words are used. Thus, interpretability methods designed for generative LLMs should be capable of leveraging knowledge of the compositionality of linguistic information. Besides, the output space of generated text can be prohibitively large for these tasks.  Owing to these challenges, only a few works, specifically the more recent studies (e.g., SyntaxShap  \cite{amara2024syntaxshap} and MExGen  \cite{paes2024multi} have proposed methods capable of providing relevancy attributions for generative natural language processing  pipelines such as LLMs. These studies leverage the spaCy open-source python library  \cite{honnibalspacy} to segment long sentences into fragments and the use SHAP and LIME to generate attributions.  spaCy is a highly proficient NLP tool that helps in understanding linguistic structure of sentences and discovering the relationships among words in sentences. It facilitates various applications including entity recognition, tokenization and dependency parsing of text.

   SyntaxShap  \cite{amara2024syntaxshap} uses  dependency parsing trees to extract syntactic relations among words in a sentence. Using this information, SyntaxShap can then determine the influence of a particular word on the next-word prediction by retrieving coalitions of words along different paths in the dependency parsing tree. Similarly, MExGen employs spaCy to divide the input text into smaller parts at various levels, namely, word-, phrase- or sentence-level granularities. This method enables attributions to be computed from a course (e.g., sentence) to fine (e.g., word) level and allows the user to specify the granularity at which attributions are generated. MExGen also uses so-called \textit{scalarizers} which help to convert the generated text to numeric values to make the output amenable to SHAP and LIME computations. Amara et al. evaluate the efficacy of SyntaxShap on text generation of state-of-the-art large language models, specifically, GPT-2  \cite{radford2019language} and Mistral 7B  \cite{jiang2023mistral}. The results show a high degree of coherence and faithfulness. Although the evaluations in this study were carried out using inputs with single sentences, the method can easily be extended to larger text fragments comprising multiple sentences by treating each sentence as a unit and then leveraging the underlying context of previous sentences to determine the next. 

\subsubsection{Counterfactual explanations of LLMs}

Counterfactual explanations  \cite{wachter2017counterfactual},  \cite{karimi2020model} belong to a broader category of example-based explainability methods.  The category also includes the use of adversarial examples. These approaches are classified as local interpretability methods because the examples are designed to reveal reasoning of the model in reference to a given prediction. In classification tasks, the goal could be finding a minimal change in the input that results in a predefined label being predicted or a new probability distribution for the predicted classes. Counterfactual examples are usually composed by varying the input in a way that leads to the desired prediction outcome. Counterfactual explanations foster human understanding of complex machine learning models by allowing causal relationships between inputs and decisions to be observed without interfering with the model’s internal architecture. In large language models, example-based explanations (e.g.,  \cite{shu2024counterfactual}) can be exploited to mitigate input prompt attacks by uncovering inconsistences in reasoning. Models are expected to exhibit the desired behavior when presented with the counterfactual examples, otherwise, unanticipated deviations could indicate serious problems. Also, by presenting appropriate hypothetical scenarios, counterfactuals can also help to discover and deal with inherent bias  \cite{howard2024uncovering},  \cite{huang2019reducing} in LLMs.

 Another important area where counterfactual explanations are of growing interest is the domain of analogical reasoning. Analogical reasoning is the process of solving new problems by drawing comparisons (or making \textit{analogies}) with the solutions of different but related problems. This feature is a critical capability needed for zero-shot generalization and open-domain task performance.  By their design, counterfactual examples also allow researchers to evaluate analogical reasoning ability of LLMs. An influential study \cite{webb2023emergent} has recently conducted direct empirical comparisons between human subjects and the instruction-tuned GPT-3 on analogical reasoning tasks and concludes that LLMs like GPT-3 can achieve competitive performance to their human counterparts in these difficult reasoning problems. However, in a direct response, two studies (\cite{hodel2023response}, \cite{lewis2024using}) leveraged counterfactual examples to present results that contradict Webb et al.’s \cite{webb2023emergent}. Hodel $\&$ West  \cite{hodel2023response} argue that  although the data used in the evaluations may be rare, they are unlikely to be entirely unseen by GPT-3 and -4 during training and, as a consequence, the impressive performance of the LLMs could not be attributed to analogical reasoning alone. The studies (Hodel $\&$ West \cite{hodel2023response} and Lewis $\&$ Mitchell  \cite{lewis2024using}) further generate counterfactual examples similar to the \textit{letter-string} analogy \cite{hofstadter1984the} (Hofstadter, 1985) by making intuitive variations of the original data. The results demonstrate GPT’s inability to reason analogically at the level of humans. However, in a sharp rebuttal, Webb et al. \cite{webb2024evidence} defend the approach (and results) in their original study  \cite{webb2023emergent} and provide new evidence that show that LLMs can in fact solve simple counterfactual reasoning problems. The authors  \cite{webb2024evidence}  instead blame the poor performance recorded in  \cite{hodel2023response},  \cite{lewis2024using} on the models’ inability to convert the alphabet sequences into numeric indices representing their positions.

debate shows that while counterfactual examples are useful in explaining the reasoning of complex models like LLMs, the approach cannot conclusively account for the behavior of these models in complex tasks such as emergent analogical reasoning.  For instance, in these settings, failures on designated tasks cannot be explained solely by task performance since there could be some underlying challenges behind the poor performance.  Some of these limitations could be addressed by leveraging other interpretability techniques such as mechanistic approaches to understand the internal mechanisms of LLM predictions. One such approach is the so-called \textit{knowledge neuron}  \cite{wang2024unveiling,chen2024journey,dai2021knowledge}, concept, which seek s to directly reveal the information captured in the hidden states of neural networks by studying neuron activations for given network. Interestingly, \textit{knowledge neurons }are mainly concerned with the storage of factual knowledge, rather than knowledge associated with complex reasoning processes.


\begin{figure*}[!htb]
	\vspace {1mm}
	\centering
	\includegraphics[width=1.0 \linewidth]{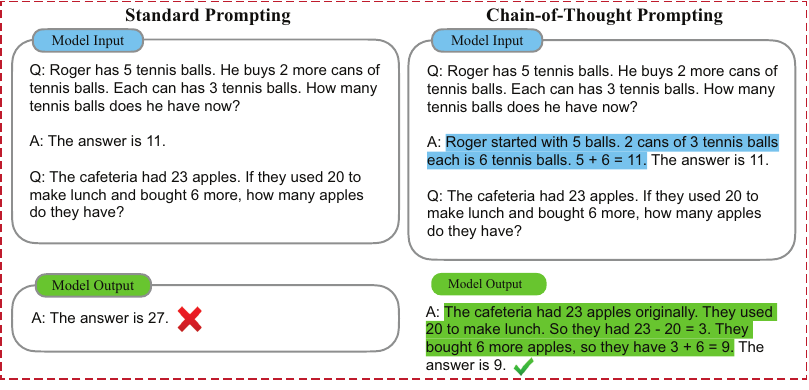} \vspace {-3mm}
	\caption{Illustrating of standard (LEFT) and chain-of-thought prompting (RIGHT) in large language models. The second example shows that the chain-of-thought technique can encourage an LLM to reason correctly toward the right answer. This output is from the original paper by Wei et al. \cite{wei2022chain}.    
	}\label{fig:18_CoT}
\end{figure*}


\subsubsection{Chain of thought prompting}

Chain-of-thought (CoT) prompting  \cite{wei2022chain}, \cite{mitra2024compositional}, which breaks the model’s decision process into intermediate logical steps, is an effective way of understanding the rationale of a large language model’s predictions. With this method of LLM interpretability, the model is prompted in a such a way as to produce a systematic flow of information eliciting the reasoning leading to the final response.  

  Wei et al.  \cite{wei2022chain}demonstrate  that chain-of-thought prompting can also help to align the responses of large language models to user needs and improve prediction accuracy (see example in Figure \ref{fig:18_CoT}). They present empirical results that show that, on difficult reasoning tasks, chain-of-thought reasoning achieves significant performance gains over standard prompting baselines. Furthermore, chain-of-thought interpretability techniques are expected to be adequately faithful since their outputs reflect the model’s inherent reasoning. Thus, the explanations are effectively the models’ own decision logic and must therefore represent the underlying rationale. However, Turpin et al.  \cite{turpin2023language} show that while chain-of-thought techniques are capable of enhancing interpretability and performance, they can sometimes fail to capture the true rationales of model decisions when LLMs are not prompted properly. The study demonstrates that popular large language models can often generate plausible chain-of-thought explanations justifying incorrect predictions. Similarly, Bao et al.  \cite{bao2024how} went further to show that state-of-the-art LLMs like ChatGPT and Llama2 can even produce correct predictions despite incorrect reasoning procedure, or conversely, wrong predictions can follow from correct chain-of-thought sequences. These findings suggest that while chain-of-thought reasoning can break down a prediction of LLM into logically sound parts, these explanations may not be faithful to the model’s decision-making process – the LLM may only be producing plausible output based on the query. 

Recently, more advanced variants of CoT, such as Tree-of-Thoughts (ToT)  \cite{yao2023tree} and Graph-of-Thoughts (GoT) \cite{besta2024graph} have been proposed to handle complex reasoning processes.   ToT casts the LLM inference pipeline in a tree structure which facilitates systematic reasoning along multiple reasoning paths and enables plausible options to be evaluated in order to arrive at the desired decision. GoT formulates the LLM’s reasoning as a graph, with vertices representing thoughts while edges represent the relations among thoughts. These methods show significant improvements over standard chain-of-thought methods and allow decisions to be reached in more difficult reasoning tasks involving sophisticated dependences of thoughts.

\subsubsection{Retrieval-augmented generation}

While the knowledge encoded by LLMs is fixed after the initial pre-training (fine-tuning) phase, information in the real world is in a state of constant evolution. Therefore, it becomes imperative to leverage external sources of information to augment the static knowledge of LLMs when responding to queries requiring updated information. Furthermore, for some tasks, it is sub-optimal to rely on generic knowledge when information specific to the question is readily accessible. For instance, it is counterproductive to use general information to provide advice on fixing problems on a specific machine tool when the product has a manufacturer’s troubleshooting guide. Retrieval-augmented generation (RAG)  \cite{guu2020retrieval},  \cite{lewis2020retrieval}  \cite{lewis2020retrieval}  \cite{guu2020retrieval} addresses this need by enabling the incorporation of external information in the inference framework of large languages models.

 In RAG, the input prompt is supplemented with external and targeted piece of information which augments the existing static knowledge of the language model. By integrating a knowledge retrieval module into the text generation process, retrieval-augmented generation methods can generate more accurate and context-appropriate responses to meet user needs. Also, predictions of RAG can be more explainable since they constrain the output within a given context. When desired, the LLM can be instructed to show relevant references to citations or the information sources in the final output.

\subsubsection{LLM interpretability via attention mechanisms}

Attention mechanisms are known to encode useful information on the reasoning of deep neural networks. Hence, as discussed in Section 3.5, the transformer attention layer has been identified as a source of information on the model’s decision. And, although relying on attention for model interpretability has been severely criticized by multiple studies, the idea remains an enticing prospect since it leverages the encoded knowledge supposedly used for prediction to explain the network’s decision. Previous attempts (e.g.,  \cite{vashishth2019attention}) to settle the debate on the interpretability of attention weights in natural language processing tasks have not led to conclusive results. 

 In natural language processing, the approach provides interpretability by analyzing attention heads to determine which tokens contributed most towards a given decision. Common techniques compute attribution scores from raw attention weights  \cite{clark2019what},  \cite{kovaleva2019revealing} and then visualize the results through saliency heatmaps or Bipartite graphs.  To achieve high performance on state-of-the-art frameworks like LLMs, more recent techniques  \cite{modarressi2023decompx},  \cite{kobayashi2023analyzing} incorporate other network components (specifically, feed forward blocks) beyond attention. This development follows previous observations that feed forward layers in LLMs like GPT-2 act as key-value memories that encode interpretable patterns  \cite{geva2022transformer},  \cite{geva2020transformer},  \cite{gurnee2024universal}. These techniques rely on decomposing the various components to obtain the final explanation outputs. For instance, they use refined attention weights which incorporate additional information from feed forward blocks to improve performance. The algorithms first decompose the operations in the multi-head attention into appropriate vectors and then compute the norm of the output to quantify the contribution of relevant features to the prediction. The method can also decompose the output of the feed forward blocks using Integrated Gradients.

\subsection{Global explanations of LLM predictions}

While local LLM interpretability methods are concerned with explaining the features of the input contributing to individual model predictions, global approaches seek to provide insights on the models’ \uline{general reasoning process}. We discuss two main family of approaches to global explanation of large language models: 1) probing techniques, which aim to understand the knowledge the LLM has learned by probing the model with appropriate queries; and 2) mechanistic explanations, which rely on uncovering the knowledge encoded in the internal representations of these models.

\subsubsection{Probing-based explanation of LLMs}

Given the impressive linguistic capabilities of modern large language models, it is natural to expect that these language models have the capacity to explain their own internal representations or their reasoning. Probing-based interpretability techniques attempt to accomplish this goal through appropriate prompting. Because these methods work by analyzing the embedding or behavior of an NLP model as a whole, they are categorized as global explainability methods.  The main approaches to probing-based interpretability of large language models include 1) classifier-based probing; and 2) zero-shot probing using only data.  The probing classifier methods are among the most widely used techniques for probing large language models for insights.                                          

\paragraph{Probing classifiers}

The method relies on extracting LLM representations to train a separate classifier which in turn is used to predict linguistic features or reasoning skills of the LLM. Two major popular lines of work exist on probing classifiers: probing for knowledge or knowledge probing; and probing for representations.

\textbf{Knowledge probing}: Knowledge probing focuses on using the classifier model to understand the knowledge captured by an LLM through the model’s responses to queries about the knowledge being tested. This approach is mainly used to evaluate whether the LLM understands a given concept or has the relevant knowledge being probed for. For example, COPEN  \cite{peng2022copen} used a linear classifier-based probing to evaluate the conceptual knowledge implicitly captured by popular LLMs including GPT-2  \cite{radford2019language}, T5  \cite{c2020exploring} and BART \cite{lewis2019bart}. In COPEN, Peng et al.  \cite{peng2022copen}investigate three key aspects of conceptual knowledge by LLMs , specifically, whether LLMs: a) recognize conceptual similarities, i.e., given an example entity, can the LLM select the most similar concept?;  b) understand basic properties of common concepts; c) understand concepts from their given contexts, for example, presented with the statement  ``Venza is fuel efficient but slow on rough roads", it would be interesting to know if the model can correctly infer from the context that \textit{Venza} is a vehicle. Peng et al observe that LLMs generally perform poorly on these tasks. For instance, the authors find that LLMs frequently suffer from what they term \textit{conceptual hallucination}, a phenomenon where the model mistakenly associates certain properties with the wrong concepts. They attribute the problem to wrong co-occurrence logic that may cause the LLM to associate entities because they occur together in the pre-training data.

\begin{figure}[H]
	\vspace {-1mm}
	\centering
	\includegraphics[width=1.0 \linewidth]{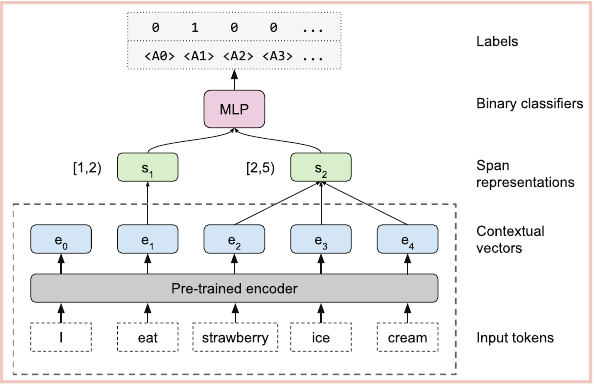} \vspace {-3mm}
	\caption{Probing classifier architecture by Tenney et al. \cite{tenney2019you}. The span pooling and MLP modules of the network learn to utilize features from the contextual vectors. The trained module receives only short-span embeddings (e.g., s1 =[1, 2)–“eat”; s2 =[2, 5)–“ strawberry ice cream”;) from a frozen pre-trained encoder (shown in dashed-line box) and predicts the label <A1>.  
	}\label{fig:19_Probing_Classifier_Architech}
\end{figure}

\textbf{ Probing for representations:} This second category of probing classifiers aims at probing the LLM to examine the similarity between the internal representations of the LLM and some known linguistic properties to understand how the LLM encode knowledge. Since the impressive performance of language models across a wide range of complex tasks is facilitated by the ability of these networks to encode world knowledge implicitly acquired through training., one of the most effective ways of determining what aspect of linguistic knowledge is learned by pre-trained language models is through training simple classifiers on the activations of the language models to identify the presence of specific linguistic features in the embedding space of the models. A popular line of work in this area (e.g.,  \cite{hewitt2019a},  \cite{tenney2019what}) applies the probing classifier approach to study how linguistic knowledge is encoded in word representations of language models. Tenney et al. \cite{tenney2019what} analyze how language models with contextualized word embeddings encode sentence structure. In the study, syntactic and semantic knowledge of LLM is tested by utilizing span representation and MLP classifiers to capture relevant information from the contextual embeddings of the pre-trained language model (see Figure \ref{fig:19_Probing_Classifier_Architech} for the basic structure and workflow of the technique). The authors find that these language models have a better understanding of syntax but performs relatively poorly on semantic reasoning tasks. Similarly, by using probing classifiers and applying appropriate norms and distance metrics on language models’ word embeddings, Hewitt $\&$ Manning  \cite{hewitt2019a} show that syntax trees are implicitly embedded in word representations of  ELMo  \cite{peters2018deep} and BERT  \cite{devlin2018bert} language models. Other aspects of linguistic knowledge that have successfully been studied through probing neural representations include hyperbole  \cite{schneidermann2023probing}. Besides determining the presence of specific linguistic features, representational probing methods also seek to establish which layers or attention heads in the given language models encode these features.

Furthermore, by probing the internal representations of LLMs, multiple studies \cite{peters2018dissecting,jawahar2019does} have shown that lower network layers usually encode word-level features, while deeper layers combine these lower-level embeddings into higher- or sentence-level representations that facilitates more sophisticated capabilities like semantic knowledge. Some other studies (e.g., \cite{gurnee2023finding}) utilize sparse probing technique that enables the analysis of feature representations at finer granularity such as neuron-level, thus facilitating greater detail in interpretability.

However, despite their popularity, some influential studies have cast doubts on the suitability of probing classifiers to truly reveal LLM behavior. For instance, Hewitt and Liang  \cite{hewitt2019designing} demonstrate that probing classifiers can achieve comparable accuracies on real linguistic representations as random baselines which do not encode meaningful linguistic properties. This means the classifier can achieve high performance by simply fitting the training data rather than actually learning relevant internal representations of the language model. The authors propose to mitigate this problem with lightweight classifiers which may not have adequate capacities to memorize the probing task. However, in  \cite{pimentel2020information}, the focus shifts back towards higher performing classifiers which are more capable of extracting relevant linguistic features and thus, can better retrieve encoded knowledge in pre-trained language models. Despite the spate of research in this area, the existence of these contrasting views only underscores the fact that definite best practices in probing classifier design are not yet fully established.

\paragraph{Data-based zero-shot probing}

\begin{figure}[H]
	\vspace {-2mm}
	\centering
	\includegraphics[width=1.0 \linewidth]{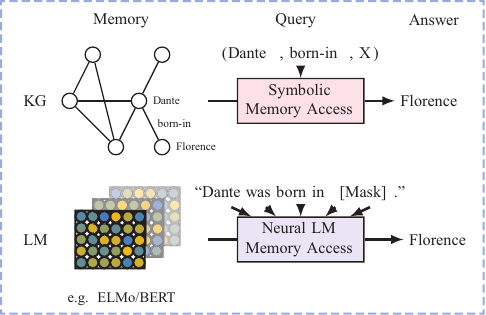} \vspace {-5mm}
	\caption{Petroni et al. \cite{petroni2019language}: A knowledge graph (KG) encodes relational information in the form of (subject, relation, object) triples like (Dante, born-in, X). To understand the factual knowledge capacity of a language model (LM), the model can be prompted to predict a target object by filling the masked token in the query “Dante was born in [Mask]”.  
	}\label{fig:20_KG}
\end{figure}

Owing to the potential faithfulness limitations of probing classifiers, some works propose to directly probe the knowledge of language models with datasets of carefully crafted queries. Here, the model’s predictions provide the needed information on the existence the linguistic knowledge being tested. The zero-shot probing approach is commonly used in testing the capabilities of language models in tasks such as factual knowledge about relationships (e.g.,  \cite{petroni2019language} (see Figure \ref{fig:20_KG}) and commonsense reasoning (e.g.,  \cite{rajani2019explain},  \cite{wang2024commonsensevis}) without additional fine-tuning on task-specific data. In these tasks, prompts are commonly presented as fill in the blanks or multi-choice selection options, although some queries may require the language model to generate its own response. however, Jiang et al.  \cite{jiang2020how} show that language models are characteristically sensitive to the phrasing of the query text, and may often fail to produce the correct response even when they possess the knowledge being tested. This often happens when the question posed is presented differently from the context in which the training data captures the given concept. Therefore, the authors  \cite{jiang2020how} posit that the knowledge elicited by language models through their responses to user queries must be seen as a lower bound on the intrinsic knowledge they typically encode. The authors then propose to raise this bound by providing prompts in a variety of forms through automatically mining and re-paraphrasing queries.

Similarly, AutoPrompt  \cite{shin2020autoprompt} automatically generates suitable prompts by combining the original inputs with so-called \textit{trigger tokens} which are found by gradient-based search method. These prompting strategies have been shown to significantly improve the responses of language models to test queries. Besides prompt design, another potential challenge of data-based probing is the tendency for the queries to fail to test the underlying knowledge of the model. Specifically, language models may exhibit shortcut learning on the pre-training data, a situation where the model directly observes patterns relevant to the target linguistic property from data regularities without having to learn the requisite linguistic structure that underlies these patterns.  

\subsubsection{Mechanistic interpretability of LLM}

 Mechanistic interpretability methods provide insights into the internal reasoning mechanisms of deep neural networks models by analyzing information at network components like layers, neurons and their interconnections. The idea is to reveal, at a global level, the underlying computations and information processing mechanisms responsible for model behavior. Mechanistic interpretability is typically realized through understanding simpler components such as features and neurons.

\paragraph{Features, neurons and circuits}

In mechanistic interpretability, a feature is often regarded as the most basic piece of information that a neural network can represent. In general, features in neural networks are understood to represent different levels of abstractions of the input, although they may also represent some intermediate abstractions not directly related to input attributes. In computer vision, networks can encode features at different granularities, from low-level features such as contours and edges to high-level features like whole object parts (e.g. ears, tails). In natural language processing, features can come in the form of words or tokens to phrases and whole sentences. There are enormous efforts aimed to fully understand the relationship between features and neurons owing to the central role these concepts play in explaining neural network behaviors. Two plausible hypotheses are the monosemantic and polysemantic views of features and neurons. The association is monosemantic if a single semantic feature is encoded by a single neuron in the network. Conversely, polysemanticity implies that a particular neuron may represent multiple features. The second hypothesis (i.e., polysemanticity) is well supported by observations from multiple studies (e.g.,  \cite{elhage2022toy},  \cite{lecomtewhat},  \cite{olah2020zoom}).

It has been suggested  \cite{elhage2022toy} that the necessity of polysemanticity could stem from the fact that the possible number of features can be significantly larger than the total number of neurons used to represent them, hence one-to-one mapping is impossible to achieve. As to how these units combine to capture complex pieces of information, the superposition hypothesis  \cite{elhage2022toy},  \cite{nanda2023emergent} posits that the internal representation of features involves linear combinations of neurons in an overlapping manner (i.e., a particular neuron may be combined with different sets of other neurons to represent different features). However, empirical results by Engels et al.  \cite{engels2024not} on state-of-the-art large language  models provides compelling evidence that shows that features can be represented in a non-linear combination of neurons. These combinations form circuits which encode features at a higher level.

\paragraph{Mechanistic interpretability via circuit discovery}

While individual neurons are important in highlighting the relevant concepts captured by large language models, they cannot directly provide explanation on the predictions or the reasoning processes of these models. Therefore, groups of these neural elements, called \textit{circuits}  \cite{olah2020zoom}, have been utilized as building blocks for understanding model behavior. In this context, a circuit refers to a fixed computational unit in a network comprising a set of neural activations and connecting pathways that can collectively process features. Essentially, the network can be viewed as a computational graph whereas circuits represent subgraphs.

 A fundamental approach in realizing mechanistic interpretability for LLMs, circuit discovery  \cite{olah2020zoom,elhage2021mathematical}, is concerned with identifying meaningful circuits in the neural network which are responsible for a given task performance. A popular line of work on circuit discovery utilizes the so-called \textit{activation patching}  \cite{meng2022locating} technique. Activation patching is a method of understanding the functions of smaller subgraphs in a network by modifying specific activations so as to understand their impact on model behavior or task performance. Thus, the technique is essentially a global counterfactual explainer in activation space. 

\paragraph{Automatic circuit discovery techniques}

Meanwhile, vanilla circuit discovery approaches require circuits to be identified manually and are therefore impractical to deploy for practical cases other than toy models or examples. Consequently, automated circuit discovery techniques are proposed to mitigate this challenge by automating various stages in the mechanistic interpretability pipeline.  ACDC  \cite{conmy2023towards} reduces the manual work in the activation patching procedure by automatically iterating through nodes and pruning redundant edges that do not impact performance on the given task as determined by a specified metric. By recursively applying this method, the procedure helps to discover sparse subgraphs that are representative of the entire network. ACDC uses naive search for edges to prune, a process which can be computationally prohibitive for large networks. Second, activation patching itself is slow as it requires three runs for each input. Consequently, this approach is limited to application settings involving small model sizes and fewer sets of behaviors. Syed et al.  \cite{syed2023attribution} propose to overcome this challenge by utilizing  \textit{attribution patching}  \cite{nanda2023attribution}, a method which employs gradient-based approximation of activation patching to reduce the computational overhead but it sacrifices performance.  

\paragraph{Other mechanistic interpretability techniques}

Other popular lines of work in mechanistic interpretability approaches are based on sparse autoencoders  \cite{cunningham2023sparse},  \cite{bricken2023towards} and \textit{logit lens} \cite{geva2022transformer}, \cite{belrose2023eliciting}. The first group of methods leverage sparse autoencoders to learn sparse sets of activations for decomposing superposed features arising from polysemanticity. Logit lens attempts to track the evolution of information across layers by progressively projecting the intermediate activations into the vocabulary space or logits. This method provides a view into (or ``lens" for viewing) the internal decision-making process of the model. 

\paragraph{Benefits of mechanistic interpretability}

Mechanistic interpretability is a unique class of methods of providing insights into the internal states of large neural networks. To achieve this goal, developers attempt to reconstruct, in discernable algorithms, the computations that these models use for processing data to arrive at their decisions. With mechanistic interpretability methods, a large neural network can be decomposed into simpler components or circuits where the role of each part can independently be studied. This compositionality approach can be immensely useful as it allows complex behaviors to be understood from simpler ones. Also, it could pave the way for not only a higher level of understanding, but also a higher control over the decision process. For instance, techniques like activation patching, which provide interpretability by modifying activations could also be used to ``switch off" circuits that drive undesirable behaviors like hallucination. Similarly, other approaches, like logit lens which allows the sequence of computations through neural networks to be analyzed, could also facilitate valuable understanding of the ``thought process" underlying complex decisions of large language models.

\section{Explainable AI using vision-language and large language models}

Owing to the enormous power of vision-language models (VLMs) and large language models (LLMs) across a vast array tasks and domains, the potential of leveraging these foundation models to enhance the explainability of other machine language models is currently attracting research interest.  Three important roles of VLMs, particularly CLIP  \cite{zhang2021learning}, and LLMs in this use case have particularly received significant attention: 1) use of vision-language models for improving or correcting the explanations generated by other methods; 2) use of large language models in explaining explanations (i.e., via translating the interpretations generated by other methods into more intelligible natural language explanations); and 3) leveraging VLMs and large LLMs to automate concept discovery in concept-based learning frameworks. We discuss each of this use cases in the following sub-sections.

\subsection{Improving gradient-based interpretability with VLMs}

One of the main functions of vision-language models in explainable AI is to refine the explanations of other methods. For example, class activation mapping techniques such as CAM and Grad-CAM do not provide accurate localization information. Despite their popularity, multiple empirical studies have outlined the most important limitations of this class of interpretability methods, namely, their problem of 1) partial activation, i.e., the tendency to activate or highlight only the most discriminative parts of the target rather than all relevant pixels; and 2) false activation of background pixels (i.e., irrelevant backgrounds objects are often highlighted). The commonest techniques employed to address these problems usually involve the following stages: 1) training the model on the main classification task; 2) generating initial class activation maps to provide coarse localization information; and 3) refining class activation maps with additional localization information from the vision-language model. 

Most of the studies discussed in this subsection were originally presented in the context of weakly supervised semantic segmentation. However. Since they essentially solve the unsupervised pixel classification problem, they can be regarded as interpretability methods for image classification. 

Prominent among this line of work includes LICO  \cite{lei2023lico} which  \uline{ }improves gradient-based interpretability methods through language supervision from CLIP by learning text prompts with corresponding image features. The method proposed in the study first extracts language features using CLIP with frozen weights. Subsequently, a loss function is defined to minimize the distances between image and language manifolds. And, to ensure that the most relevant features of a class have the highest similarity to the language prompt for the given class, the authors propose to\uline{ }align the linguistic tokens and image feature maps using optimal transport theory. Empirical results by the authors show that while retaining competitive classification accuracy, LICO achieves favorable interpretability performance with common interpretability methods like Grad-CAM, Score-CAM and Group-CAM.

\begin{figure}[H]
 	\vspace {-1mm}
 	\centering
 	\includegraphics[width=1.0 \linewidth]{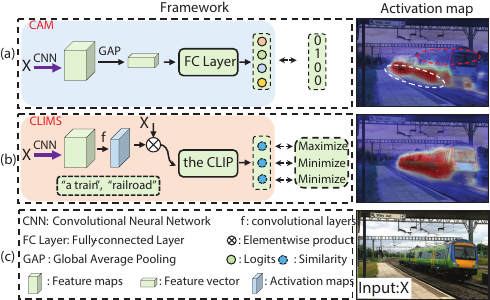} \vspace {-3mm}
 	\caption{(a) saliency maps produced by the standard CAM technique is characterized by imprecise activations (RIGHT) and these are obtained from pooled average features at the GAP layer (LEFT); (b) CLIMS replaces the GAP and FC layers with additional convolution operations which are supervise by CLIP (LEFT) to generate more accurate saliency maps that highlight correct target regions (RIGHT); (c) meanings of abbreviations, symbols and illustrations used in (a) and (b). 
 	}\label{fig:21_SaliencyMaps}
 \end{figure}

 Similarly, in CLIMS  \cite{xie2022clims}, Xie leverages the multimodal vision and language understanding of CLIP to activate whole target regions and suppress background pixels. CLIMS relies on the original neural network backbone for predicting the initial class activation maps. Then, CLIP  \cite{zhang2021learning} is used to refine the generated CAMs using the knowledge of image content and textual descriptions or labels. It ensures complete target activation by maximizing the foreground and text label matching loss. It also suppresses redundant activations by minimizing background and text label matching loss. Additionally, co-occurring bias is a phenomenon that causes activation maps of classifiers to wrongly focus on objects that frequently occur together with the target in natural images. To mitigate this problem, CLIMS incorporate a co-occurring region suppression loss which is minimized to remove these superfluous regions from the generated CAMs. Figure \ref{fig:21_SaliencyMaps} (a) illustrates CAM-generated irrelevant attributions for the class ``train" shown by the dash white ellipse (possibly due to the co-occurrence of trains and railroads in practice). In Figure \ref{fig:21_SaliencyMaps} (b), CLIMS produces saliency map which is more focused on the target train.

  While CLIMS employs CLIP to refine already generated class activation maps by using a separate neural network, in CLIP-ES  \cite{lin2023clip}Lin  et al. propose to simplify the architecture by directly leveraging CLIP in the pipeline to compute and improve final CAM scores. The authors utilize the Softmax loss to optimize Grad-CAM with the aid of CLIP to remove superfluous, non-target classes from the activations. The approach vastly improved on CLIMS in terms of total computational time and memory required to produce the final heatmap.

\subsection{Explaining explanations using LLMs}

Owing to their rich knowledge base and the ability to articulate contextually relevant descriptions of complex phenomena in a human-understandable, natural language form, LLMs have a natural appeal for explaining the predictions of machine learning models. The approach has received enormous research interest in practical domains, including in high-stakes applications like cybersecurity  \cite{ali2023huntgpt},  \cite{khedirienhancing}. While attribution methods like Gradients or SHAP can present their results in various forms such as visualization plots to aid human understanding, such outputs may only be meaningful to machine learning experts or developers but not end-users who do not usually possess the requisite knowledge of the machine learning systems or the interpretability methods. For this reason, stakeholders cannot fully exploit the intended benefits of explainable AI. Consequently, to enhance the intelligibility and utility of the generated explanations, multiple studies have explored the use of large language models as a means to translate these outputs to into human-understandable natural language explanations. TalkToModel \cite{slack2023explaining},InterroLang   \cite{feldhus2023interrolang} and XAIstories  \cite{martens2023tell} are examples of promising works in this direction.

TalkToModel uses the importance scores of attribution-based methods like SHAP to explain machine learning model predictions to end-users in an interactive dialog. The framework leverages large language models to parse the user’s words and to generate natural language responses to explain the predictions. This method is suitable for feature attributions on tabular data and classifiers based on this data type, as well as other methods like counterfactual explanations. Similarly, Castelnovo et al.  \cite{castelnovo2024augmenting}, QoEXplainer  \cite{wehner2024qoexplainer} and  XAIstories  \cite{martens2023tell} propose  to generate natural language explanations from Shapley scores through human-computer interactions.  XAIstories includes separate modules for counterfactual explanations and direct natural language explanations. The modules are leveraged toexplain the importance of features based on SHAP generated attributions. InterroLang extends TalkToModel to explain NLP models. Likewise, Zytek et al.  \cite{zytek2024llms} evaluate the suitability of GPT-based models in converting SHAP-generated explanations into human-readable natural language explanations. The authors propose prompt designs to enable the LLM to generate accurate and contextually relevant explanations. To accomplish this, GTP-3.5 or -4 is used to generate explanations for the most relevant features ranked by their SHAP scores. The study evaluates the LLM generated explanations on their soundness, clarity, completeness, and context-awareness. In a user, experiment participants show overwhelming preference for LLM-based explanations over the original attributions, citing their clarity, informativeness and overall usability as factors for this choice. These results highlight the potential of LLMs in improving the explainability of conventional attribution-based methods like SHAP.

In a practical use case in cybersecurity  \cite{ali2023huntgpt},  \cite{khedirienhancing}, LLM integration allows  operators to readily understand the generated explanations in order to quickly assess the underlying threats behind the generated alerts. Since LIME and SHAP scores present prediction interpretation in visualization plots indicating how different features influence the model's decision for a given prediction, these methods do not directly convey their explanations in human-understandable form. Therefore, the final results still need to be interpreted by operators with some level of understanding of these results. Additionally, incorporating LLMs also facilitates human-machine interactions. For example, HuntGPT  \cite{ali2023huntgpt} convert LIME or Shapley scores into LLM-powered natural language explanations in an interactive platform where operators can ask follow-up questions about the predicted threat or possible intervention measures. Similarly, Zhang and Chen  \cite{zhang2024large} utilize SHAP scores to quantify the contributions of various factors in a model predictive control of an HVAC (heating, ventilation, and air conditioning) system, and then leverage LLM to generate human-understandable explanations. Zhang and Chen also demonstrate the ability of their LLM-based explainable AI system to incorporate interactive features through question answering. This feature improves trust and understanding of the model’s decisions, and enables system operators to gain further insight on the use case. 

\subsection{Improving concept-based learning with VLM and LLM}

Conventional concept-based reasoning methods face significant challenges both in terms of implementation and performance. Notably, (i) the human effort and time required to produce hand-crafted concept labels can be prohibitive (ii) the need to enforce similarity constraints with human-interpretable concepts may result in degraded performance compared to equivalent black-box models. In fact, the first problem can even aggravate the second. Specifically, for high explainability and performance, concept annotations must be carefully crafted.  However, the tremendous difficulty in labeling large datasets with fine-grained class-discriminative concepts can lead to sub-optimal concept design and consequently poor performance.  

Fortunately, recent studies such as Post-hoc Concept Bottleneck Models (Post-Hoc CBM or P-CBM)  \cite{yuksekgonul2022post} are  proposed to simultaneously address both of these limitations by leveraging the impressive generalization capability of the CLIP vision-language model, particularly, its ability to discover visual concepts (see, for example,  \cite{zang2024pre}) from natural images.  Post-Hoc CBM solves the first problem by automating the creation of concept annotations and addresses the second problem by adopting a post-hoc scheme for concept learning, rather than the conventional ante-hoc approach. Post-Hoc CBM leverages CLIP to generate human-interpretable concepts using concept descriptions from ConceptNet  \cite{speer2013conceptnet}. ConceptNet is a well-known knowledge graph that encodes a vast amount of world knowledge. Also, Post-Hoc CBM can encode labeled concepts features using concept activation vectors (CAVs). To ensue acceptable performance, P-CBM appends a residual predictor module (i.e., after the final layer of the original black-box model) whose role is to recover the black-box model’s predictions when the learned concepts are inadequate to achieve satisfactory results.  

   Post-hoc concept bottleneck models also facilitates additional performance improvements at test time by allowing human experts to debug and refine incorrect concept attributions on a global level, rather than restricting human interventions to individual predictions, as is the case with conventional CBM frameworks models\textit{ }  \cite{koh2020concept}.  Although Post-hoc CBM achieves an impressive performance, its reliance on static knowledge graphs to build the concept subspace constrains its use to application domains where these knowledge bases exist.   Unfortunately, owing to the limited coverage of knowledge graphs, these use cases are very restrictive. Furthermore, as a post-hoc explainability framework which essentially learns CLIP-generated concepts on top of an already trained network, Post-hoc CBM’s explanations may not accurately reflect the original model’s reasoning. This faithfulness problem is a general concern for post-hoc interpretability methods.

 To mitigate the potential drop in faithfulness, Concept Discovery Models (CDMs)  \cite{panousis2023sparse} avoid training a separate concept bottleneck layer in a post-hoc scheme, and instead propose to infer the presence of relevant concepts from CLIP’s image and text embeddings using variational Bayesian technique  \cite{kingma2013auto} during model training . The entire pipeline, which includes prediction performance optimization, concept discovery, sparsification and alignment, is realized at training time in an end-to-end fashion. CDM achieves impressive performance that even surpasses Post-hoc CBMs. 

  Despite the immense performance advantage of CBMs that leverage vision-language models like CLIP, these methods are not without their challenges. Most importantly, CLIP’S ability to automatically capture fine-grained concept descriptions from natural images in domain-specific settings is limited. Also, augmenting CLIP with knowledge graphs like ConceptNet (as in the case of Post-Hoc CBM) is inadequate in most practical cases as these static knowledge bases do not encode nuanced contextual information useful for fine-grained recognition.  Above all, knowledge graphs have limited coverage and methods relying on them are restricted to applications where xx.

To address these challenge, other state-of-the-art approaches (e.g., 
\cite{menon2022visual,yang2023language,oikarinen2023label,kazmierczak2023clip}) are proposed to augment vision-language frameworks with the more expressive large language models, whose role in these frameworks is to automatically generate concept sets from human knowledge expressed via natural language explanations. The most commonly employed large language models in these applications are GPT-3 \cite{brown2020language},  GPT-3.5 
\cite{wang2022performance}, GPT-4 \cite{achiam2023gpt} and Llamma-2 \cite{touvron2023llama}. To generate relevant concepts, these large language models are prompted or asked questions requiring them to list descriptive features of the given class or name objects commonly associated with the class. They return a list of words or phrases satisfying the given queries. LLMs have a rich store of domain knowledge about many different objects and situations, and with appropriate prompting can determine which natural language concepts are correctly describe the target. 

Specifically, Oikarinen et al.  \cite{oikarinen2023label} propose   Label-Free Concept Bottleneck Model (Label-free CBM or LF-CBM) which utilizes CLIP-Dissect  \cite{oikarinen2022clip} to learn a projection from the model’s feature space into human- interpretable concept space. CLIP-Dissect is a state-of-the-art model agnostic framework that utilizes the CLIP visual foundation model to describe the semantic meanings of network neurons in terms of high-level concepts. LF-CBM subsequently uses the learned embeddings to train a Concept Bottleneck Layer (CBL). Rather than obtaining concepts from knowledge graphs, LF-CBM leverage GPT-3 to generate the concept sets by supplying appropriate prompts. This ensures that a large, diverse and semantically-rich collection of concepts are obtained.  To enhance explainability and reduce computational burden, the authors further apply filtering to remove redundant or undesirable concepts. Similarly, In CLIP-QDA, Kazmierczak et al.  \cite{kazmierczak2023clip} utilized  CLIP to automatically recover relevant concepts from images in zero-shot manner without the need for prior manual annotation. Initial concept labels are obtained by prompting GPT-3 to list a set of descriptive attributes of the images to represent relevant concepts for a class. Given an input image, The method first computes CLIP scores for each concept. Then a classifier is trained to predict the class of the image based on the detected concepts. CLIP-LIME follows the conventional LIME approach by training a surrogate model to approximate the original classifier in the neighborhood of the relevant CLIP scores by perturbing the input. Rather than presenting raw concept scores, CLIP-LIME presents its explanations as attribution weights for each concept. Similarly, CLIP-SHAP utilizes KernelSHAP  \cite{lundberg2017a} to compute attribution score for each concept by perturbing the input, and the resulting Shapley values represent the importance of each concept in the model’s decision. CLIP-LIME and CLIP-SHAP provide a more user-friendly presentation, highlighting the relative influence of all concepts in a graphical form. The method is especially useful in settings involving multiple concepts, as it allows easy visualization and comparison of concepts based on their strengths.

\begin{figure*}[!htb]
	\vspace {-1mm}
	\centering
	\includegraphics[width=1.0 \linewidth]{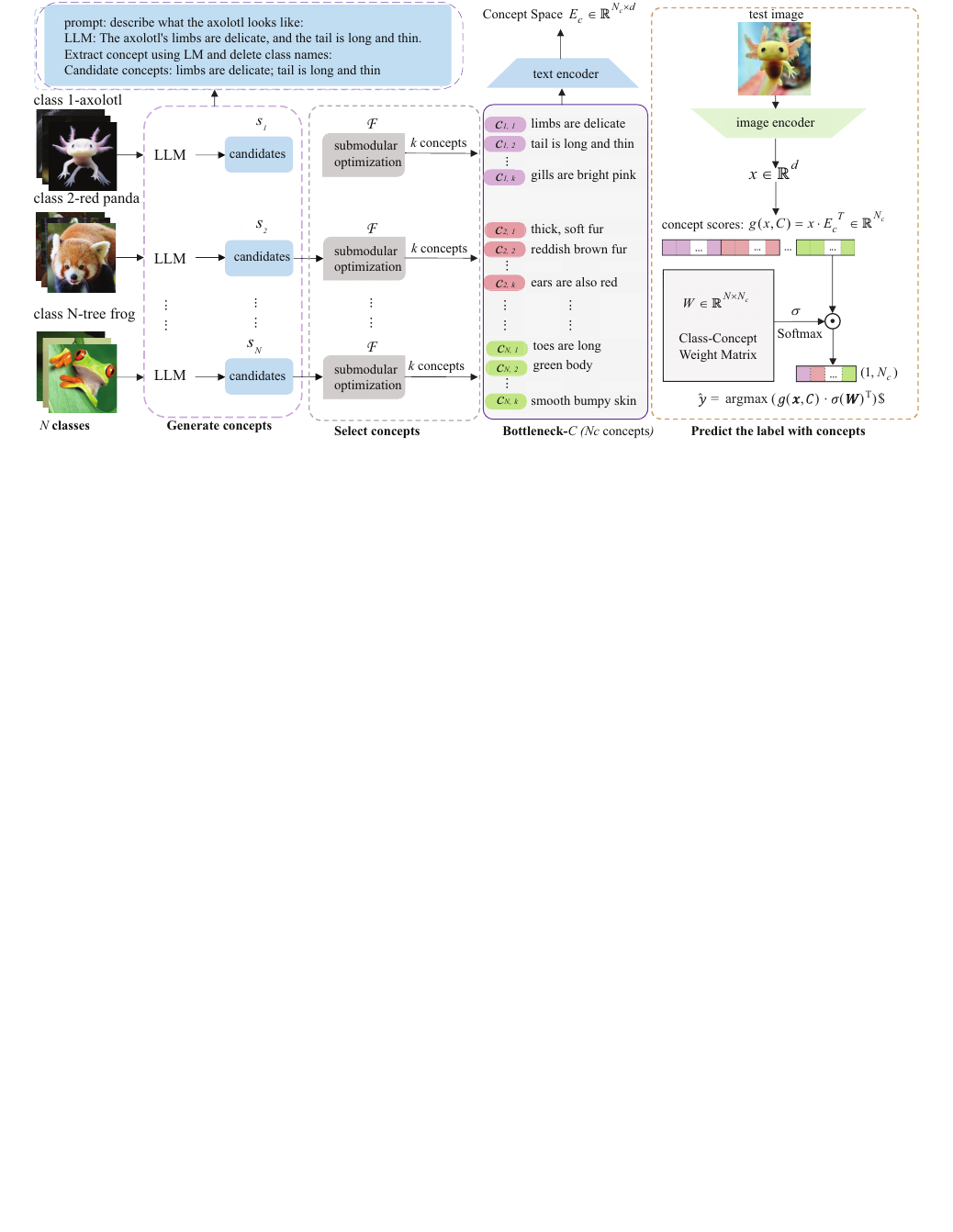} \vspace {-140mm}
	\caption{Principle of the LaBo \cite{yang2023language} framework.  
	}\label{fig:22_LaBo}
\end{figure*}

Another unique philosophy, exemplified by LaBo  \cite{yang2023language}, involves the use of very large numbers of concepts to achieve high accuracy.  In LaBo, Yang et al. propose to prompt GPT-3 to generate precise descriptions of the given categories. Then, class names are removed from the descriptions leaving only descriptive attributes used to form candidate concepts. Next, a specialized module is employed to select a quality discriminative subset from the generated concept pool. Further, a pre-trained CLIP model with frozen weights is used to align the selected GPT-3-generated linguistic concepts and visual attributes from CLIP embeddings. The resulting concept scores are used to construct a bottleneck layer. Finally, the classification task is trained to associate image-text concept scores with target category labels. The framework is illustrated in Figure \ref{fig:22_LaBo}. \textbf{ }Although LaBo achieves high prediction accuracy, the large number of concepts employed (as high as 10, 000 for ImageNet dataset) makes the decisions difficult to explain. This challenge arises because humans cannot intuitively keep track of the high number of concepts that are associated with each classification decision.   Furthermore, recent studies (e.g., LM4CV  \cite{yan2023learning}) show that such large vocabularies generated from LLMs are characterized by redundancies and excessive noise that may include concepts that are irrelevant to the target category or even non-visual attributes such as "loud music". The study even shows that a set of concepts randomly drawn from the English vocabulary can achieve competitive performance with standard LLM-generated baselines, implying that most of the generated concepts are not unique to the target classes and hence cannot be considered class-discriminative. This ultimately undermines the faithfulness of explanations derived by associating these noisy linguistic features with physical attributes of targets in images. 

  It is possible to limit the number of concepts by enforcing sparsity constraint at a final layer following the concept bottleneck layer (e.g., Label-free CBM). However, as this method primarily caps the desired number of concepts on a global per-class level and not the semantic importance of each concept to the given example, it can lead to sub-optimal results in terms of the quality of concepts retained. Specifically, irrelevant concepts for certain classes may be admitted at the expense of some important concepts for other classes. In order to overcome this drawback, more recent studies employ techniques designed to select the most important concepts, thus allowing them to achieve sparsity without severely sacrificing relevant concepts. For instance, CDM  \cite{panousis2023sparse} achieves  sparsity by inferring relevant concepts with the aid of Bayesian technique. LM4CV \cite{yan2023learning} leverage dictionary learning to extract concise attributes from CLIP embeddings which help to prune the raw attributes generated by the LLM. VLG-CBM \cite{srivastava2024vlg} leverage the  GroundingDINO  \cite{liugrounding} open- vocabulary object detection framework to find visual concepts relevant to target categories. GroundingDINO is a state-of-the-art model that is capable of detecting arbitrary objects from their natural language descriptions.    

\subsection{Comparison of LLM-assisted concept learning methods}

\subsubsection{Datasets}

To compare the results of different methods, we summarize results from empirical studies from state-of-the-art research in the field. The review considers results on publicly available datasets with distinct characteristics. The selected datasets are the CIFAR-10 and CIFAR-100  \cite{krizhevsky2009learning}, the Caltech-UCSD Birds-200 (CUB)  \cite{wah2011},  ImageNet  \cite{deng2009imagenet} and Places365  \cite{zhou2018places}. These are among the most widely used datasets for testing the efficacy of concept-based explainability frameworks.  The datasets vary widely in terms of size and complexity. The CIFAR-10 and CIFAR-100 datasets each comprises low resolution (32x32) images of common objects and animals. There is one target per image, and the images are cropped to reveal these target objects. CUB is a large dataset of bird photographs grouped into 200 classes. In comparison, Places365 and ImageNet comprise 365 and 1000 classes, and each contain over a million images. Table \ref{tab:4summary_datasets} summarizes the datasets.

\begin{table*}[htb!]
	\caption{Summary of datasets commonly employed in evaluating CBMs. $\ast$ -different versions have different number of images and categories; + Places365 -the standard version contains 365 classes.}
	\label{tab:4summary_datasets}
\renewcommand{\arraystretch}{1.3}
\begin{adjustbox}{max width=\textwidth}
	
		\scalebox{0.95}{
			
\begin{tabular}{p{4.12cm}p{4.12cm}p{4.12cm}}
\hline
\multicolumn{1}{|p{4.12cm}}{\textbf{Dataset}} & 
\multicolumn{1}{|p{4.12cm}}{\textbf{Number of images}} & 
\multicolumn{1}{|p{4.12cm}|}{\textbf{Number of classes}} \\ 
\hline
\multicolumn{1}{|p{4.12cm}}{CIFAR-10 \cite{krizhevsky2009learning}} & 
\multicolumn{1}{|p{4.12cm}}{60, 000} & 
\multicolumn{1}{|p{4.12cm}|}{10} \\ 
\hline
\multicolumn{1}{|p{4.12cm}}{CIFAR-100 \cite{krizhevsky2009learning}} & 
\multicolumn{1}{|p{4.12cm}}{60, 000} & 
\multicolumn{1}{|p{4.12cm}|}{100} \\ 
\hline
\multicolumn{1}{|p{4.12cm}}{CUB \cite{wah2011caltech}} & 
\multicolumn{1}{|p{4.12cm}}{11, 788} & 
\multicolumn{1}{|p{4.12cm}|}{200} \\ 
\hline
\multicolumn{1}{|p{4.12cm}}{ImageNet$\ast$ \cite{deng2009imagenet} } & 
\multicolumn{1}{|p{4.12cm}}{14, 197, 122/ 1,431,167} & 
\multicolumn{1}{|p{4.12cm}|}{21,841/1000} \\ 
\hline
\multicolumn{1}{|p{4.12cm}}{Places365$\ast$\textsuperscript{+} \cite{zhou2017places}} & 
\multicolumn{1}{|p{4.12cm}}{10,624,92/ 2,150,210} & 
\multicolumn{1}{|p{4.12cm}|}{434/365} \\ 
\hline
\end{tabular}

} 
\end{adjustbox}
\end{table*}
\subsubsection{Comparison results}

The top-1 accuracies of various methods are presented in Table \ref{tab:5comparison_sota}. The first row shows the results for the ``standard" baseline, a conventional CNN framework trained without concept learning. The model uses a ResNet50 backbone trained on the ImageNet dataset. This network is a black-box model without additional accuracy-limiting constraints imposed, as in the case with CBMs.  Hence, the model achieves state-of-the-art performance across all datasets, with top-1 accuracy as high as 88.80$\%$ on CIFAR-10 and 76.13$\%$ on ImageNet. In contrast, CB2M  \cite{steinmann2023learning} which  is a representative state-of-the-art (SoTA) concept bottleneck model that does not leverage the power of vision-language or large language models, achieves a much lower accuracy when tested on the CUB dataset. CB2M is an example of a case-based learning model that can be improved through user intervention at test time. However, owing to interpretability constraints, the vanilla framework performs far worse than the standard baseline. Specifically, the accuracy on the CUB dataset dramatically drops to 69.1$\%$ compared to 76.70$\%$ for the ``standard" model. Hard-AR  \cite{havasi2022addressing} addresses concept correlation and leakage in CBMs and achieves state-of-the-art performance on CUB. However, Hard-AR, like CB2M, does not utilize foundation models (VLM or LLM) to automate concept annotation. For this reason, it is difficult to apply the method to large or complex datasets like ImageNet that contain thousands of classes. Consequently, these methods have only been applied to domains in which labeled concepts exist (e.g., CUB) and small-scale or low-dimensional datasets like MIMIC-III EWS \cite{johnson2016mimic}  and colorMNIST \cite{lecun1998mnist,rieger2020interpretations}. MIMIC-III EWS is an\textbf{ }Early Warning Score (EWS) dataset from electronic health records that is used for the prediction of cardiac conditions. MNIST and colorMNIST are simple handwritten digit recognition datasets.

The top-1 accuracies of various methods are presented in Table \ref{tab:5comparison_sota}. The first row shows the results for the ``standard" baseline, a conventional CNN framework trained without concept learning. The model uses a ResNet50 backbone trained on the ImageNet dataset. This network is a black-box model without additional accuracy-limiting constraints imposed, as in the case with CBMs.  Hence, the model achieves state-of-the-art performance across all datasets, with top-1 accuracy as high as 88.80$\%$ on CIFAR-10 and 76.13$\%$ on ImageNet. In contrast, CB2M  \cite{steinmann2023learning} which  is a representative state-of-the-art (SoTA) concept bottleneck model that does not leverage the power of vision-language or large language models, achieves a much lower accuracy when tested on the CUB dataset. CB2M is an example of a case-based learning model that can be improved through user intervention at test time. However, owing to interpretability constraints, the vanilla framework performs far worse than the standard baseline. Specifically, the accuracy on the CUB dataset dramatically drops to 69.1$\%$ compared to 76.70$\%$ for the ``standard" model. Hard-AR  \cite{havasi2022addressing} addresses concept correlation and leakage in CBMs and achieves state-of-the-art performance on CUB. However, Hard-AR, like CB2M, does not utilize foundation models (VLM or LLM) to automate concept annotation. For this reason, it is difficult to apply the method to large or complex datasets like ImageNet that contain thousands of classes. Consequently, these methods have only been applied to domains in which labeled concepts exist (e.g., CUB) and small-scale or low-dimensional datasets like MIMIC-III EWS \cite{johnson2016mimic} and colorMNIST \cite{lecun1998mnist,rieger2020interpretations}. MIMIC-III EWS is an \textit{Early Warning Score} (EWS) dataset from electronic health records that is used for the prediction of cardiac conditions. MNIST and colorMNIST are simple handwritten digit recognition datasets.

\begin{table*}[!htbp]
	
	\caption{Comparison of SoTA methods. Note that methods that use the more powerful backbones (e.g., the CLIP ViT variants) generally achieve higher accuracies than their ResNet counterparts. RN50/18 represent ResNet varients.}
	\label{tab:5comparison_sota}
	
	\renewcommand{\arraystretch}{1.3}
	\begin{adjustbox}{max width=\textwidth}
		
			\scalebox{0.80}{
		\begin{tabular}{p{2.86cm}p{1.7cm}p{1.7cm}p{1.82cm}p{1.92cm}p{1.84cm}p{2.97cm}p{1.69cm}}
			\hline
			\multicolumn{1}{|p{2.86cm}}{\textbf{Model}} & 
			\multicolumn{1}{|p{1.7cm}}{\textbf{CIFAR-10}} & 
			\multicolumn{1}{|p{1.7cm}}{\textbf{CIFAR-100}} & 
			\multicolumn{1}{|p{1.82cm}}{\textbf{CUB-200}} & 
			\multicolumn{1}{|p{1.92cm}}{\textbf{Places365}} & 
			\multicolumn{1}{|p{1.84cm}}{\textbf{ImageNet}} & 
			\multicolumn{1}{|p{2.97cm}}{\textbf{Back-b}} & 
			\multicolumn{1}{|p{1.69cm}|}{\textbf{Metric}} \\ 
			\hline
			\multicolumn{1}{|p{2.86cm}}{Standard } & 
			\multicolumn{1}{|p{1.7cm}}{88.80$\%$} & 
			\multicolumn{1}{|p{1.7cm}}{ 70.10$\%$$\ast$ } & 
			\multicolumn{1}{|p{1.82cm}}{76.70$\%$ } & 
			\multicolumn{1}{|p{1.92cm}}{48.56$\%$ } & 
			\multicolumn{1}{|p{1.84cm}}{76.13$\%$ } & 
			\multicolumn{1}{|p{2.97cm}}{ResNet50} & 
			\multicolumn{1}{|p{1.69cm}|}{} \\ 
			\hline
			\multicolumn{1}{|p{2.86cm}}{CB2M \cite{steinmann2023learning}} & 
			\multicolumn{1}{|p{1.7cm}}{} & 
			\multicolumn{1}{|p{1.7cm}}{} & 
			\multicolumn{1}{|p{1.82cm}}{69.1} & 
			\multicolumn{1}{|p{1.92cm}}{} & 
			\multicolumn{1}{|p{1.84cm}}{} & 
			\multicolumn{1}{|p{2.97cm}}{Inception-v3} & 
			\multicolumn{1}{|p{1.69cm}|}{} \\ 
			\hline
			\multicolumn{1}{|p{2.86cm}}{Hard-AR \cite{havasi2022addressing}} & 
			\multicolumn{1}{|p{1.7cm}}{N/A } & 
			\multicolumn{1}{|p{1.7cm}}{N/A } & 
			\multicolumn{1}{|p{1.82cm}}{81.7} & 
			\multicolumn{1}{|p{1.92cm}}{N/A } & 
			\multicolumn{1}{|p{1.84cm}}{N/A } & 
			\multicolumn{1}{|p{2.97cm}}{Inception v3 } & 
			\multicolumn{1}{|p{1.69cm}|}{} \\ 
			\hline
			\multicolumn{1}{|p{2.86cm}}{P-CBM \cite{yuksekgonul2022post}} & 
			\multicolumn{1}{|p{1.7cm}}{84.50$\%$$\ast$} & 
			\multicolumn{1}{|p{1.7cm}}{56.00$\%$$\ast$} & 
			\multicolumn{1}{|p{1.82cm}}{N/A } & 
			\multicolumn{1}{|p{1.92cm}}{N/A } & 
			\multicolumn{1}{|p{1.84cm}}{N/A } & 
			\multicolumn{1}{|p{2.97cm}}{CLIP(RN50)} & 
			\multicolumn{1}{|p{1.69cm}|}{} \\ 
			\hline
			\multicolumn{1}{|p{2.86cm}}{LF CBM \cite{oikarinen2023label}} & 
			\multicolumn{1}{|p{1.7cm}}{86.40$\%$} & 
			\multicolumn{1}{|p{1.7cm}}{65.13$\%$} & 
			\multicolumn{1}{|p{1.82cm}}{74.31$\%$} & 
			\multicolumn{1}{|p{1.92cm}}{43.68$\%$} & 
			\multicolumn{1}{|p{1.84cm}}{71.95$\%$} & 
			\multicolumn{1}{|p{2.97cm}}{CLIP(RN50) /RN18} & 
			\multicolumn{1}{|p{1.69cm}|}{top-1} \\ 
			\hline
			\multicolumn{1}{|p{2.86cm}}{} & 
			\multicolumn{1}{|p{1.7cm}}{} & 
			\multicolumn{1}{|p{1.7cm}}{} & 
			\multicolumn{1}{|p{1.82cm}}{} & 
			\multicolumn{1}{|p{1.92cm}}{} & 
			\multicolumn{1}{|p{1.84cm}}{} & 
			\multicolumn{1}{|p{2.97cm}}{} & 
			\multicolumn{1}{|p{1.69cm}|}{} \\ 
			\hline
			\multicolumn{1}{|p{2.86cm}}{CDL \cite{zang2024pre}} & 
			\multicolumn{1}{|p{1.7cm}}{96.5} & 
			\multicolumn{1}{|p{1.7cm}}{77.8} & 
			\multicolumn{1}{|p{1.82cm}}{64.7} & 
			\multicolumn{1}{|p{1.92cm}}{-} & 
			\multicolumn{1}{|p{1.84cm}}{75.7} & 
			\multicolumn{1}{|p{2.97cm}}{CLIP ViT-B/32} & 
			\multicolumn{1}{|p{1.69cm}|}{} \\ 
			\hline
			\multicolumn{1}{|p{2.86cm}}{CDM (RN50, w/ Z)} & 
			\multicolumn{1}{|p{1.7cm}}{86.50} & 
			\multicolumn{1}{|p{1.7cm}}{67.60} & 
			\multicolumn{1}{|p{1.82cm}}{72.26} & 
			\multicolumn{1}{|p{1.92cm}}{52.70} & 
			\multicolumn{1}{|p{1.84cm}}{72.20} & 
			\multicolumn{1}{|p{2.97cm}}{CLIP(RN50)} & 
			\multicolumn{1}{|p{1.69cm}|}{} \\ 
			\hline
			\multicolumn{1}{|p{2.86cm}}{CDM (ViT-B/16, w/ Z) } & 
			\multicolumn{1}{|p{1.7cm}}{95.30} & 
			\multicolumn{1}{|p{1.7cm}}{80.50} & 
			\multicolumn{1}{|p{1.82cm}}{79.50} & 
			\multicolumn{1}{|p{1.92cm}}{52.58} & 
			\multicolumn{1}{|p{1.84cm}}{79.30} & 
			\multicolumn{1}{|p{2.97cm}}{ViT-B/16} & 
			\multicolumn{1}{|p{1.69cm}|}{} \\ 
			\hline
			\multicolumn{1}{|p{2.86cm}}{LM4CV \cite{yan2023learning}} & 
			\multicolumn{1}{|p{1.7cm}}{87.99} & 
			\multicolumn{1}{|p{1.7cm}}{77.29} & 
			\multicolumn{1}{|p{1.82cm}}{64.05} & 
			\multicolumn{1}{|p{1.92cm}}{-} & 
			\multicolumn{1}{|p{1.84cm}}{-} & 
			\multicolumn{1}{|p{2.97cm}}{CLIP ViT-B/32} & 
			\multicolumn{1}{|p{1.69cm}|}{} \\ 
			\hline
			\multicolumn{1}{|p{2.86cm}}{VLG-CBM \cite{srivastava2024vlg}} & 
			\multicolumn{1}{|p{1.7cm}}{88.63$\%$} & 
			\multicolumn{1}{|p{1.7cm}}{66.48$\%$} & 
			\multicolumn{1}{|p{1.82cm}}{75.82$\%$} & 
			\multicolumn{1}{|p{1.92cm}}{42.55$\%$} & 
			\multicolumn{1}{|p{1.84cm}}{73.98$\%$} & 
			\multicolumn{1}{|p{2.97cm}}{CLIP(RN50) /RN18} & 
			\multicolumn{1}{|p{1.69cm}|}{} \\ 
			\hline
			\multicolumn{1}{|p{2.86cm}}{WP2 - FVLC \cite{lai2023faithful} $\ast$+} & 
			\multicolumn{1}{|p{1.7cm}}{86.22$\%$} & 
			\multicolumn{1}{|p{1.7cm}}{65.34$\%$} & 
			\multicolumn{1}{|p{1.82cm}}{74.44$\%$} & 
			\multicolumn{1}{|p{1.92cm}}{44.55 } & 
			\multicolumn{1}{|p{1.84cm}}{NA} & 
			\multicolumn{1}{|p{2.97cm}}{CLIP(RN50) /RN18} & 
			\multicolumn{1}{|p{1.69cm}|}{} \\ 
			\hline
			\multicolumn{1}{|p{2.86cm}}{CSS VL-CBM \cite{selvaraj2024improving}} & 
			\multicolumn{1}{|p{1.7cm}}{} & 
			\multicolumn{1}{|p{1.7cm}}{} & 
			\multicolumn{1}{|p{1.82cm}}{83.89} & 
			\multicolumn{1}{|p{1.92cm}}{} & 
			\multicolumn{1}{|p{1.84cm}}{} & 
			\multicolumn{1}{|p{2.97cm}}{CLIP ViT-B-16} & 
			\multicolumn{1}{|p{1.69cm}|}{} \\ 
			\hline
			\multicolumn{1}{|p{2.86cm}}{CF-CBM \cite{panousis2024coarse}} & 
			\multicolumn{1}{|p{1.7cm}}{} & 
			\multicolumn{1}{|p{1.7cm}}{} & 
			\multicolumn{1}{|p{1.82cm}}{79.50} & 
			\multicolumn{1}{|p{1.92cm}}{-} & 
			\multicolumn{1}{|p{1.84cm}}{77.40} & 
			\multicolumn{1}{|p{2.97cm}}{, ViT-B-16} & 
			\multicolumn{1}{|p{1.69cm}|}{} \\ 
			\hline
			\multicolumn{1}{|p{2.86cm}}{SPARSE-CBM \cite{semenov2024sparse}} & 
			\multicolumn{1}{|p{1.7cm}}{91.17$\%$ } & 
			\multicolumn{1}{|p{1.7cm}}{91.17$\%$ } & 
			\multicolumn{1}{|p{1.82cm}}{80.02$\%$} & 
			\multicolumn{1}{|p{1.92cm}}{41.34$\%$} & 
			\multicolumn{1}{|p{1.84cm}}{71.61$\%$ } & 
			\multicolumn{1}{|p{2.97cm}}{CLIP-ViT-L/14} & 
			\multicolumn{1}{|p{1.69cm}|}{} \\ 
			\hline
		\end{tabular}
	} 
	\end{adjustbox}
	\end{table*}

The third group of methods in Table \ref{tab:5comparison_sota} are those that automate concept discovery with the aid of vision-language and large language models. For this group of methods, the results presented in Table \ref{tab:5comparison_sota} show that most studies achieve far more impressive performance compared to the first two categories of methods (i.e., standard CNN and regular CBM). Besides the differences among these categories of methods, the results also show that performance also varies widely across datasets. Specifically, the CIFAR-10 dataset gives the best results for all methods. In contrast, the accuracies on the remaining datasets range from high (e.g., on CIFAR-100 and CUB) to modest (e.g., on Places365 and ImageNet). There are a number of reasons for this observation. Fundamentally, images in the smaller datasets are also less complex and are easier to distinguish compared to the larger datasets. This is because the larger the dataset, the higher the likelihood of the need to use more fine-grained features for recognition.   For example, on the CIFAR-100, it is much more challenging to find discriminative attributes that distinguish a ``girl"  from a ``woman" than, say a ``ship" from a ``bird" on the CIFAR-10.  Likewise, on the CUB, the task is to identify bird species using fine-grained concepts such as wing color and beak shape. Unfortunately, some of the attributes are shared among classes in the dataset. Moreover, owing to factors like partial occlusion, some images may not reveal certain concepts critical to the classification task. Despite these challenges, some methods achieve high accuracies on these images. For instance, CSS VL-CBM \cite{selvaraj2024improving}, CF-CBM \cite{panousis2024coarse} and SPARSE-CBM \cite{semenov2024sparse} achieve  substantial accuracies on CUB-200 while CF-CBM also excels on ImageNet.  CDM \cite{panousis2023sparse}with  ViT-B-16 backbone maintains a consistently high accuracy across all the complex datasets (CUB-200, ImageNet Places365). All these frameworks mentioned outperform the Standard black-box baseline on the respective datasets by a wide margin. This means that, besides facilitating interpretability, leveraging VLMs and LLMs in case-based learning can help to improve accuracy on difficult classification tasks. The underlining principle is that grounding object recognition on known visual attributes can help to significantly improve performance. Overall, the Places365 scene recognition is by far the most difficult among all the tasks presented in Table \ref{tab:5comparison_sota}. This task involves classifying whole scenes based on the objects present. Given the high number of scene categories in the dataset, and the fact that these scenes overlap widely in the kind of objects they contain, it is difficult to find representative concepts that exclusively describe each category.

\begin{figure}[H]
	\vspace {-4mm}
	\centering
	\includegraphics[width=1.0 \linewidth]{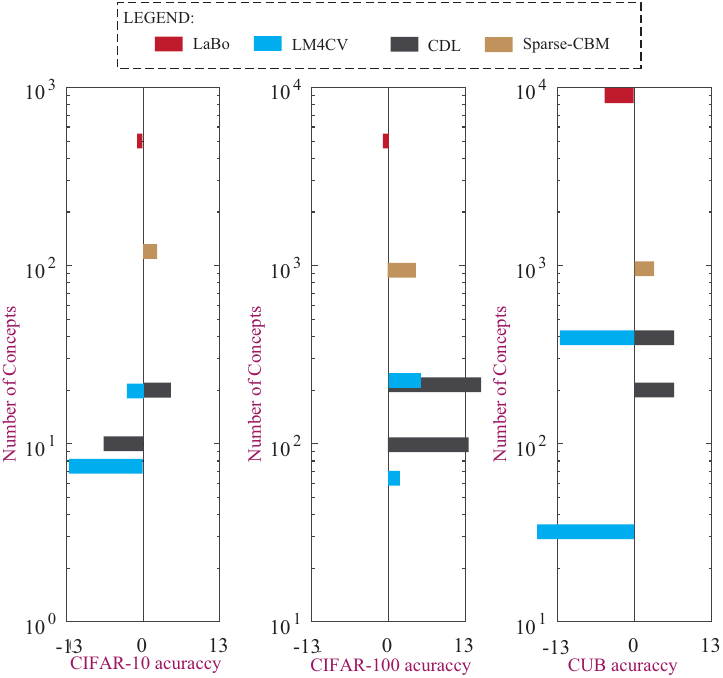} \vspace {-3mm}
	\caption{Relationship between number of concepts used and accuracy of CBMs that utilize VLMs and LLMs. The horizontal bars accuracies with respect to a standard ResNet baseline.
	}\label{fig:23_Acc_LLM_VLM}
\end{figure}

Finally, as shown in Figure \ref{fig:23_Acc_LLM_VLM}, for a given model, classification accuracy correlates positively with number of concepts. However, this does not seem to be the case for different methods. That is, a method is likely to be more accurate when using a large number of concepts but its accuracy compared to another framework cannot be judged only based on the number of concepts. For example, LM4CV \cite{yan2023learning} and CDL \cite{zang2024pre} achieve higher accuracies using more concepts. Yet, LaBo uses a far larger number of concepts but generally attains less accuracy than CDL. This observation can partly be explained by the fact that with a large number of concepts, some of them are likely to be unimportant or noisy. 

Another important observation is that the number of concepts to achieve satisfactory accuracy depends on the number of classes in the dataset. However, because of the tendency for a large number of concepts to render CBMs difficult to interpret, some works opt to sacrifice accuracy for explainability. By employing filtering and other discriminative techniques, some state-of-the-art methods are able to reduce a large number of attributes to much smaller number of more discriminative subsets.   For instance, CDL \cite{zang2024pre} achieves decent accuracies on all evaluated datasets with number of concepts not more than the number of categories. Similarly, on the CUB dataset, LM4CV uses a much smaller number of concepts (i.e., 32) than classes  \cite{marconato2022glancenets} and is able to achieve a top-1 accuracy of 60.27.

\section{Discussions}

A great deal of value addition has been achieved in machine learning through explainability. Despite recent progress, researchers and practitioners continue to question state-of-the-art interpretability methods for not truly eliciting the rationales of model decisions. The main challenge is that most methods are concerned with attributing importance to low-level input features by scrutinizing the internal structure and mechanisms of the underlying machine learning models. These techniques are therefore not able to provide high-level, semantically faithful explanations about model decisions. The reasons for this behavior are obvious. First, machine learning methods mainly rely on learning useful correlations between input data and final predictions. Consequently, they are not expected to naturally gain high-level insights into model behavior beyond showing features which are frequently associated with given outcomes in the training data. However, as long as these methods produce consistently accurate attributions that can “communicate” clearly the intended explanation, they are sufficient for many use-cases. Moreover, in many practical application scenarios, XAI techniques only serve as an auxiliary function to complement users’ understanding by enabling additional insights rather than providing complete explanations of entire decision processes. For example, in a typical disease diagnosis using X-Ray images, an interpretability method that highlights suspected features or pixels can draw experts’ attention to review and verify the decision in a situation where it would have been difficult for the human eye to recognize the defect un-aided. Thus, if attribution methods are precise enough, they can satisfy most practical needs.
Unfortunately, as highlighted in this paper, popular attribution methods, particularly techniques based on gradients and class activation mapping, are susceptible to producing poor explanations with partial or incomplete coverage of relevant features. They may also show excessive noise, background activations or even learning spurious relationships in data and using these as the basis for their decisions. Explanatory interactive learning is a technique that is proposed to address this problem by leveraging human expertise to review and edit incorrect explanations or predictions. The corrected instances can be fed back into the model as counter examples to discourage the learner from relying on confounders for its decision. This approach not only enhances interpretability but it also helps to improve prediction accuracy by leveraging expert knowledge in correcting mispredictions made on the basis of undesirable rationales. Despite its appeal, the human effort required by the method to review individual samples is highly prohibitive, especially for large scale data. Furthermore, humans are limited in their ability to accurately delineate target boundaries, a weakness which could be severely aggravated by fatigue.  More recently, some studies seek to address these challenges by leveraging vision-language models like CLIP for generating improved saliency maps with minimal human effort. However, CLIP’s open-world knowledge may not be adequate in the given domain. Owing to this limitation, these frameworks usually involve fine-tuning the pre-trained vision-language model on appropriate image-text pairs. The resulting saliency maps are generally far superior to those generated using conventional class activation mapping techniques. However, these frameworks still struggle in leveraging coarse image level annotations to accurately supervise pixel level localization, especially for rare categories in the fine-tuning data. 

Owing to this challenge, we expect future work to fully leverage diffusion models (DMs) and optimize the VLM- DM framework to generate accurate saliency maps in a zero-shot manner by maximizing the open vocabulary recognition ability of CLIP and the fine-grained localization of diffusion models. Although some studies have already explored the combination of VLMs and diffusion models in semantic segmentation settings, so far impressive results have not been demonstrated in challenging settings such as small or inconspicuous target localization. Another open problem in this domain concerns handling multi-class recognition problems with vision-language models where the most prominent class often dominates and obscures the less prominent ones.

Also, as discussed in this paper, most XAI approaches, including the widely used attribution methods, produce explanations based on low-level feature correlations. As a result, their explanations do not connote semantically faithful relationships between input features and model predictions. Concept bottleneck models (CBMs) address this problem by learning human-interpretable concepts from manually annotated datasets and using them for explaining the rationales of network predictions. Despite their appeal, CBMs were limited to relatively simpler or smaller datasets such as CUB, for which concept labels exist. However, more recent studies have leveraged the immense knowledge of vision-language and large language models to automatically discover relevant concepts from data. Automating the concept annotation process has allowed the approach to be scaled to large and complex datasets like ImageNet and the Places365 scene recognition datasets. On these tasks, CBMs have surpassed their standard black box baselines while providing interpretability. Owing to this success, research interest in this area of explainable AI is growing fast with multiple works emerging to address key accuracy and explainability challenges. Unfortunately, there is a glaring lack of objective explainability metrics that enable developers and users to quantitatively measure and compare different techniques. Current methods of comparing the interpretabilities of these methods are based on qualitative results from user studies or relating the number of concepts to explainability. However, user understanding or preferences are highly subjective and while concept count alone cannot be used as a basis of comparison without considering the explainability of the concepts themselves. Therefore

The interpretability of large language models is another important subject that this paper addresses. Large language models encode substantial world knowledge and hence, boast incredible power and ability to solve many complex problems. They are usually prompted to generate desirable responses. However, in many specialized domains or applications, it takes expert knowledge and enormous care to design quality prompts required for optimum performance. Owing to this challenge, current approaches are difficult to scale to larger problems or highly complex use-cases. This limitation often leads to under-utilization of LLM power.  To address this drawback, future prompt designs could leverage LLM-based embodied agents, preferably, those with artificial general intelligence capability. These agents should understand human needs or preferences, and be able to align their own goals towards designing effective prompts that generate desired LLM predictions and explanations.

As with other categories of models, it is extremely challenging to evaluate the faithfulness of LLM explainability methods. For some LLM explanation types, it is necessary to not only report on the faithfulness or correctness, but also on the intelligibility and soundness.  For instance, if the explanations are expressed in natural language, it is important to evaluate how well they communicate the intended explanation to the stakeholder. For correctness evaluation, natural language explanations are difficult to objectively evaluate quantitatively. In the absence of ground truth dataset on correct explanations, human evaluators are usually employed to score results. This method may not be adequately reliable for cases which need subjective judgement. The human effort required also poses a significant limitation. To alleviate this challenge, LLMs are also sometimes fine-tuned to score explanations on their truthfulness. However, like human evaluators, the answers demanded by LLMs in this pipeline cannot be regarded as gold standards. Besides, different sets of metrics are used for evaluating different kinds of explanations. For example, attention-based explainability methods can be evaluated using feature perturbation techniques while user studies can be useful in assessing natural language explanations. These methods are difficult to unify for easy comparison of explanability methods. Therefore, better evaluation metrics are required to help developers and users select the right explainability methods for their needs.

\section{Conclusion}

Explainable artificial intelligence is attracting significant attention in research and in many real-world domains. This need is caused by the high utility of The importance of explainability stems from the extensive use of machine learning models in critical applications coupled with the opacity of their predictions that require transparency in the decision-making processes. This paper highlights the fundamental principles used in achieving the most important goals of model explainability. The study presents many useful insights on key methods and discusses the state of explainable AI with respect to contemporary needs. The paper also presents various ways in which state-of-the-art explainability methods help to improve the accuracy of machine learning frameworks.

The review shows that XAI has achieved a significant success and many powerful techniques have been developed across many domains. Many new techniques are being developed while limitations of current methods are being address. Despite this progress, a number of serious limitations still remain, some of which span the whole range of model types and application domains. Crucially, it is still challenging to establish reliable metrics and benchmarks to evaluate different explainability techniques. Given the importance of this need, we expect extensive effort to be committed to this goal.

 



\end{multicols}

\end{document}